%% file: PaperForReview.tex
\documentclass[10pt,twocolumn,letterpaper]{article}

\usepackage[pagenumbers]{cvpr} %

\usepackage{graphicx}
\usepackage{amsmath}
\usepackage{amssymb}
\usepackage{booktabs}

\usepackage{times}
\usepackage{epsfig}
\usepackage{grffile}
\usepackage{multirow}
\usepackage{comment}
\usepackage[export]{adjustbox}

\usepackage{url}

\usepackage{dsfont}
\usepackage{caption}
\usepackage{subcaption}

\makeatletter
\@namedef{ver@everyshi.sty}{}
\makeatother
\usepackage{tikz}
\usetikzlibrary{shapes, arrows, calc, positioning}

\usepackage[pagebackref,breaklinks,colorlinks,urlcolor=black]{hyperref}

\usepackage[capitalize]{cleveref}
\crefname{section}{Sec.}{Secs.}
\Crefname{section}{Section}{Sections}
\Crefname{table}{Table}{Tables}
\crefname{table}{Tab.}{Tabs.}

\newcommand{\rulesep}{\unskip\ \vrule\ }
\newcommand{\joinR}{\hspace{-.1em}}
\newcommand{\RomanI}{I}
\newcommand{\RomanII}{\mbox{\RomanI\joinR\RomanI}}
\DeclareMathOperator*{\argmin}{\arg\!\min}
\DeclareMathOperator*{\argmax}{\arg\!\max}

\def\cvprPaperID{10032} %
\def\confName{CVPR}
\def\confYear{2022}

\begin{document}

\title{Representing 3D Shapes with Probabilistic Directed Distance Fields}

\author{
Tristan Aumentado-Armstrong$^\text{1,2,3}$~~~~Stavros Tsogkas$^\text{1}$~~~~Sven Dickinson$^\text{1,2,3}$~~~~Allan Jepson$^\text{1}$ \\
$^\text{1}$Samsung AI Centre Toronto~~~~~~$^\text{2}$University of Toronto~~~~~~$^\text{3}$Vector Institute for AI\\
{\tt\small tristan.a@partner.samsung.com,~\{stavros.t,s.dickinson,allan.jepson\}@samsung.com}
}

\maketitle

\begin{abstract}
   Differentiable rendering is an essential operation in modern vision,
    allowing inverse graphics approaches to 3D understanding 
    to be utilized in modern machine learning frameworks.
Explicit shape representations (voxels, point clouds, or meshes), 
    while relatively easily rendered,
    often suffer from limited geometric fidelity 
    or topological constraints.
On the other hand, 
    implicit representations 
    (occupancy, distance, or radiance fields) 
    preserve greater fidelity,
    but suffer from complex or inefficient rendering processes, 
    limiting scalability.
In this work, we endeavour to address both shortcomings with
    a novel shape representation that allows 
    fast differentiable rendering within an implicit architecture.
Building on implicit distance representations,
    we define \emph{Directed Distance Fields (DDFs)},
    which map an oriented point (position and direction) to surface visibility and depth.
Such a field can 
    render a depth map with a single forward pass per pixel,
    enable differential surface geometry extraction (e.g., surface normals and curvatures) via network derivatives,
    be easily composed,
    and permit extraction of classical unsigned distance fields.
Using probabilistic DDFs (PDDFs), 
    we show how to model inherent discontinuities in the underlying field.
Finally, we apply our method to fitting single shapes, unpaired 3D-aware generative image modelling, and single-image 3D reconstruction tasks, showcasing strong performance with simple architectural components via the versatility of our representation.
\end{abstract}

\newcommand{\fwone}{0.16\textwidth}

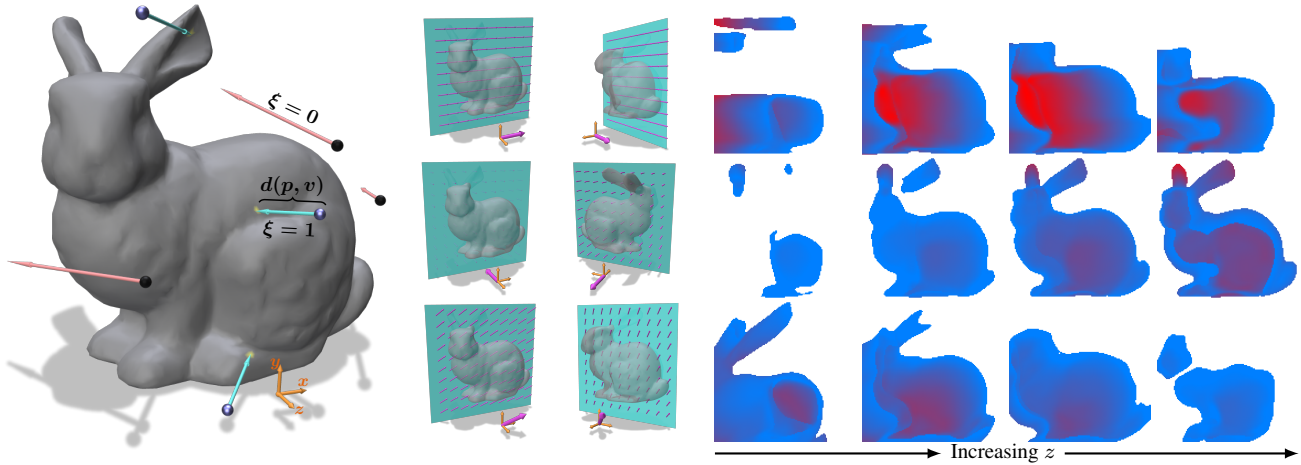
\begin{figure*}[t]
    \centering
    \begin{minipage}[c]{0.31\textwidth}%
    \includegraphics[width=0.99\textwidth]{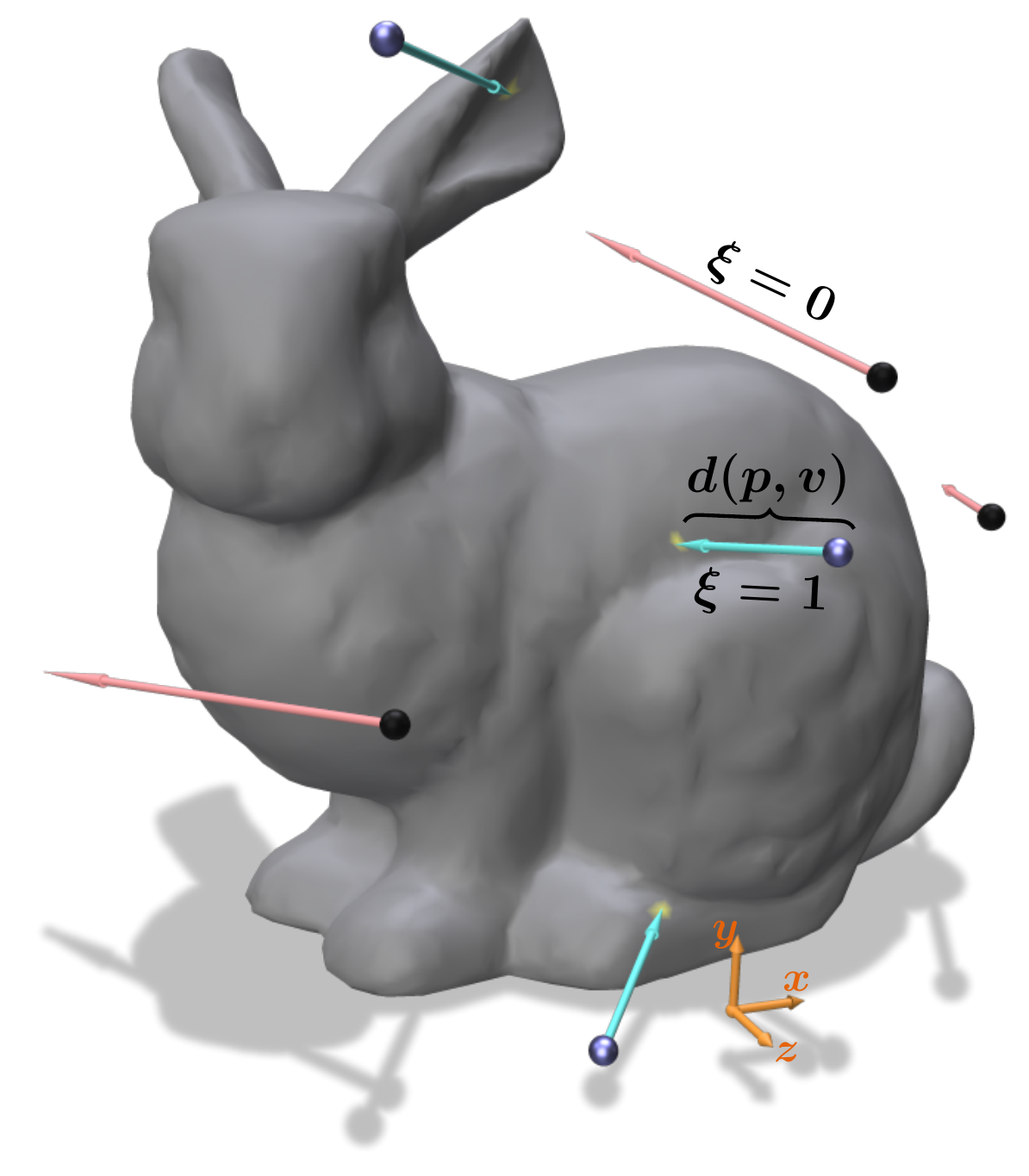}
    \end{minipage}\hfill%
    \begin{minipage}[l]{0.67\textwidth}%
        \includegraphics[height=\fwone]{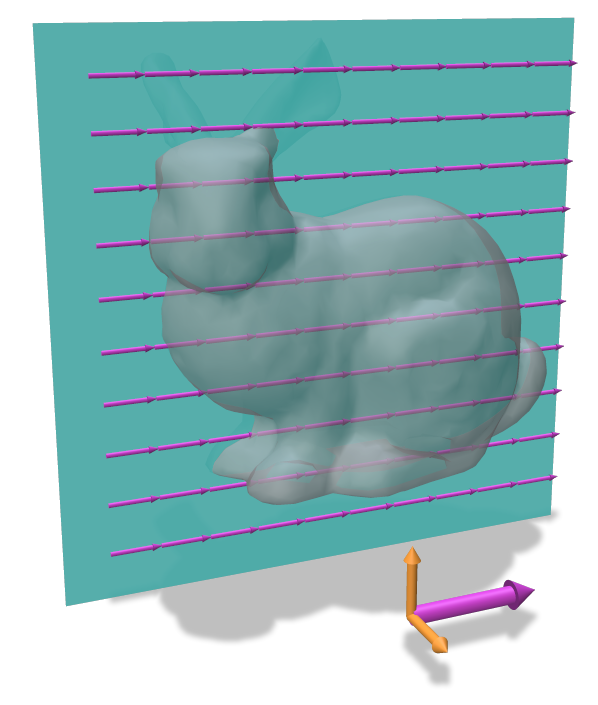}\hfill
        \includegraphics[height=\fwone]{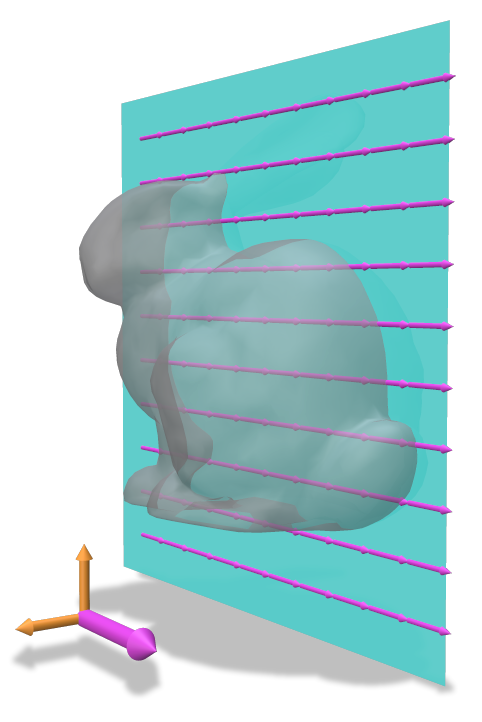}\hfill
        \includegraphics[height=\fwone]{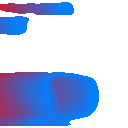}
        \includegraphics[height=\fwone]{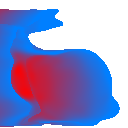}
        \includegraphics[height=\fwone]{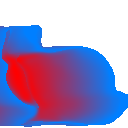}
        \includegraphics[height=\fwone]{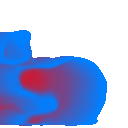}
        \\
        \includegraphics[height=\fwone]{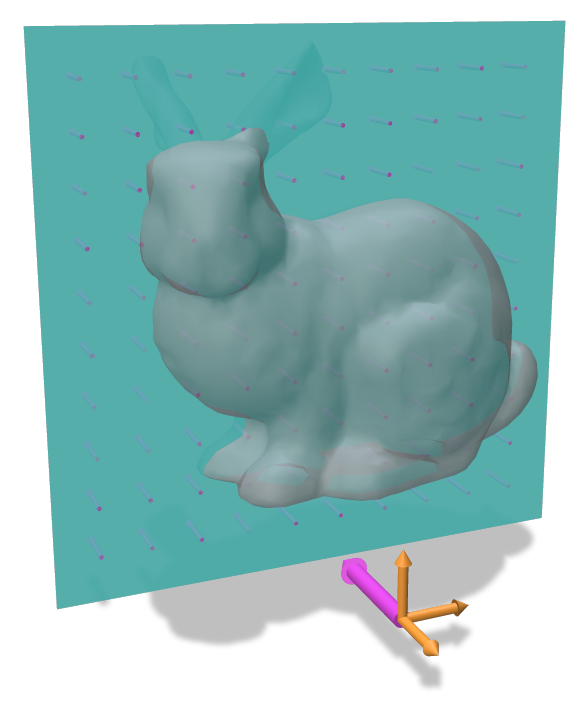}\hfill
        \includegraphics[height=\fwone]{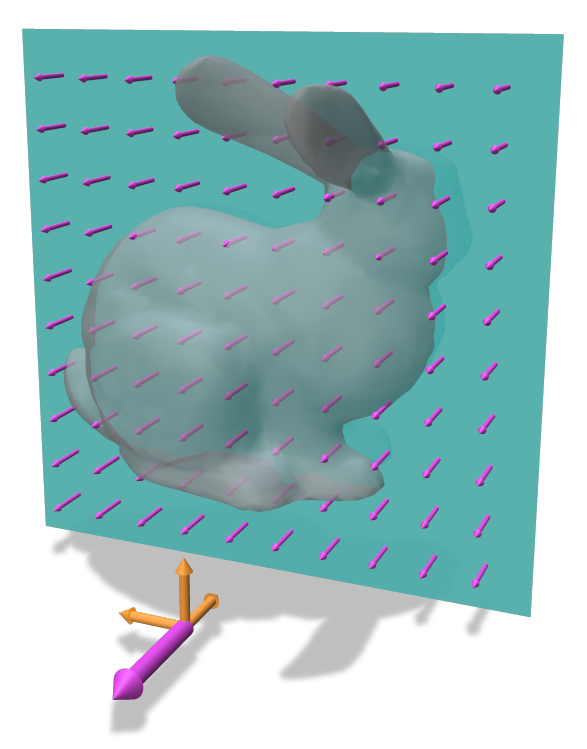}\hfill
        \includegraphics[height=\fwone]{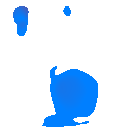}       
        \includegraphics[height=\fwone]{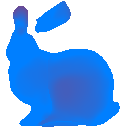}
        \includegraphics[height=\fwone]{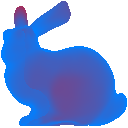}
        \includegraphics[height=\fwone]{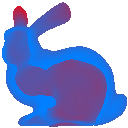}
        \\
        \includegraphics[height=\fwone]{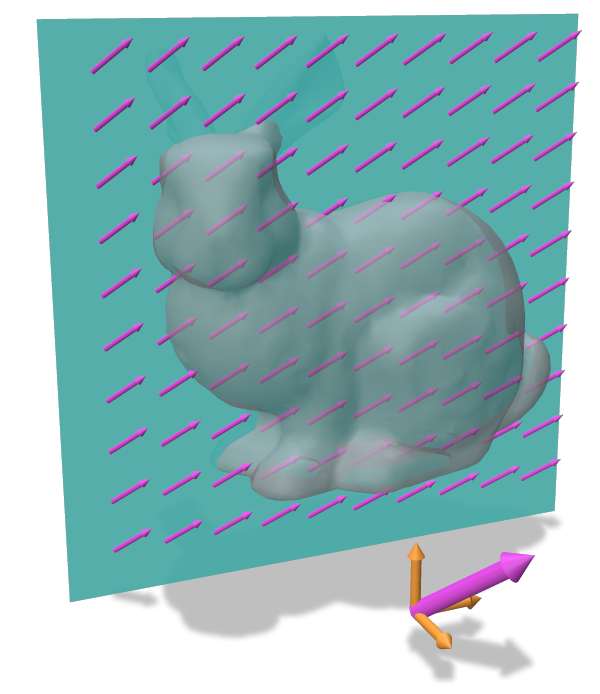}\hfill
        \includegraphics[height=\fwone]{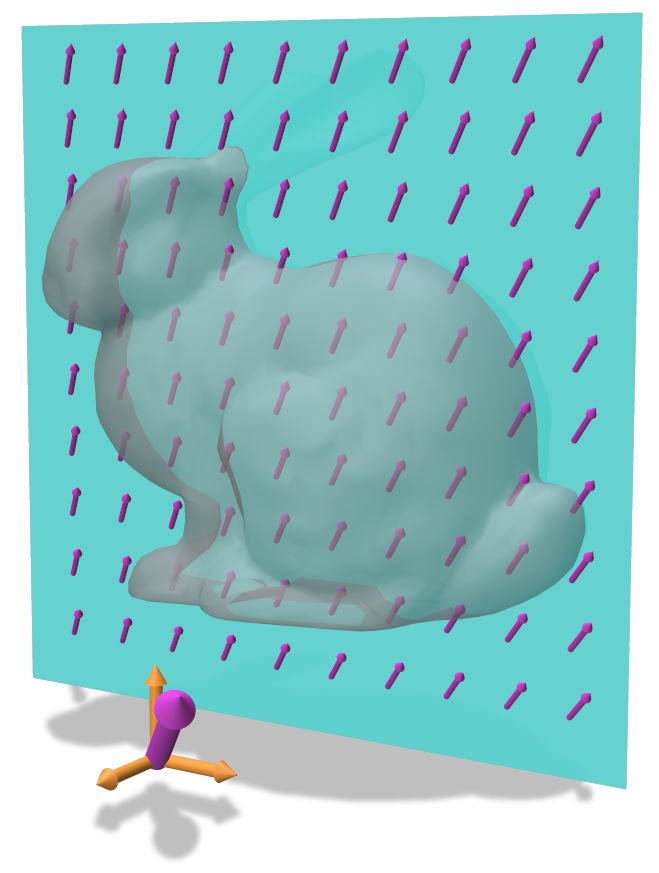}\hfill
        \includegraphics[height=\fwone]{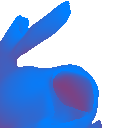}
        \includegraphics[height=\fwone]{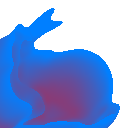}
        \includegraphics[height=\fwone]{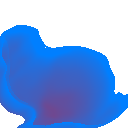}
        \includegraphics[height=\fwone]{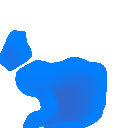}
    \end{minipage}
    \makebox[0pt][r]{%
    \begin{tikzpicture}[overlay]
    \node[xshift=0cm,yshift=0.4mm] (hidden) at (0,0) {};
    \node[xshift=1cm,yshift=0.4mm] (s) at (-8.75,-2.95) {};
    \node[xshift=1cm,yshift=0.4mm] (e) at (-0.99,-2.95) {};
    \node[xshift=1cm,yshift=0.4mm] (t) at (-4.92,-2.95) {\footnotesize Increasing $z$};
    \draw [-latex,thick] (s.center) --  node[right,above] {}(t.west) ;
    \draw [-latex,thick] (t.east) --  node[right,above] {}(e.center) ;
    \end{tikzpicture}
    }
    \caption{Illustrative example of a directed distance field, 
    fit to the Stanford bunny.
    \textit{Left}: depiction of visible oriented points (blue points, turquoise directions) that intersect the shape and those that miss the shape (black points, red directions) with $\xi = 0$. 
    \textit{Middle}: per row, illustrations of one slice plane (from two different views) and the fixed $v$ vector per slice plane (pink arrows), corresponding to the insets on the right (i.e., $v$ is the same across all $p$ for each row). 
    \textit{Right}: resulting depth field evaluated across positions at fixed orientations $v$ (rows: top, middle, and bottom show different $v$ values, parallel to $(1,0,0)$, $(0,0,-1)$, and $(1,1,1)$, respectively; columns: different slices in 3D with each having fixed $z$). Each pixel value is coloured with the distance value $d(p,v)$ obtained for that position $p$ and direction $v$ (red to blue meaning further to closer).
    Non-visible oriented points ($\xi=0$) are shown as white. 
    Notice the depth changes at intersections between the slice plane and shape (i.e., when $p$ moves through $S$).
    } %
    \label{fig:teaser}
\end{figure*}

\section{Introduction}

Three-dimensional shapes are represented in a variety of ways in modern computer vision and machine learning systems, 
with differing utilities depending on the task to which they are applied.
Recent advances in representation learning, however, capitalize on the inherent 3D structure of the world, 
and its link to generating the 2D images seen by our eyes and algorithms, 
via \textit{differentiable rendering} procedures compatible with neural network architectures. 
This enables an analysis-by-synthesis paradigm~\cite{yuille2006vision} that treats vision as ``inverse graphics''~\cite{kulkarni2015deep,romaszko2017vision},
wherein the model attempts to infer the 3D factors 
(e.g., shape, pose, texture, lighting)
that gave rise to its 2D perceptions. %

This can permit learning more powerful representations with weaker supervision.
Neural radiance fields (NeRFs) \cite{mildenhall2020nerf}, for instance, can be used for 3D inference \cite{yu2021pixelnerf} and 3D-aware generative image modelling \cite{chan2021pi,niemeyer2021giraffe,schwarz2020graf}, trained entirely on 2D data.
Similarly, implicit geometric fields, 
    such as occupancy fields \cite{mescheder2019occupancy}
    and
    signed distance fields (SDFs) \cite{park2019deepsdf},
    have recently been used in conjunction with differentiable renderering as well
    \cite{liu2020dist,niemeyer2020differentiable,jiang2020sdfdiff}.
Other works have learned textured mesh inference and/or generation
via rendering-based approaches 
(e.g., \cite{henderson2020leveraging,aumentado2020cycle,pavllo2020convolutional,goel2020shape,tulsiani2020implicit}).

Nevertheless, 
    it is still not always clear which representation is best for a given task.
Voxels and point clouds tend to have reduced geometric fidelity, while meshes suffer from the difficulties inherent in discrete structure generation, often leading to topological and textural fidelity constraints, or dependence of rendering efficiency on shape complexity \cite{liu2019soft}.
While implicit shapes can have superior fidelity, they struggle with complex or inefficient rendering procedures, requiring multiple network forward passes and/or complex calculations per pixel \cite{mildenhall2020nerf,sitzmann2019scene,liu2020dist}, and may be difficult to use for certain tasks (e.g., deformation, segmentation, or correspondence).
Thus, a natural question is how to design a method capable of fast differentiable rendering, yet still retaining high-fidelity geometric information 
that is useful for a variety of downstream applications. %

In this work, 
we explore \textit{directed distance fields} (DDFs), a representation that 
(i) captures the detailed geometry of a scene or object, including higher-order differential quantities and internal geometry,
(ii) can be differentiably rendered efficiently, compared to common implicit shape or radiance-based approaches,
(iii) is trainable with (pointwise) depth data,
(iv) can be easily composed, 
and 
(v) allows extraction of classical unsigned distance fields.
The definition is simple: 
for a given shape, we learn a field that maps any position and orientation to 
\textit{visibility} 
(i.e., whether the surface exists from that position along that direction) 
and 
\textit{distance}  
(i.e., how far the surface is along that ray, if it is visible).
Fig.\ \ref{fig:teaser} illustrates how DDFs can be viewed as implicitly storing all possible depth images of a given shape (i.e., from all possible cameras), 
reminiscent of a light field, but with geometric distance instead of radiance (see Fig.\ \ref{fig:teaser}).
Such a field is inherently discontinuous (see Fig.\ \ref{fig:discontillus}), 
presenting issues for differentiable neural networks, 
but has a powerful advantage in rendering, since a depth image can be computed with a single forward pass per pixel.
Its high input dimension (5D) incurs greater difficulty in learning,
    but the additional information 
    increases its versatility (e.g., higher-order local geometry, internal structure);
    furthermore, several geometric properties define constraints on the field and its derivatives, reducing the effective degrees of freedom.
We summarize our contributions as follows:
\begin{enumerate}
    \item We define directed distance fields (DDFs), a 5D mapping from any position and viewpoint to depth and visibility (\S\ref{sec:ddf:def}), 
    and a probabilistic variant (PDDFs) that can model surface and occlusion discontinuities (\S\ref{sec:app:pddfs}).    %
    \item By construction, our representation allows differentiable rendering via a single forward pass per pixel (\S\ref{sec:app:rendering}), 
    without restrictions on the shape (topology, water-tightness) or field queries (internal structure).
    \item We prove several geometric properties of DDFs %
    (\S\ref{sec:ddf:properties}), and use them in our method. %
    \item We apply DDFs to fitting shapes (\S\ref{sec:results:singlefieldfitting}), %
    single-image reconstruction (\S\ref{sec:app:si3dr}), %
    and generative modelling (\S\ref{sec:genmodel}).
\end{enumerate}

\section{Related Work}

\textbf{Implicit Shape Representations}~
Our work is most similar to distance field representations of shape, 
which have a long history in computer vision \cite{rosenfeld1968distance}, most recently culminating in signed and unsigned distance fields (S/UDFs) \cite{park2019deepsdf,chibane2020neural,venkatesh2020dude}. %
In comparison to explicit representations, implicit shapes can capture arbitrary topologies with high fidelity.
Several works examine differentiable rendering of implicit fields \cite{liu2020dist,jiang2020sdfdiff,niemeyer2020differentiable,sitzmann2019scene,liu2019learning,takikawa2021neural} (or combine it with neural volume rendering \cite{Kellnhofer:2021:nlr,oechsle2021unisurf,yariv2021volume,wang2021neus}).
In contrast, by conditioning on both viewpoint and position, DDFs can flexibly render depth, with a single field query per pixel.
Further, a UDF can actually be extracted from a DDF
    (see \S\ref{sec:ddf:properties}).

The closest current model to ours is the Signed Directional Distance Field (SDDF), 
    independently and concurrently developed by
    Zobeidi et al.\ \cite{zobeidi2021deep}, 
    which also maps position and direction to depth.
However, the lack of a sign in DDFs introduces a fundamental difference 
    in structure modelling:
starting from a point $p$, consider a ray that intersects with a wall; 
evaluating a DDF at a point after the intersection provides 
the distance to the next object, while the SDDF continues to measure the signed distance to the wall.
This reduces complexity and dimensionality, 
    but may limit representational utility for some tasks and/or shapes.

\textbf{Neural Radiance Fields}~
NeRFs \cite{mildenhall2020nerf} are powerful 3D representations, capable of novel view synthesis for reconstruction \cite{yu2021pixelnerf} and image generation \cite{chan2021pi} with very high fidelity. 
However, the standard differentiable volume rendering formulation of NeRFs is computationally expensive, requiring many forward passes per pixel, though recent work has improved on this 
(e.g., \cite{barron2021mip,garbin2021fastnerf,hedman2021baking,rebain2021derf,reiser2021kilonerf,lindell2021autoint,yu2021plenoctrees,Kellnhofer:2021:nlr}).
Furthermore, the distributed nature of the density makes extracting explicit geometric details (including higher-order surface information) more difficult (e.g., \cite{yariv2021volume,oechsle2021unisurf}).

Most similar to DDFs are Light Field Networks (LFNs) \cite{sitzmann2021light},
    which enable rendering with a single forward pass per pixel, 
    and permit sparse depth map extraction (assuming a Lambertian scene).
Unlike LFNs, DDFs model geometry rather than radiance as the primary quantity, 
    computing depth with a single forward pass, and surface normals with a single backward pass,
    while LFNs predict RGB and sparse depth from such a forward-backward operation.
Finally, the 4D parameterization of LFNs 
does not permit rendering from viewpoints between occluded objects.

\section{Directed Distance Fields}

\label{sec:ddf:def}

\textbf{Definition}
Let $S \subset \mathcal{B}$ be a 3D shape, 
where $\mathcal{B} \subset \mathbb{R}^3$ is a bounding volume that will act as the domain of the field. 
Consider a position $p\in\mathcal{B}$ and view direction $v\in\mathbb{S}^2$.
We define $S$ to be \textit{visible} from an oriented point $(p,v)$ 
if the line $\ell_{p,v}(t) = p + tv$ 
intersects $S$ for some $t \geq 0$.
We write the binary visibility field for $S$ as
$\xi(p,v) = \mathds{1}[(p,v)\text{ is visible}]$. 
For convenience, we refer to an oriented point $(p,v)$ as visible if $\xi(p,v)=1$.

We then define a \textit{directed distance field} (DDF) as a non-negative scalar field $d : \mathcal{B} \times \mathbb{S}^2 \rightarrow \mathbb{R}_+$,
which maps from any visible position and orientation in space to the minimum distance from $p$ to $S$ along $v$ (i.e., the first intersection of $\ell_{p,v}(t)$ with $S$).
In other words,
$q(p,v) = d(p,v) v + p$ 
is a map to the shape, and thus
satisfies $q(p,v) \in S$ for visible $(p,v)$ (meaning $\xi(p,v) = 1$).
See Fig.\ \ref{fig:teaser} for an illustration.

\begin{figure}
    \centering
    \includegraphics[width=0.19\textwidth]{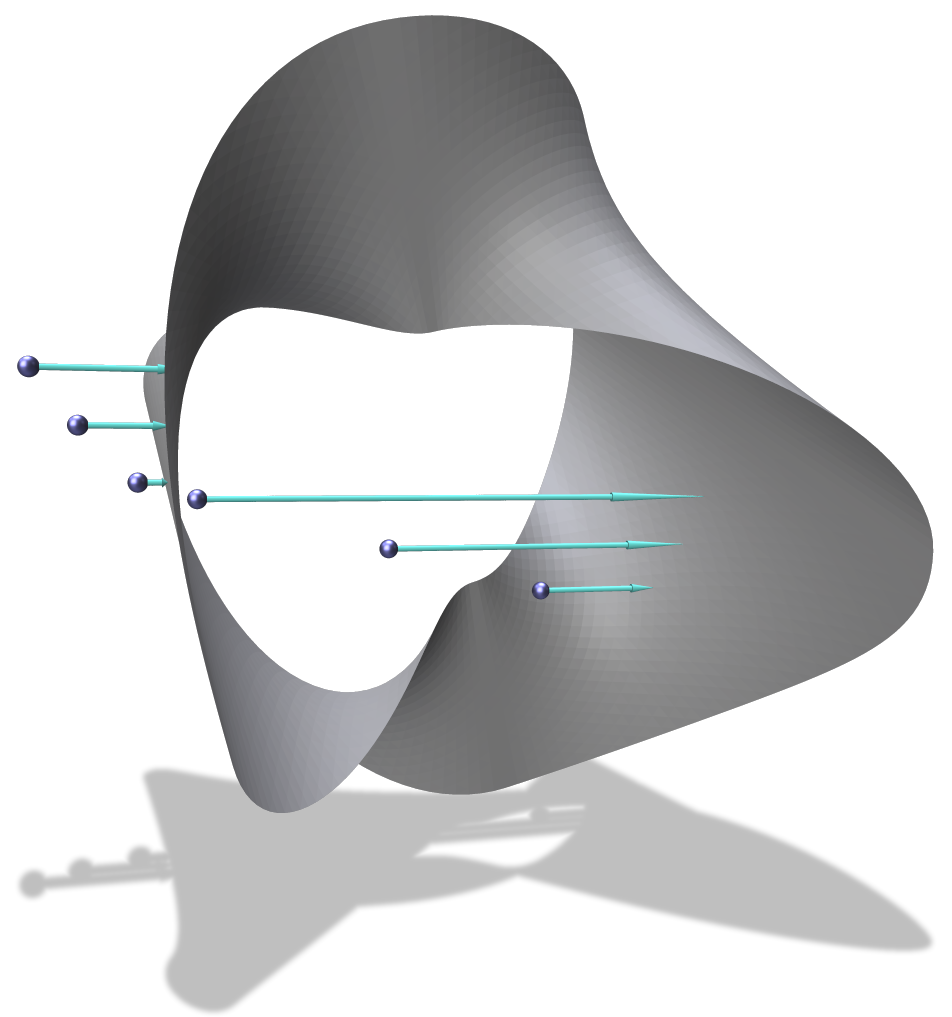}\rulesep
    \includegraphics[width=0.195\textwidth]{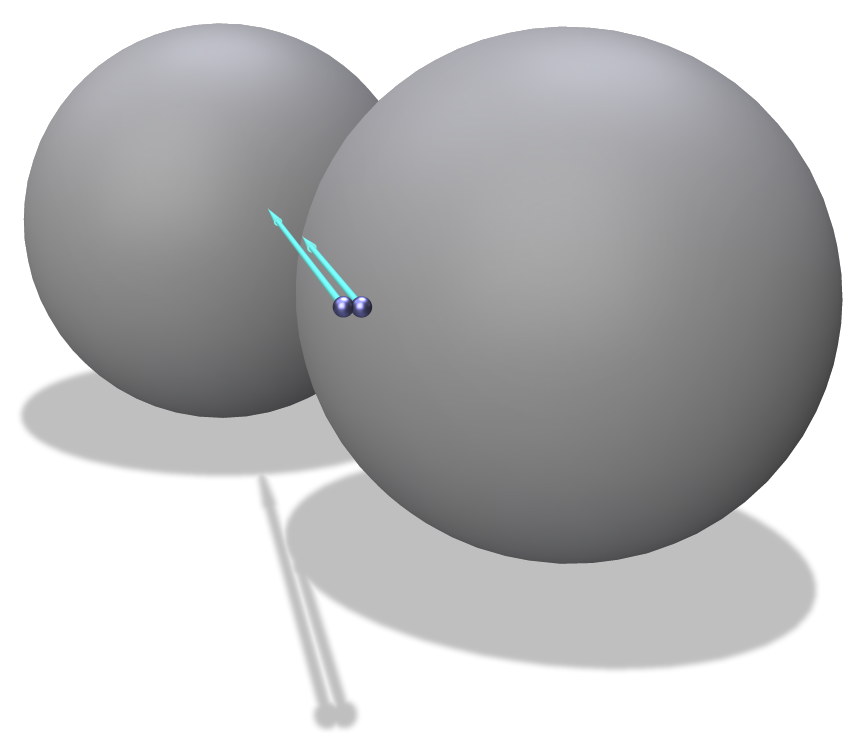}
    {%
    \setlength{\fboxsep}{0pt}%
    \setlength{\fboxrule}{0.5pt}%
    \raisebox{0.25\height}{%
    \fbox{\includegraphics[width=0.04\textwidth]{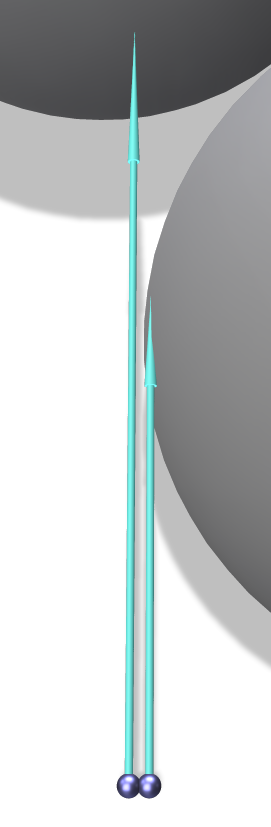}}%
    }%
    }%
    \caption{
    Inherent discontinuities with DDFs.
    Left: \textit{surface discontinuities}, where $p$ passes through $S$. Right: \textit{occlusion discontinuities}, where $v$ or $p$ is moved over an occluding boundary edge. }
    \label{fig:discontillus}
\end{figure}

\subsection{Geometric Properties}
\label{sec:ddf:properties}

DDFs satisfy several useful geometric properties.
We provide proofs in Appendix \ref{appendix:proofs}.

\phantomsection
\label{property1}
\textbf{Property I: Directed Eikonal Equation.}
Similar to SDFs,
which satisfy the eikonal equation 
$||\nabla_p \text{SDF}(p)||_2 = 1$, 
a DDF enforces a directed version of this property.
In particular, 
    for any visible $(p,v)$,
    we have 
    $ \nabla_p d(p,v) v = -1 $, with $\nabla_p d(p,v)\in\mathbb{R}^{1\times 3}$.
Note this implies 
$||\nabla_p d(p,v)||_2 \geq 1$ 
as well.
There is also a directed eikonal property for the visibility field,
as
locally moving along the viewing line cannot change visibility:
$ \nabla_p \xi(p,v) v = 0 $.

\phantomsection
\label{property2}
\textbf{Property II: Surface Normals.}
The derivatives of implicit fields are closely related to the surface normals $n \in\mathbb{S}^2$ 
of $S$;
e.g., $\nabla_q \text{SDF}(q)^T = n(q)$ for any $q\in S$.
For DDFs, a similar relation holds (\textit{without} requiring $p\in S$):
$ %
\nabla_p d(p,v) = {-n(p,v)^T} / ({n(p,v)^T v}),
$ %
for any visible $ (p,v) $ such that
$n(p,v) := n(q(p,v))$
are the normals at
$q(p,v) = d(p,v) v + p \in S$
and $n(p,v) \not\perp v$ 
(i.e., the change in $d$ moving \textit{off} the surface is undefined).
This allows recovering the surface normals of any point $q \in S$, simply by querying any $(p,v)$ on the line that ``looks at'' $q$, and computing 
$ n(p,v) = 
 \varsigma \nabla_p d(p,v)^T / ||\nabla_p d(p,v)||_2 $,
 where we choose $\varsigma \in \{-1,1\}$
 such that $n^Tv < 0$ 
 (so that $n$ always points back to the query oriented point%
)\footnote{This defines the normal via $v$, even for non-orientable surfaces.}. In this sense, $n(p,v)$ is the visible surface normal on $S$, as seen from $(p,v)$. 

\phantomsection
\label{property3}
\textbf{Property III: Gradient Consistency.}
Consider a visible $(p,v)$.
Notice that changing the viewpoint by some infinitesimal $\delta_v$ would seem to have a similar effect as pushing the position $p$ in the direction $\delta_v$.
In fact, it can be shown that
$ \nabla_v d(p,v) \delta_v  = d(p,v) \nabla_p d(p,v) \delta_v $, where $\delta_v = \omega \times v $ for any $\omega \in \mathbb{R}^3$.
This relates the directional derivatives of $d$,
along a rotational perturbation $\delta_v$, 
with respect to both viewpoint and position (see also Appendix \ref{appendix:proofs:prop3} for alternative expressions).
As with Property \hyperref[property1]{I},
any $d$ must satisfy gradient consistency to be a true DDF.

\phantomsection
\label{property4}
\textbf{Property IV: Deriving Unsigned Distance Fields.}
We remark that an unsigned distance field (UDF) can be extracted from a DDF via the following optimization problem:
$ \text{UDF}(p) = \min_{v \in \mathbb{S}^2} d(p,v) $, constrained such that $\xi(p,v) = 1$, allowing them to be procured if needed (see \S\ref{sec:app:udfextract}).
UDFs remove the discontinuities from DDFs (see \S\ref{sec:app:pddfs} and Fig.\ \ref{fig:discontillus}), but are not rendered as easily nor can they be queried for distances in arbitrary directions.

\phantomsection
\label{property5}
\textbf{Property V: Local Differential Geometry.}
For any visible $(p,v)$, the geometry of a 2D manifold $S$ near $q(p,v)$ is completely characterized by $d(p,v)$ and its derivatives. 
In particular,
we can estimate the first and second fundamental forms 
using the gradient and Hessian of $d(p,v)$ 
(see Appendix \ref{appendix:proofs:prop5}).
This allows computing surface properties, 
such as curvatures,
from any visible oriented position, simply by querying the network; 
see Fig.\ \ref{fig:singobjfits}
for an example.

\textbf{Neural Geometry Rendering.} 
Many methods utilize differentiable rendering of geometric quantities, such as depth and surface normals 
(e.g., \cite{yan2016perspective,wu2017marrnet,tulsiani2017multi,nguyen2018rendernet}).
Often, such methods can be written as parallelized DDFs (see Appendix \ref{appendix:neurren}).
Thus, the properties above hold, regardless of architecture;
we believe this can improve 
such %
frameworks.

\subsection{Rendering} 
\label{sec:app:rendering}

A primary application of DDFs is rapid differentiable rendering.
In contrast to some differentiable mesh renderers (e.g., \cite{liu2019soft}), there is no dependence on the complexity of the underlying shape, after training.
Unlike classical NeRFs \cite{mildenhall2020nerf} or other standard implicit shape fields \cite{liu2020dist,sitzmann2019scene}, 
DDFs only require a single forward pass per pixel.

The process itself is a straightforward ray casting procedure. 
Given a camera with position $p_0\in\mathcal{B}$, for a pixel with 3D position $\rho$, we effectively cast a ray $r(t) = p_0 + t v_\rho $ with $v_\rho = (\rho - p_0)/||\rho - p_0||_2$ into the scene with a single query $d(p_0, v_\rho)$, which provides the depth image pixel value.
Note that $\rho$, and thus $d(p_0, v_\rho)$, 
depend on the camera parameters.
Finally, consider $p\notin \mathcal{B}$.
In this case, we first compute the intersection point $p_r\in\partial\mathcal{B}$ between the ray $r$ and the boundary $\partial \mathcal{B}$. 
We then use $d(p_r,v) + ||p - p_r||_2$ as the output depth (or set $\xi(p_r,v)=0$ if no intersection exists).
This allows querying the network from arbitrary positions and directions, including those unseen in training.

\newcommand{\ctwa}{0.19\textwidth}
\newcommand{\ctwb}{0.159\textwidth}
\begin{figure}
    \centering
    \begin{subfigure}[t]{.5\textwidth}
    \centering
    \includegraphics[width=\textwidth]{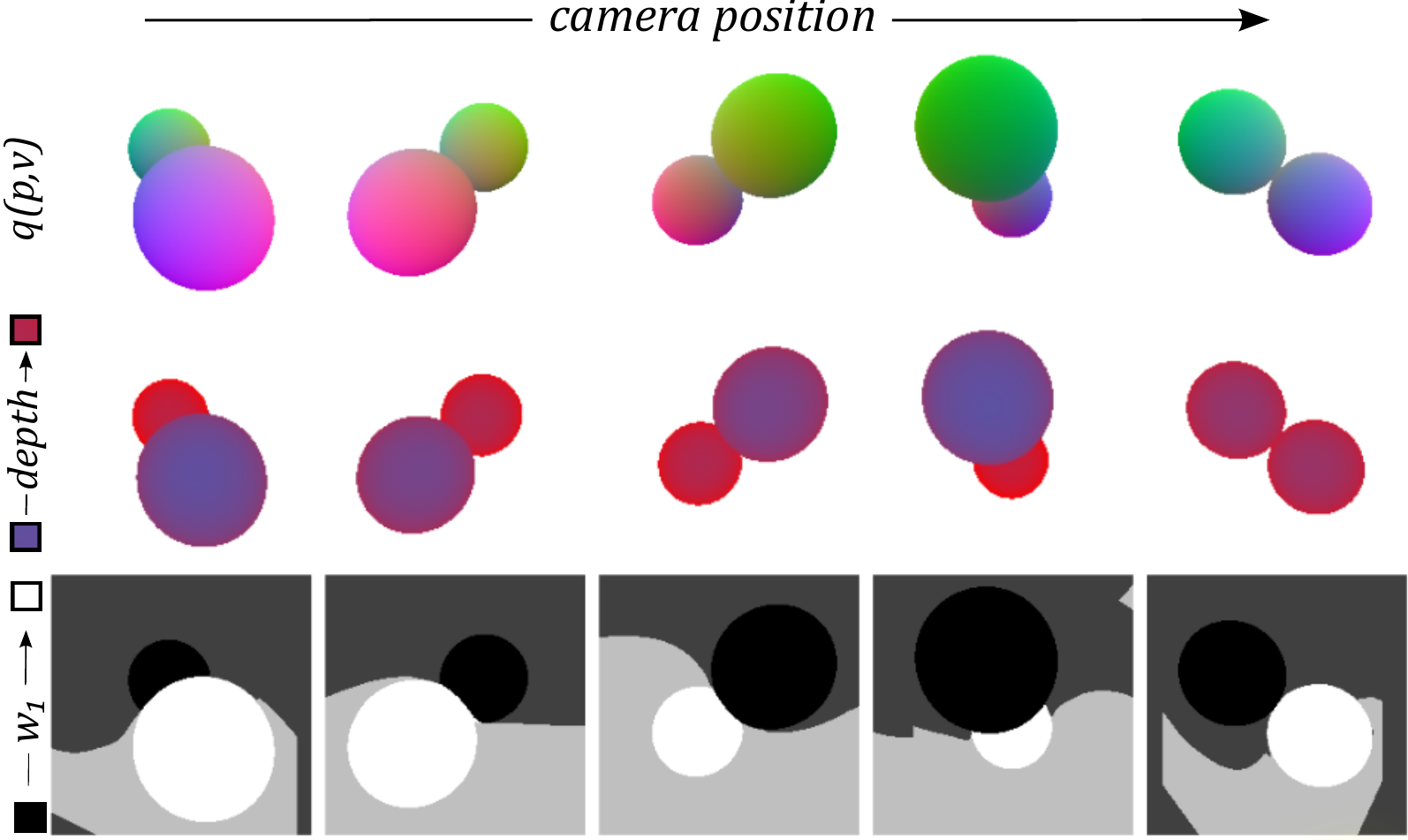}
    \caption{Weight field transitions in DDF renders. 
    In row three, white color marks high $w_1$ values, as well as which surface ($d_1$ vs $d_2$) is active. 
    Light and dark grey demarcate their non-visible counterparts, with low $\xi$. 
    The change in dominant weight ($w_1$ vs $1-w_1$) at occlusion edges permits discontinuities. }
  \end{subfigure}
  \begin{subfigure}[t]{.5\textwidth}
    \centering
    \includegraphics[width=\textwidth]{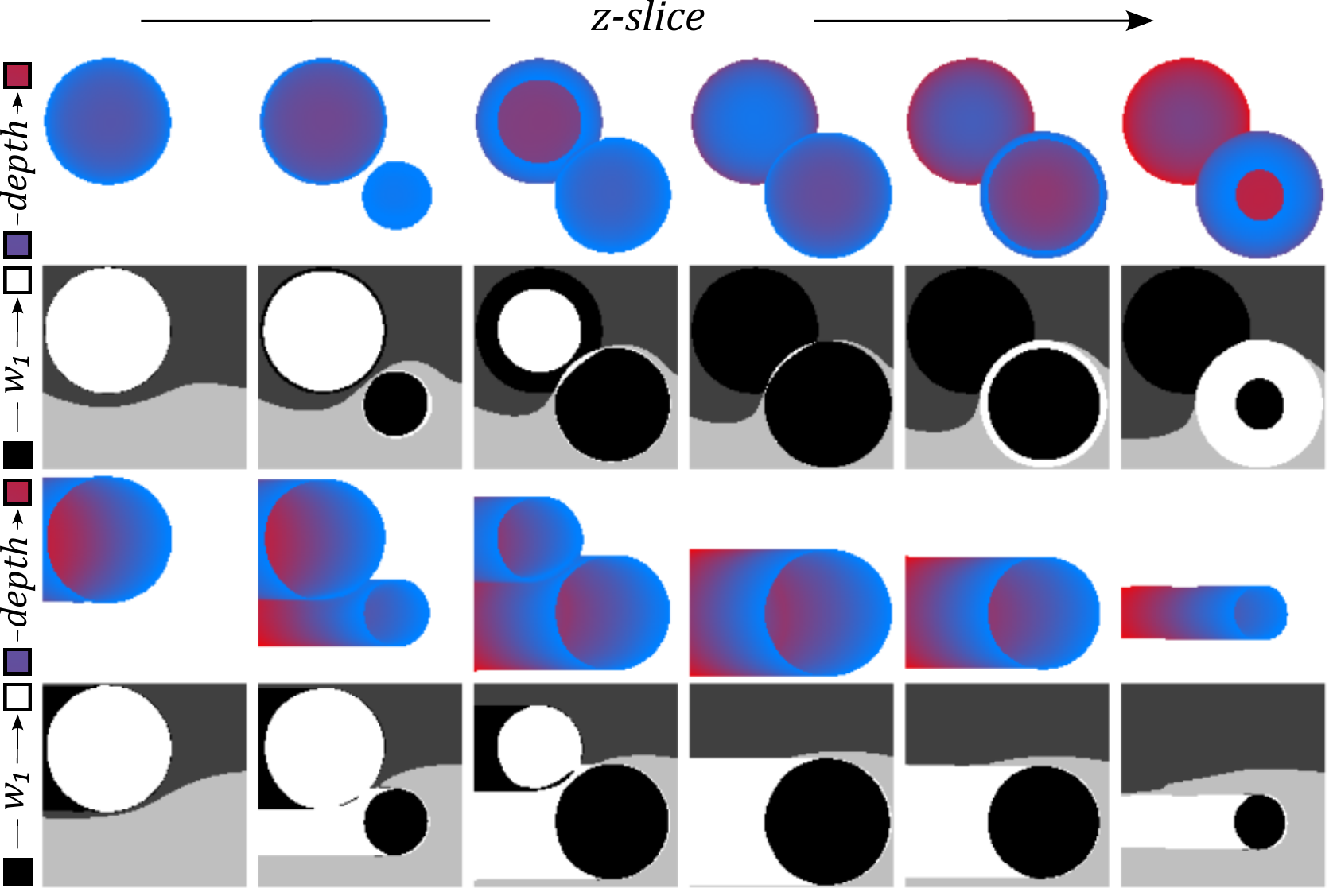}
    \caption{Weight field transitions using 3D slices in $z$. 
    Rows 1 and 3 depict (discontinuous) distance values, with fixed $v$ 
        ($(0,0,-1)$ and $(1,0,0)$, respectively) and varying $p$ across image pixels. 
    Rows 2 and 4 show weight values for $w_1$ and $\xi$, 
        as in (a) above.
    Notice the field switching upon $p$ transitioning through a surface discontinuity. }
  \end{subfigure}
    \caption{
    Illustration of probabilistic DDFs for discontinuous depth modelling, on a simple two-sphere scene with renders (a) and spatial slices (b). Here, $K = 2$, so $w_1 = 1 - w_2$ (see \S\ref{sec:app:pddfs}).
    }
    \label{fig:prob_illus}
\end{figure}

\newcommand{\rmtl}{0.071}
\newcommand{\rmtll}{0.066}
\begin{figure}
    \centering
    \includegraphics[width=\linewidth]{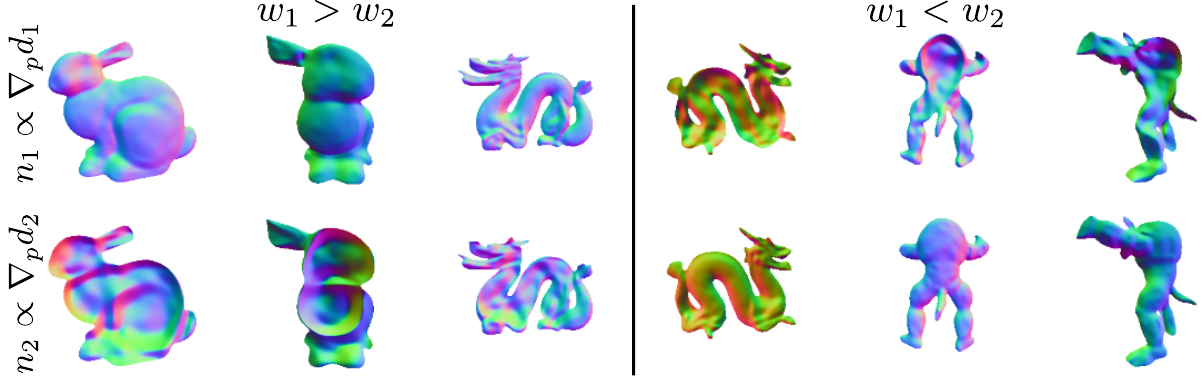}
    \caption{
        PDDF renders of $n_1$ and $n_2$.
        Though not explicitly enforced, a ``see-through effect'' occurs when the lower-weight field models the surface behind the currently visible one.
    }
    \label{fig:seethrough}
\end{figure}

\subsection{Discontinuity Handling: Probabilistic DDFs}
\label{sec:app:pddfs}

DDFs are inherently discontinuous functions of $p$ and $v$. 
As shown in Fig.\ \ref{fig:discontillus}, whenever 
(i) $p$ passes through the surface $S$ or 
(ii) $p$ or $v$ is moved across an occlusion boundary, a discontinuity in $d(p,v)$ will occur.
We therefore modify the DDF formulation, to allow a $C^1$ 
network to represent the discontinuous field.
In particular, we alter $d$ to output probability distributions over rays, rather than a single value.
Let $\mathcal{P}_\ell$ be the set of probability distributions with support on some ray $\ell_{p,v}(t) = p+tv,\, t\geq 0$.
Then $d : \mathcal{B} \times \mathbb{S}^2 \rightarrow \mathcal{P}_\ell$ is a \textit{probabilistic DDF} (PDDF). 
The visibility field, $\xi(p,v)$, is unchanged in the PDDF.

For simplicity, herein we restrict $\mathcal{P}_\ell$ to be the set of mixtures of Dirac delta functions with $K$ components.
Thus, the network output is a density field
$
P_{p,v}(d) = %
\sum_i w_i \delta(d - d_i)
$ 
 over depths,
where $w_i$'s are the mixture weights, with $\sum_i w_i = 1$, and $d_i$'s are the delta locations.
Our output depth is then $d_{i^*}$, where
$i^* = \argmax_i w_i$; i.e., the highest weight delta function marks the final output location.
As $w_i$ changes continuously, $w_{i^*}$ will switch from one $d_i$ to another $d_j$, which may be arbitrarily far apart, resulting in a discontinuous jump.
Thus, by having the weight field $w(p,v)$ smoothly \textit{transition} from one index ${i^*}$ to another, at the site of a surface or occlusion discontinuity, we can obtain a discontinuity in $d$ as desired.
In this work, we use $K=2$, to represent discontinuities without sacrificing efficiency.
Fig.\ \ref{fig:prob_illus} showcases example transitions, with respect to  (a) occlusion discontinuities and (b) surface collision; 
Fig.\ \ref{fig:seethrough} visualizes the difference in the normals fields.
Notationally, we may treat a PDDF as a DDF, by setting
$d(p,v) := d_{i^*}(p,v)$.

\newcommand{\ctwc}{0.077\textwidth}
\begin{figure*}[ht]
    \centering
    \includegraphics[width=\ctwc]{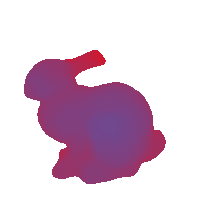}
    \includegraphics[width=\ctwc]{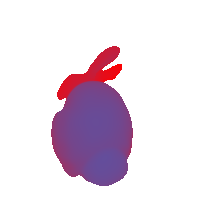}
    \includegraphics[width=\ctwc]{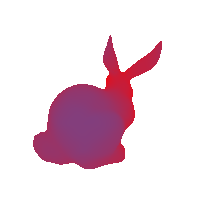}
    \includegraphics[width=\ctwc]{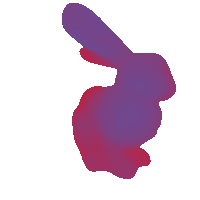} 
    \includegraphics[width=\ctwc]{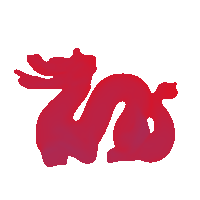}
    \includegraphics[width=\ctwc]{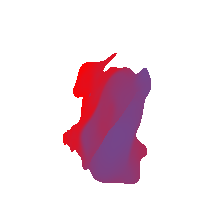}
    \includegraphics[width=\ctwc]{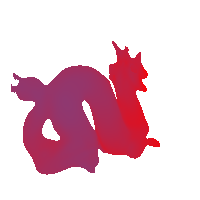}
    \includegraphics[width=\ctwc]{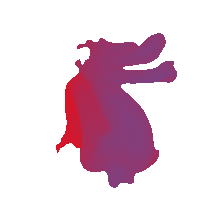}
    \includegraphics[width=\ctwc]{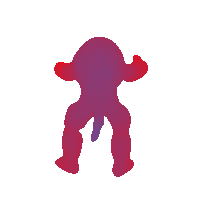}
    \includegraphics[width=\ctwc]{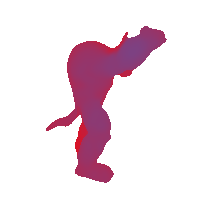}
    \includegraphics[width=\ctwc]{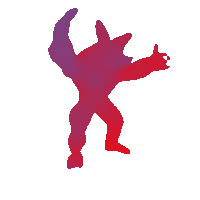}
    \includegraphics[width=\ctwc]{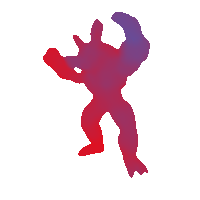}
    \includegraphics[width=\ctwc]{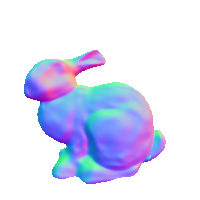}
    \includegraphics[width=\ctwc]{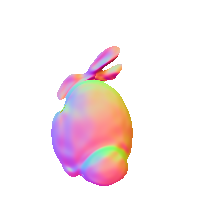}
    \includegraphics[width=\ctwc]{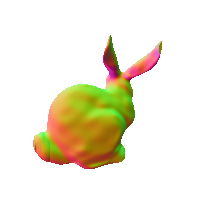}
    \includegraphics[width=\ctwc]{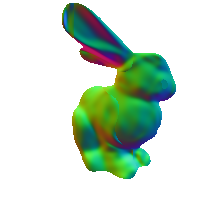} 
    \includegraphics[width=\ctwc]{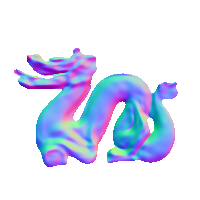}
    \includegraphics[width=\ctwc]{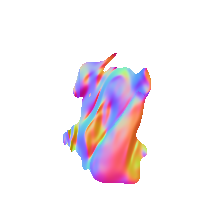}
    \includegraphics[width=\ctwc]{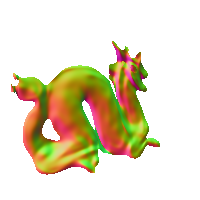}
    \includegraphics[width=\ctwc]{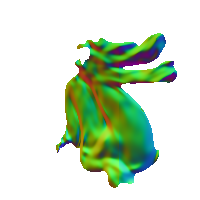}
        \includegraphics[width=\ctwc]{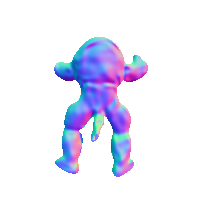}
    \includegraphics[width=\ctwc]{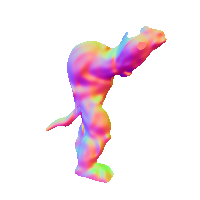}
    \includegraphics[width=\ctwc]{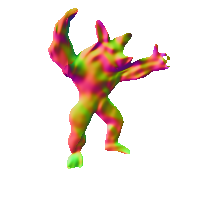}
    \includegraphics[width=\ctwc]{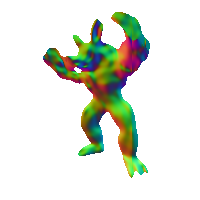}
    \includegraphics[width=\ctwc]{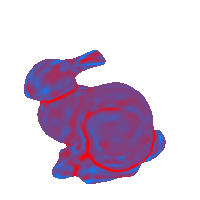}
    \includegraphics[width=\ctwc]{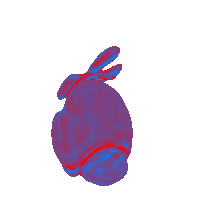}
    \includegraphics[width=\ctwc]{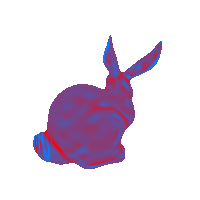}
    \includegraphics[width=\ctwc]{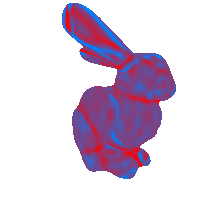} 
    \includegraphics[width=\ctwc]{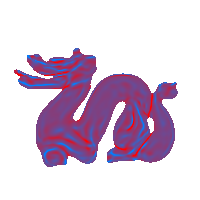}
    \includegraphics[width=\ctwc]{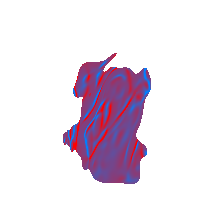}
    \includegraphics[width=\ctwc]{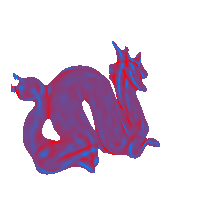}
    \includegraphics[width=\ctwc]{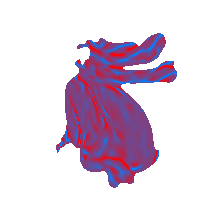}
    \includegraphics[width=\ctwc]{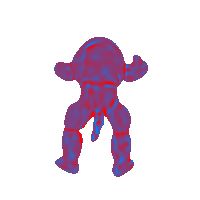}
    \includegraphics[width=\ctwc]{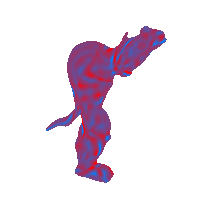}
    \includegraphics[width=\ctwc]{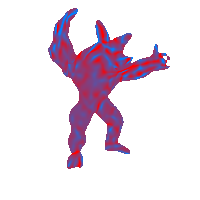}
    \includegraphics[width=\ctwc]{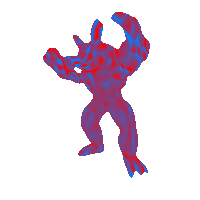}    
    \caption{
    Renders of DDF fits to shapes. %
    Rows: depth, normals, and mean curvature. %
    Columns: different camera positions per object. 
    Each quantity is directly computed from the learned field, using network derivatives at the query oriented point $(p,v)$ per pixel. 
    }
    \label{fig:singobjfits}
\end{figure*}

\subsection{Learning DDFs}
\label{sec:app:learning}

\label{sec:app:data}
\textbf{Mesh Data Extraction}
Given a mesh specifying $S$, 
we can obtain visibility $\xi$ and depth $d$
by ray-casting
from any %
$(p,v)$.
In total, we consider six types of data samples %
(visualized in Appendix \ref{sec:datatypes}):
\textit{uniform} (U) random samples for $(p,v)$; 
\textit{at-surface} (A) samples, where $(p,v)$ must be visible;
\textit{bounding} (B) samples, where $p\in\partial\mathcal{B}$ and $v$ points to the interior of $\mathcal{B}$;
\textit{surface} (S) samples, where $p\in S$ and $v$ is uniform;
\textit{tangent} (T) samples, 
where $v$ is in the tangent space of $q(p,v)\in S$;
and 
\textit{offset} (O) samples, 
which offsets $p$ from T-samples
along
$n(p,v)$ by a small value.

\textbf{Loss Functions}
Our optimization objectives are defined per oriented point 
$(p,v)$. 
We denote $\xi$, $n$, and $d$ 
as the ground truth visibility, surface normal, and depth values, 
and let $\widehat{\xi}$, $\widehat{d}_i$, and $w_i$ denote the network predictions.
Recall $i^* = \arg\max_{j} w_j$ is the maximum likelihood PDDF index.

The \textit{minimum distance loss} ensures that the correct depth is output for the highest probability component:
$%
    \mathcal{L}_d = \xi| \widehat{d}_{i^*} - d |^2.
$%
The \textit{visibility objective}, $L_\xi = \mathrm{BCE}(\xi,\widehat{\xi})$, is the binary cross entropy  between the visibility prediction and the ground truth.
A first-order \textit{normals loss} (as in \cite{gkioxari2019mesh}), 
$
\mathcal{L}_n = -\xi| n^T \widehat{n}_{i^*}(p,v) |,
$ 
uses Property \hyperref[property2]{II} to match surface normals to the underlying shape, 
via $ \nabla_p \widehat{d}_{i^*} $.
A \textit{Directed Eikonal regularization}, 
based on Property \hyperref[property1]{I}, is given by
\begin{equation}
    \mathcal{L}_{\mathrm{DE}} = 
    \gamma_{\mathrm{E},d}
    \sum_i \xi\left[ \nabla_p \widehat{d}_i v + 1 \right]^2
    +
    \gamma_{\mathrm{E},\xi}
    [\nabla_p \widehat{\xi} v]^2
    ,
\end{equation}
applied on the visibility and each delta component of $d$, analogous to prior SDF work
(e.g., \cite{gropp2020implicit,lin2020sdf,yang2021deep}).

Finally, we utilize two {weight field regularizations}, 
which encourage 
(1) low entropy PDDF outputs 
(to prevent $i^*$ from switching unnecessarily),
and
(2) the maximum likelihood delta component to \textit{transition} (i.e., change $i^*$) 
when a discontinuity is required:
$ 
\mathcal{L}_W = \gamma_V \mathcal{L}_V + \gamma_T \mathcal{L}_T 
$. 
The first is a \textit{weight variance loss}: 
$\mathcal{L}_V = \prod_i w_i$.
The second is a \textit{weight transition loss}:
$
\mathcal{L}_T = %
\max( 0, \varepsilon_T - |\nabla_p w_1 n| )^2,
$
where $\varepsilon_T$ is a hyper-parameter controlling the desired transition speed. %
Since $K=2$, using $w_1$ alone is sufficient to enforce changes along the normal. %
Note that $\mathcal{L}_T$ is \textit{only} applied to oriented points that we wish to undergo a transition 
(i.e., where a discontinuity is desired, 
as illustrated in Fig.\ \ref{fig:discontillus} and \ref{fig:prob_illus}), 
namely surface (S) and tangent (T) data. 
The complete PDDF shape-fitting loss is then
\begin{equation} \label{eq:singlefitall}
    \mathfrak{L}_S = \gamma_d \mathcal{L}_d +
                     \gamma_\xi \mathcal{L}_\xi +
                     \gamma_n \mathcal{L}_n +
                     \mathcal{L}_{\mathrm{DE}} +
                     \mathcal{L}_W.
\end{equation}
Other regularizations could be applied 
(e.g., gradient and view consistency; see Property \hyperref[property3]{III} and Appendix \ref{appendix:viewconsis}), 
but for simplicity we leave them to future work.

\section{Empirical Results}

\subsection{Single Field Fitting}
\label{sec:results:singlefieldfitting}

We use the SIREN neural architecture \cite{sitzmann2020implicit} for all field parameterizations,
as it both  allows for higher order derivative calculations and has shown powerful representational capabilities in other works 
(e.g., \cite{chan2021pi,martel2021acorn}).
We use $K=2$, an axis-aligned bounding box for $\mathcal{B}$, Adam \cite{kingma2014adam} for optimization, and PyTorch \cite{pytorch} for all implementations.
(See Appendix \ref{appendix:singlefits} for details.)
In Fig.\ \ref{fig:singobjfits}, we show results for fitting single objects, via PDDF renderings with a single network evaluation per pixel.
Surface normals and curvatures are obtained 
using only additional backward passes for the same oriented points used in the single forward pass.
Using SIREN enables fairly detailed renders of the geometry.

\begin{figure}
    \centering
    \includegraphics[width=\linewidth]{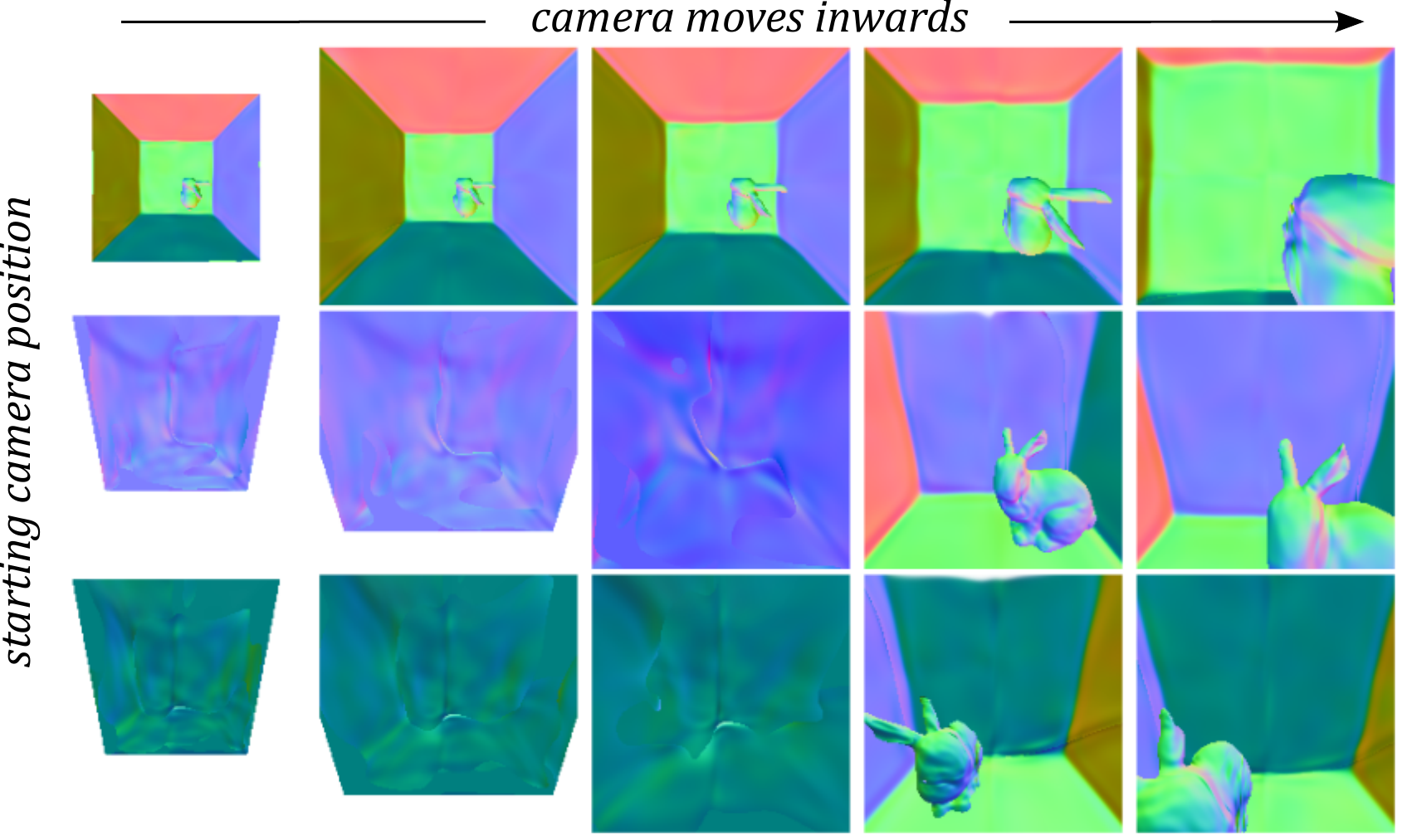}
    \caption{Example of internal structure rendering and compositional scene construction. 
             Colours correspond to surface normals 
             (as in Fig.\ \ref{fig:singobjfits}), 
             estimated via the DDF (with Property \hyperref[property2]{II}). }
    \label{fig:simpleroom}
\end{figure}

We discuss two additional modelling capabilities of DDFs: 
(i) internal structure representation
and
(ii) compositionality.
The first refers to the ability of our model to handle multi-layer surfaces:
we are able to place a camera inside a scene, within or between multiple surfaces, along a given direction.
This places our representation in contrast with recent work \cite{zobeidi2021deep,sitzmann2021light}, which does not model internal structure.
The second lies in the ease with which we can combine multiple DDFs, 
    which is useful for manipulation without retraining
    and
    scaling to more complex scenes.
Our approach is inspired by prior work on soft rendering \cite{liu2019soft,gao2020learning}.
Formally, given a set of $N$ DDFs 
    $ \zeta = \{ T^{(i)}, \xi^{(i)}, d^{(i)}, \mathcal{B}^{(i)} \}_{i=1}^N$,
    where $T^{(i)}$ is a transform on oriented points converting world to object coordinates for the $i$th DDF 
    (e.g., scale, rotation, and translation),
    we can aggregate the visibility and depth fields 
    into a single combined DDF.
For visibility of the combination of objects, we ask that \textit{at least} one surface is visible, implemented as:
    \begin{equation}
        \xi_\zeta(p,v) = 1 - \prod_k (1 - \xi^{(k)}(T^{(k)}(p,v))).
    \end{equation}
For depth, we want the closest visible surface to be the final output. 
One way to perform this is via a linear combination
    \begin{equation}
        d_\zeta(p,v) = 
        \sum_i a_\zeta^{(k)}(p,v)\, d^{(k)}(T^{(k)}(p,v)),
    \end{equation}
    where $a_\zeta^{(k)}$ are computed via visibility and distance:
    \begin{equation}
        a_\zeta(p,v) = \mathrm{Softmax}
                        \left(  \left\{
                            \frac{ \eta_T^{-1} \xi^{(k)}(T^{(k)}(p,v))}{ 
                            \varepsilon_s + d^{(k)}(T^{(k)}(p,v)) }
                            \right\}_k\,
                        \right),
    \end{equation}
    with temperature $\eta_T$ and maximum inverse depth scale $\varepsilon_s$ as hyper-parameters.
This upweights contributions when distance is small, but visibility is high.
We exhibit these capabilities in Fig.\ \ref{fig:simpleroom}, 
    which consists of two independently trained DDFs 
    (one fit to five planes, forming a simple room, and
        the other to the bunny mesh), %
    where we simulate a camera starting outside the scene and entering the room.

\newcommand{\ctwca}{0.09\textwidth}

\subsection{UDF Extraction}
\label{sec:app:udfextract}
\begin{figure}
    \centering
    \includegraphics[width=\linewidth]{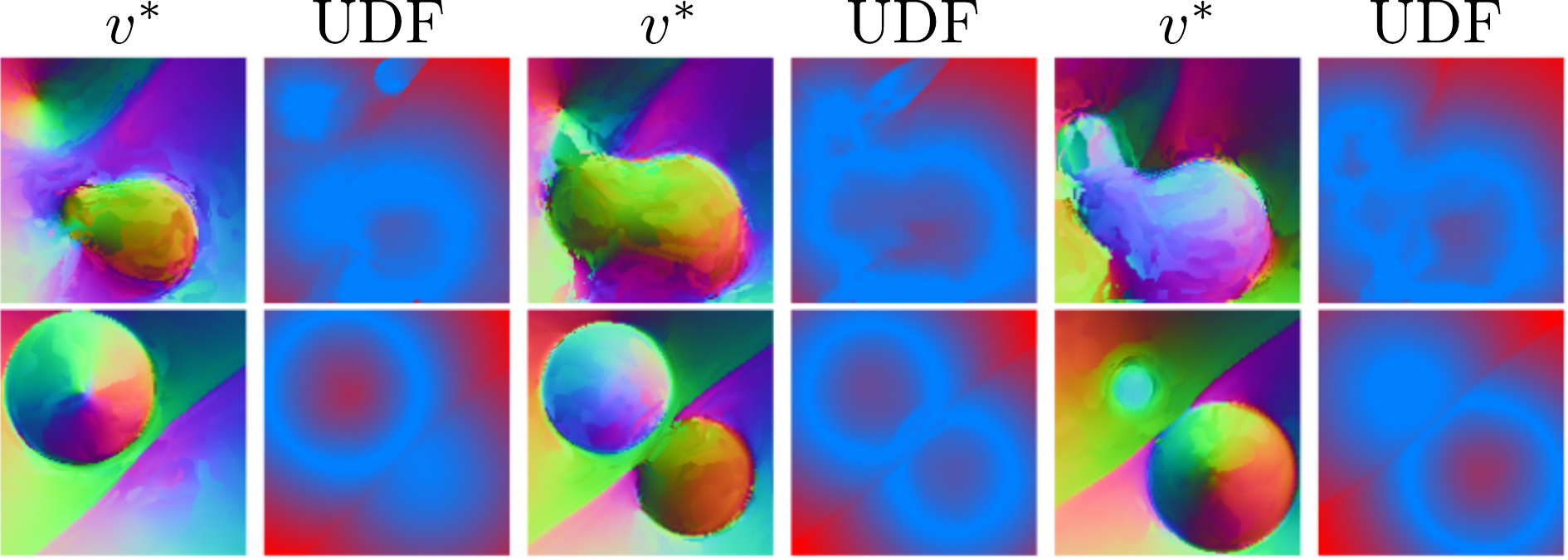}%
    \caption{
        $v^*$ fields (colours are 3D components) and respective UDFs (blue and red correspond to near and far distances).
        Each image is a slice in $z$, 
            with adjacent pairs having the same $z$.
        The differing colours in $v^*$ for the multisphere in
            column 3 are due to the slice breaching the front versus the back of the two spheres.
    }
    \label{fig:udf}
\end{figure}

As noted in Property \hyperref[property4]{IV},  
one can extract a UDF from a DDF.
In particular, we optimize a field 
$v^* : \mathcal{B} \rightarrow \mathbb{S}^2 $, 
such that $\mathrm{UDF}(p) = d(p, v^*(p))$.
We solve this by gradient descent on a loss that maximizes visibility while minimizing depth for a given $v^*(p)$.
(See Fig.\ \ref{fig:udf} for visualizations and Appendix \ref{appendix:udf} 
for optimization details.)

The vector field $v^*$ points in the direction of the closest point on $S$. 
Notice that discontinuities in $v^*$ occur at surfaces as before, but also on the medial surface of $S$ in $\mathcal{B}$.\footnote{At such positions, there are multiple valid values of $v^*$.}
When the surface normals exist, $v^*$ is closely related to them: 
$v^*(p) = -n(p,v^*(p))$, in the notation of Property \hyperref[property2]{II}.
Recent work has highlighted the utility of UDFs over SDFs %
\cite{venkatesh2020dude,chibane2020neural};
in the case of DDFs, 
extracting a UDF or $v^*$ may provide useful auxiliary information for some tasks.

\newcommand{\ctwcb}{0.075\textwidth} %

\begin{table*}[t] %
    \centering
    \begin{tabular}{cr|ccccc | cccc | c c}
         &  & \multicolumn{5}{c|}{DDF} & 
              \multicolumn{4}{c|}{PC-SIREN} & \multirow{2}{*}{P2M} & \multirow{2}{*}{3DR} \\
         &  & $\Pi_g$-L & $\Pi_g$-S & $\widehat{\Pi}_\nabla$-S & $\widehat{\Pi}$-L  & $\widehat{\Pi}$-S  
         & $\Pi_g$-L  & $\Pi_g$-S  & $\widehat{\Pi}$-L  & $\widehat{\Pi}$-S &  &  \\\hline 
    & $D_C$  $ \downarrow$ & 0.459 & 0.512 & 0.823 & 0.855  & 0.919 
                            & 0.431 & 0.465 & 0.876 & 0.915 
                            & 0.610 & 1.432 \\ 
Chairs & $F_{\tau}$ $ \uparrow$ & 55.47 & 48.40 & 42.28 & 47.51 & 41.08 
                             & 62.25 & 56.36 & 50.56 & 45.57 
                             & 54.38 & 40.22 \\ 
     & $F_{2\tau}$ $ \uparrow$ & 72.82 & 67.75 & 60.39 & 63.81 & 58.98 
                             & 77.38 & 74.56 & 65.43 & 62.76 
                             & 70.42 & 55.20 \\\hline 
    & $D_C$  $ \downarrow$ & 0.210 & 0.239 & 0.673 & 0.793 & 0.836 
                            & 0.215 & 0.227 & 0.829 & 0.844 
                            & 0.477 & 0.895 \\ 
Planes & $F_{\tau}$  $ \uparrow$ & 80.46 & 76.62 & 63.32 & 63.75 & 60.54 
                             & 81.49 & 80.11 & 63.64 & 62.34 
                             & 71.12 & 41.46 \\ 
     & $F_{2\tau}$  $ \uparrow$ & 90.05 & 88.55 & 76.16 & 74.96 & 73.47 
                             & 89.71 & 89.18 & 74.76 & 74.18 
                             & 81.38 & 63.23 \\ \hline
    & $D_C$  $ \downarrow$ & 0.231 & 0.288 & 0.390 & 0.541 & 0.606 
                            & 0.371 & 0.400 & 0.737 & 0.768 
                            & 0.268 & 0.845 \\ 
Cars & $F_{\tau}$  $ \uparrow$ & 70.91 & 59.93 & 54.16 & 62.69 & 52.47 
                             & 64.57 & 57.82 & 56.01 & 50.04
                             & 67.86 & 37.80 \\ 
     & $F_{2\tau}$  $ \uparrow$ & 86.57 & 79.66 & 74.68 & 79.71 & 72.78 
                             & 78.72 & 76.00 & 71.22 & 68.52 
                             & 84.15 & 54.84 \\ 
    \end{tabular}
    \caption{
        Single-image 3D reconstruction results.
        Rows: ShapeNet categories and performance metrics.
        Columns: %
        L/S refer to sampling 5000/2466 points for evaluation (2466 being the output size of P2M), $\Pi_g$/$\widehat{\Pi}$ denote using the true versus predicted camera for evaluation (the former case removing camera prediction error), and $\widehat{\Pi}_\nabla$ 
        test-time camera correction from the predicted position using gradient descent.
        Metrics: $D_C$ is the Chamfer distance ($\times 1000$), $F_{\tau}$ is the F-score ($\times 100$) at threshold $\tau = 10^{-4}$.
        PC-SIREN is our matched-architecture baseline;
        Pixel2Mesh (P2M) \cite{wang2018pixel2mesh,wang2020pixel2mesh} 
        and 
        3D-R2N2 (3DR) \cite{choy20163d} 
        are baselines using different shape modalities 
        (numbers from \cite{wang2020pixel2mesh}).
        Note that scenarios using $\Pi_g$ (effectively evaluating shapes in canonical object coordinates) are not directly comparable to P2M or 3DR.
    }
    \label{tab:si3dr}
\end{table*}

\subsection{Single-Image 3D Reconstruction}
\label{sec:app:si3dr}

We next utilize DDFs for single-image 3D reconstruction.
Given a colour image $I$, 
we predict the underlying latent shape $z_s$ 
and camera ${\Pi}$ that gave rise to the image, via an encoder $ E(I) = (\widehat{z}_s, \widehat{\Pi}) $. 
For decoding, we use a \textit{conditional} PDDF (CPDDF), which computes depth $\widehat{d}(p,v|z_s)$ and visibility $\widehat{\xi}(p,v|z_s)$. 
For evaluation,
we use the extrinsics of a camera: either the predicted $\widehat{\Pi}$ or ground-truth $\Pi_g$ (to separate shape and camera errors).

We use three loss terms: 
(a) shape DDF fitting in canonical pose $\mathfrak{L}_S$ (eq.\ \ref{eq:singlefitall}),
(b) camera prediction $\mathcal{L}_\Pi = ||\Pi_g - \widehat{\Pi}||_2^2$,
and 
(c) mask matching $ \mathcal{L}_M = \mathrm{BCE}(I_\alpha, \mathcal{R}_\xi(z_s|\widehat{\Pi})) $, where $I_\alpha$ is the input alpha channel and $ \mathcal{R}_\xi $ renders the DDF visibility.
The total objective, 
$\mathcal{L}_\mathrm{SI3DR} = \gamma_{R,S}\mathfrak{L}_S + \gamma_{R,\Pi}\mathcal{L}_\Pi  + \gamma_{R,M} \mathcal{L}_M $, is optimized by AdamW \cite{loshchilov2018decoupled}.
We implement $E$ as two ResNet-18 networks \cite{he2016deep}, while the CPDDF is a modulated SIREN \cite{mehta2021modulated}.
See Appendix \ref{appendix:si3dr} for details.

\textbf{Explicit Sampling.}  Evaluating 3D reconstruction often involves metrics based on point clouds (PCs). We present a simple approach to PC extraction from  DDFs, though it cannot guarantee uniform sampling over the shape. 
Analogous to prior work \cite{chibane2020neural}, we recall that $q(p,v) = p + d(p,v)v \in S$, if $\xi(p,v) = 1$. 
Thus, we sample $p\sim \mathcal{U}[\mathcal{B}]$, and wish to compute their projections $q$ onto the shape.
However, we cannot choose an arbitrary $v$, as many will not be visible.
Instead, for each $p$, we uniformly sample directions $V(p) = \{v_i(p)\sim\mathcal{U}[\mathbb{S}^2]\}_{i=1}^{n_v}$ and utilize our composition technique to estimate $\widehat{v}^{\,*}(p)$ by weighted average over $V(p)$ (as in \S\ref{sec:app:udfextract}, but without optimization), giving $q(p,\widehat{v}^{\,*}(p))$ as a point on the shape.
Repeating this process $N_H$ times (starting from $p\leftarrow q$) can also help, if depths are less accurate far from the shape.
We set $n_v=128$ and $N_H = 3$ (see Appendix \ref{appendix:si3dr:nh1} for ablation with $N_H=1$).

\textbf{Baselines.}  
Our primary baseline is designed to alter the \textit{shape representation}, while keeping the remaining architecture and training setup as similar as possible. We do this by using the same encoders as the DDF and an almost identical network for the decoder (changing only the input and output layer dimensionalities), but altered to output PCs directly (denoted PC-SIREN).
In particular, we treat the decoder as an implicit shape mapping $f_b : \mathbb{R}^3 \rightarrow \mathbb{R}^3$, which takes $p\sim\mathcal{U}[-1,1]^3$ as input and directly returns $q = f_b(p) \in S$ as output. 
Training uses the Chamfer distance $D_C$ to the ground truth 
in object coordinates and $\mathcal{L}_\Pi$.
We also consider two other baselines:
the mesh-based Pixel2Mesh (P2M) \cite{wang2018pixel2mesh,wang2020pixel2mesh} and voxel-based 3D-R2N2 (3DR) \cite{choy20163d}.

\newcommand{\alw}{0.274\textwidth}
\newcommand{\alh}{0.49in}
\newcommand{\aalh}{0.50in}
\newcommand{\cch}{0.30}
\newcommand{\ach}{0.01}
\newcommand{\cuthch}{0.11}
\begin{figure*}
    \adjincludegraphics[height=\aalh,trim={ {\ach\width} {\cuthch\height} {\ach\width}  {\cuthch\height}},clip]{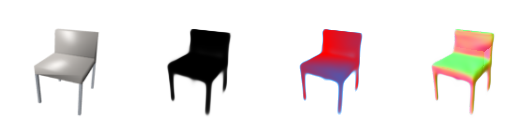}
    \adjincludegraphics[height=\alh,trim={ {\cch\width} {\cuthch\height} {\cch\width}  {\cuthch\height}},clip]{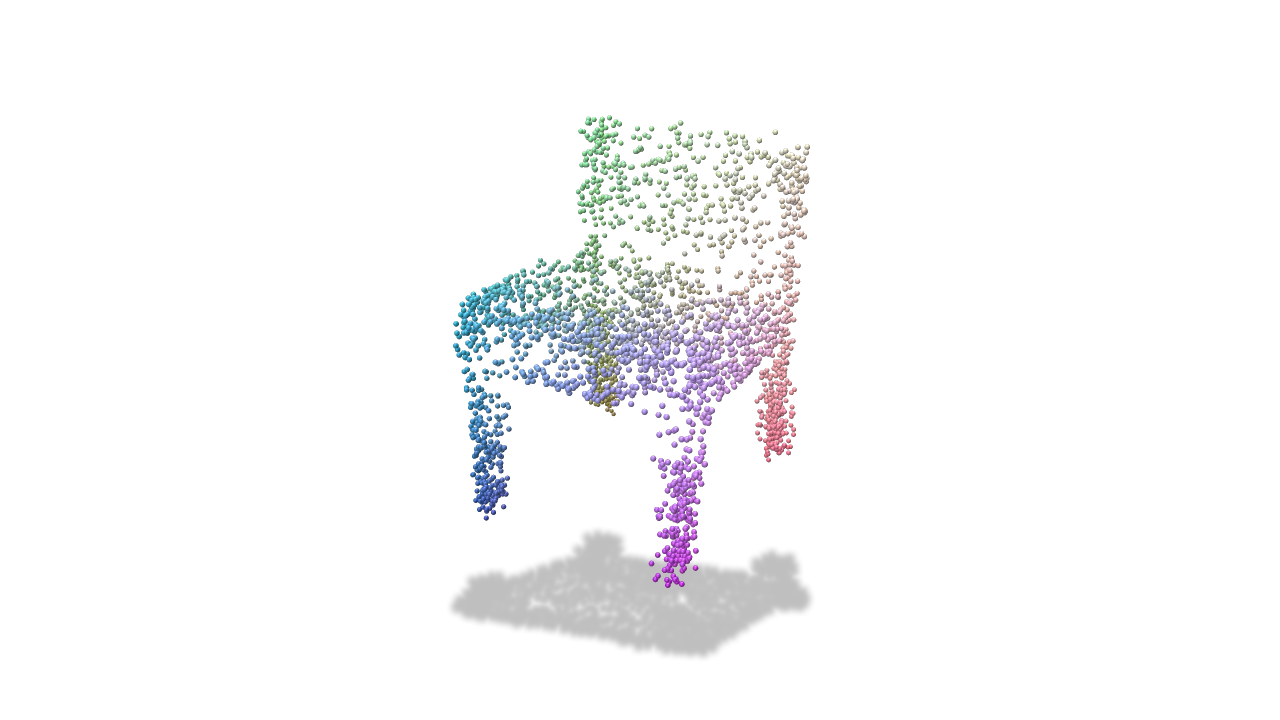}\hfill
    \adjincludegraphics[height=\alh,trim={ {\cch\width} {\cuthch\height} {\cch\width}  {\cuthch\height}},clip]{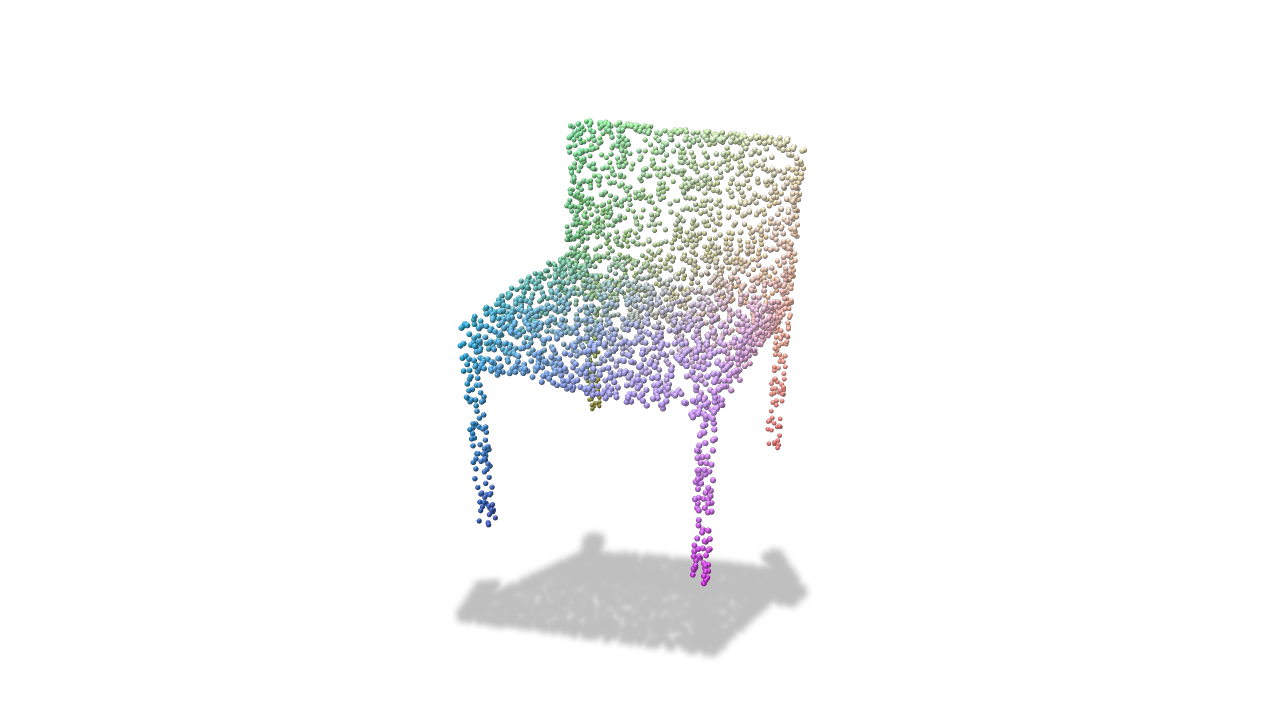}\hfill %
    \adjincludegraphics[height=\aalh,trim={ {\ach\width} {\cuthch\height} {\ach\width}  {\cuthch\height}},clip]{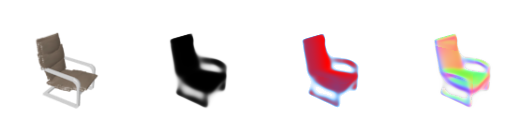}
    \adjincludegraphics[height=\alh,trim={ {\cch\width} {\cuthch\height} {\cch\width}  {\cuthch\height}},clip]{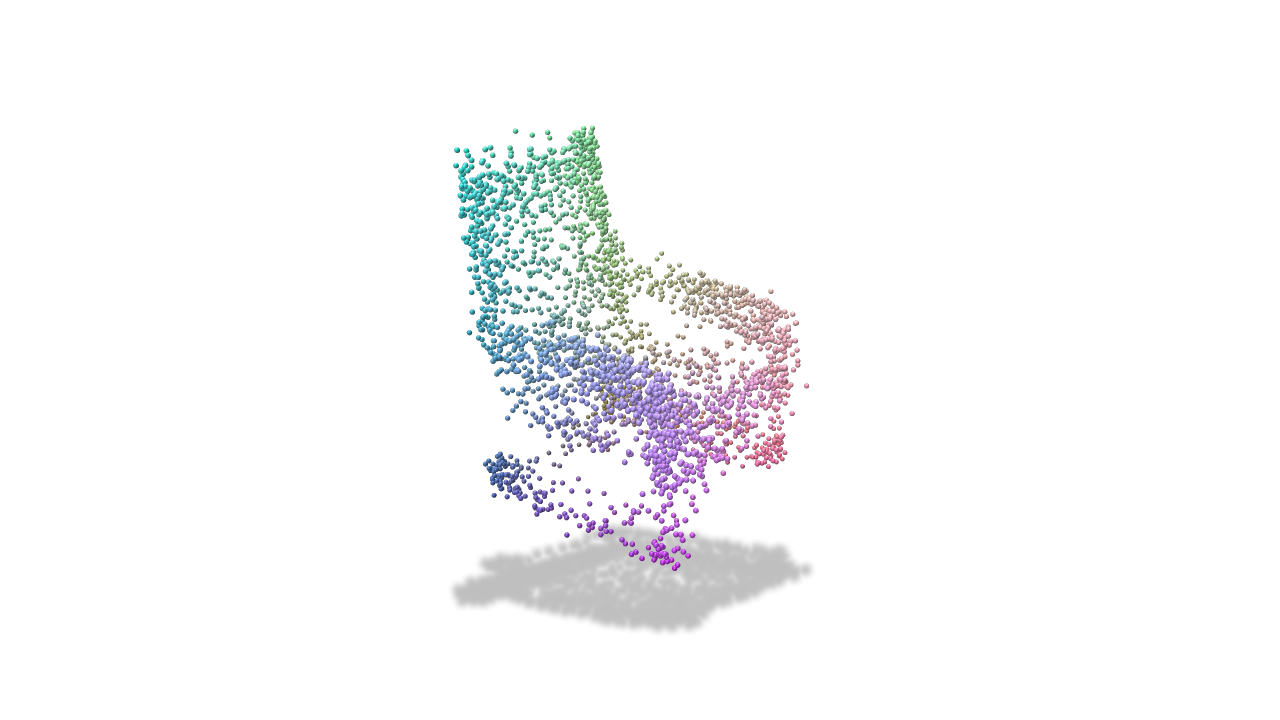}\hfill
    \adjincludegraphics[height=\alh,trim={ {\cch\width} {\cuthch\height} {\cch\width}  {\cuthch\height}},clip]{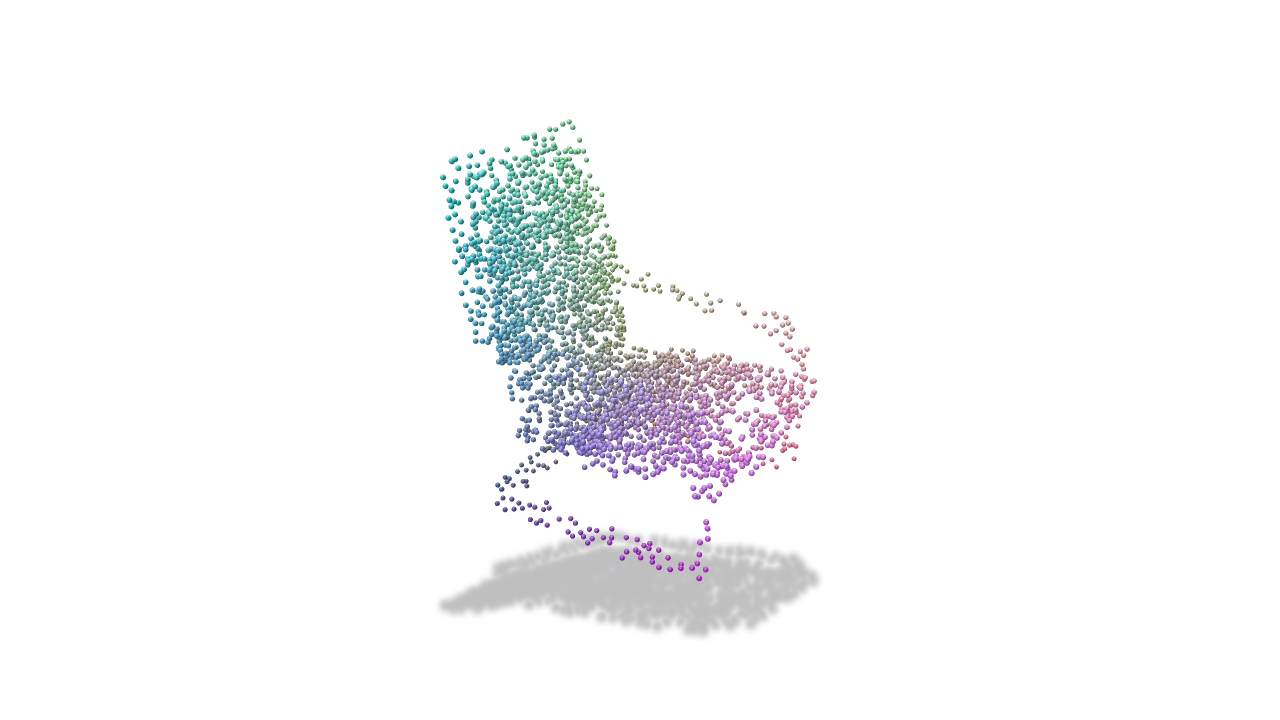} \\
    \adjincludegraphics[height=\aalh,trim={ {\ach\width} {\cuthch\height} {\ach\width}  {\cuthch\height}},clip]{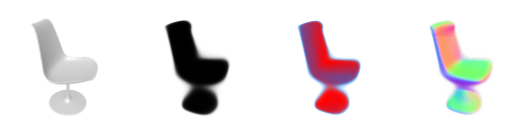}
    \adjincludegraphics[height=\alh,trim={ {\cch\width} {\cuthch\height} {\cch\width}  {\cuthch\height}},clip]{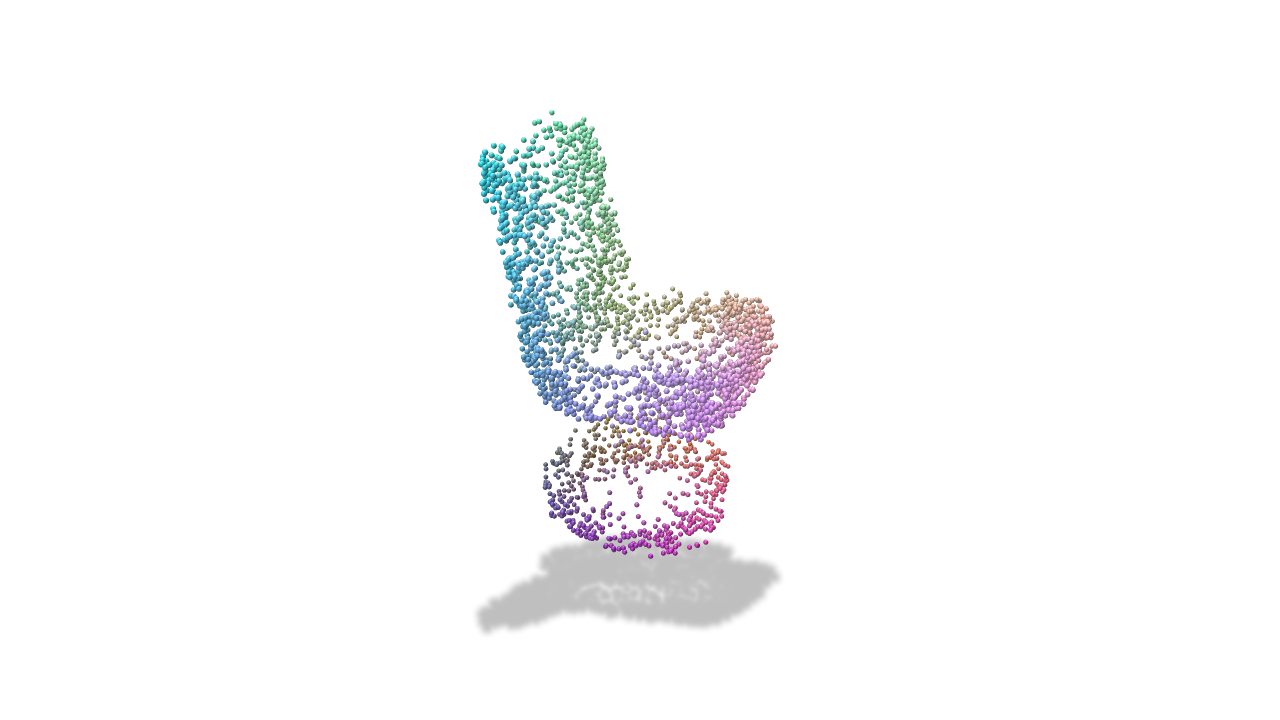}\hfill
    \adjincludegraphics[height=\alh,trim={ {\cch\width} {\cuthch\height} {\cch\width}  {\cuthch\height}},clip]{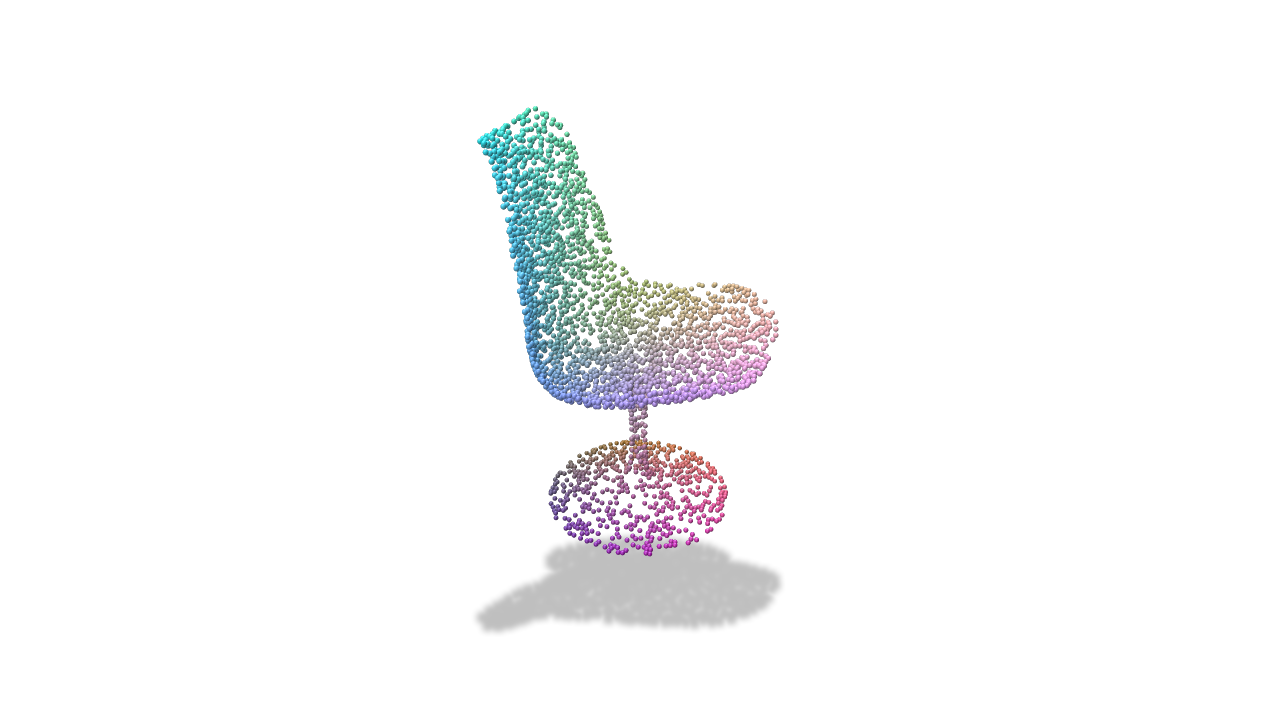}\hfill %
    \adjincludegraphics[height=\aalh,trim={ {\ach\width} {\cuthch\height} {\ach\width}  {\cuthch\height}},clip]{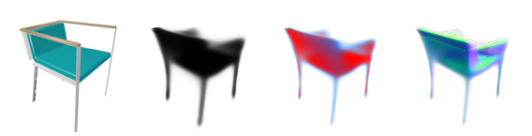}
    \adjincludegraphics[height=\alh,trim={ {\cch\width} {\cuthch\height} {\cch\width}  {\cuthch\height}},clip]{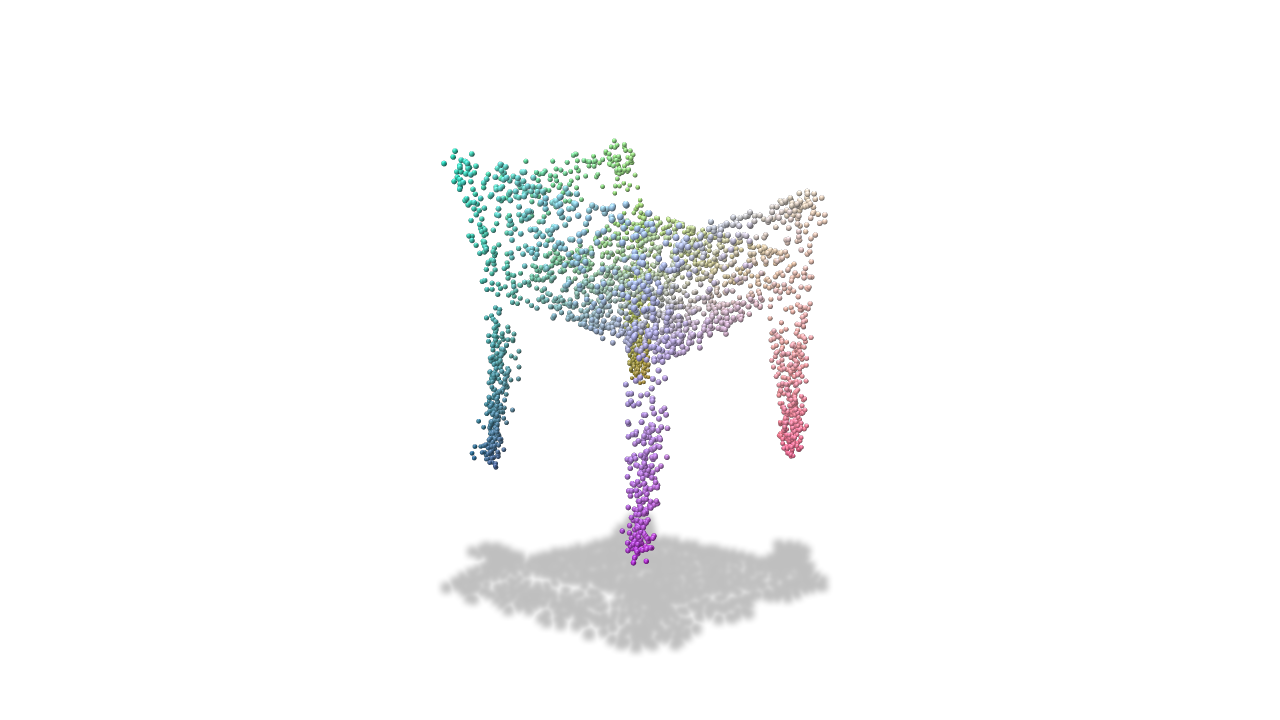}\hfill
    \adjincludegraphics[height=\alh,trim={ {\cch\width} {\cuthch\height} {\cch\width}  {\cuthch\height}},clip]{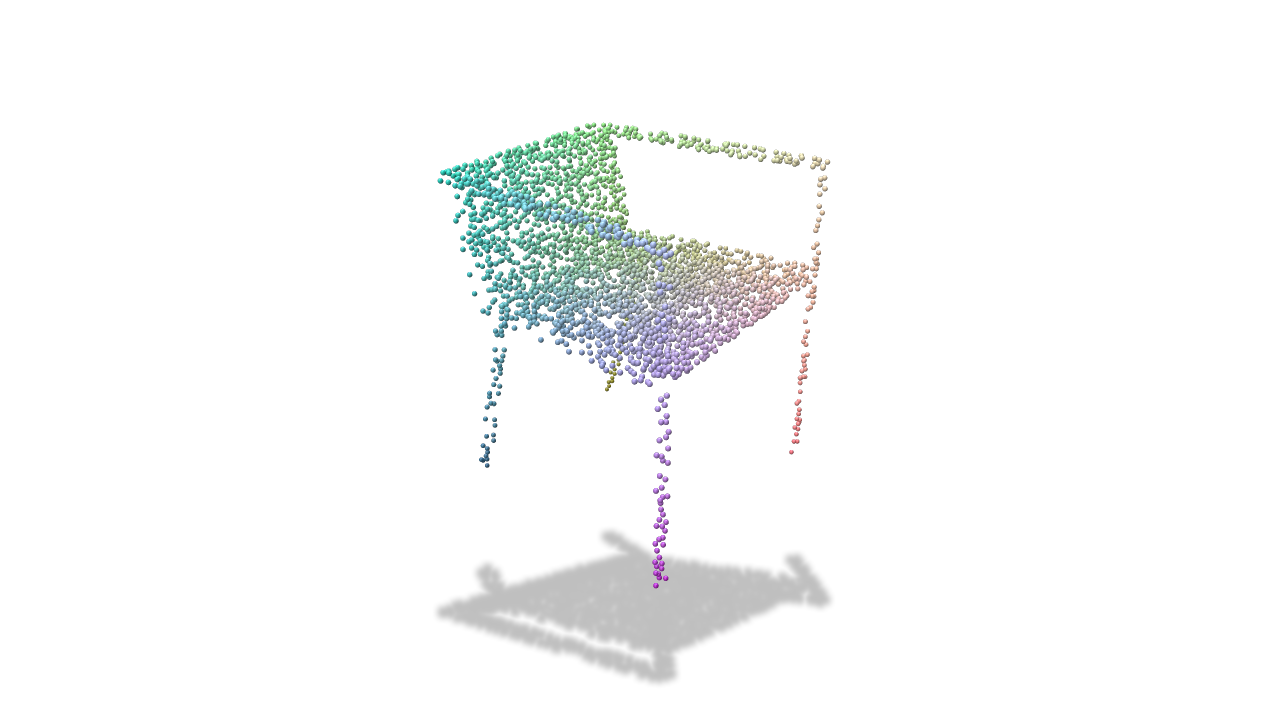} \\
    \adjincludegraphics[height=\aalh,trim={ {\ach\width} {\cuthch\height} {\ach\width}  {\cuthch\height}},clip]{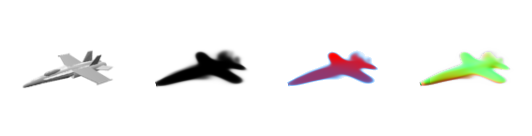}
    \adjincludegraphics[height=\alh,trim={ {\cch\width} {\cuthch\height} {\cch\width}  {\cuthch\height}},clip]{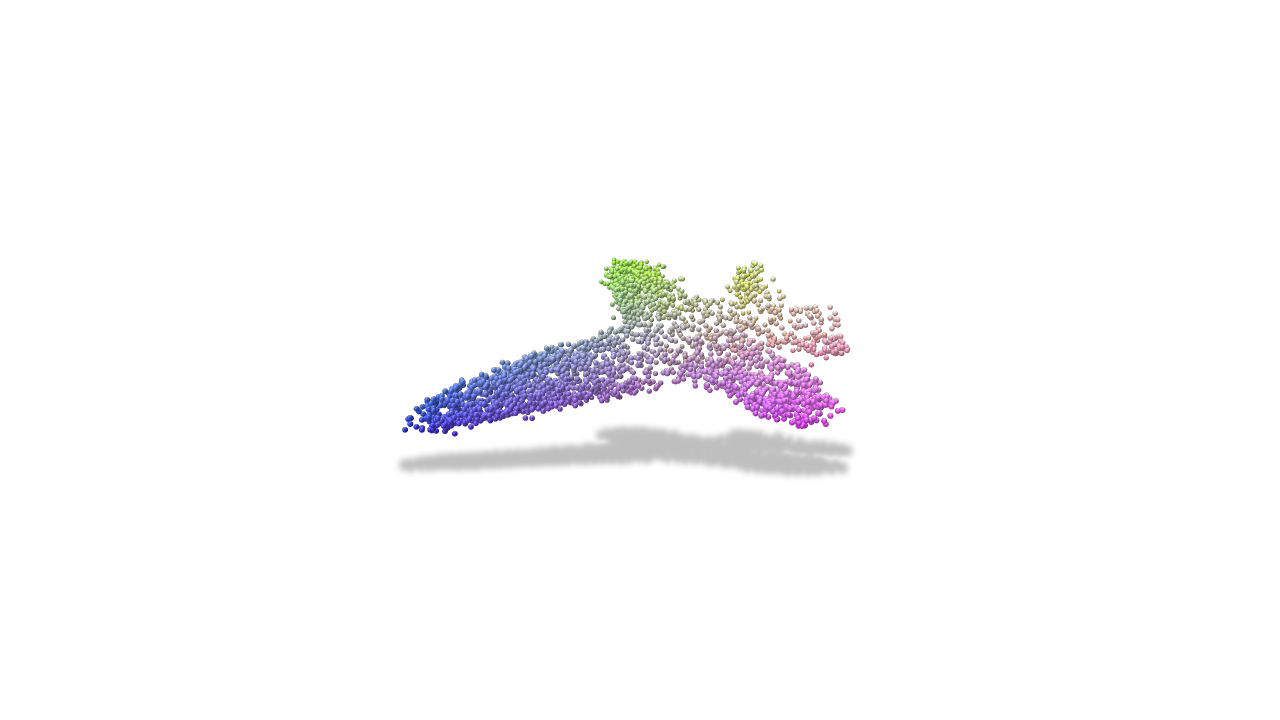}\hfill
    \adjincludegraphics[height=\alh,trim={ {\cch\width} {\cuthch\height} {\cch\width}  {\cuthch\height}},clip]{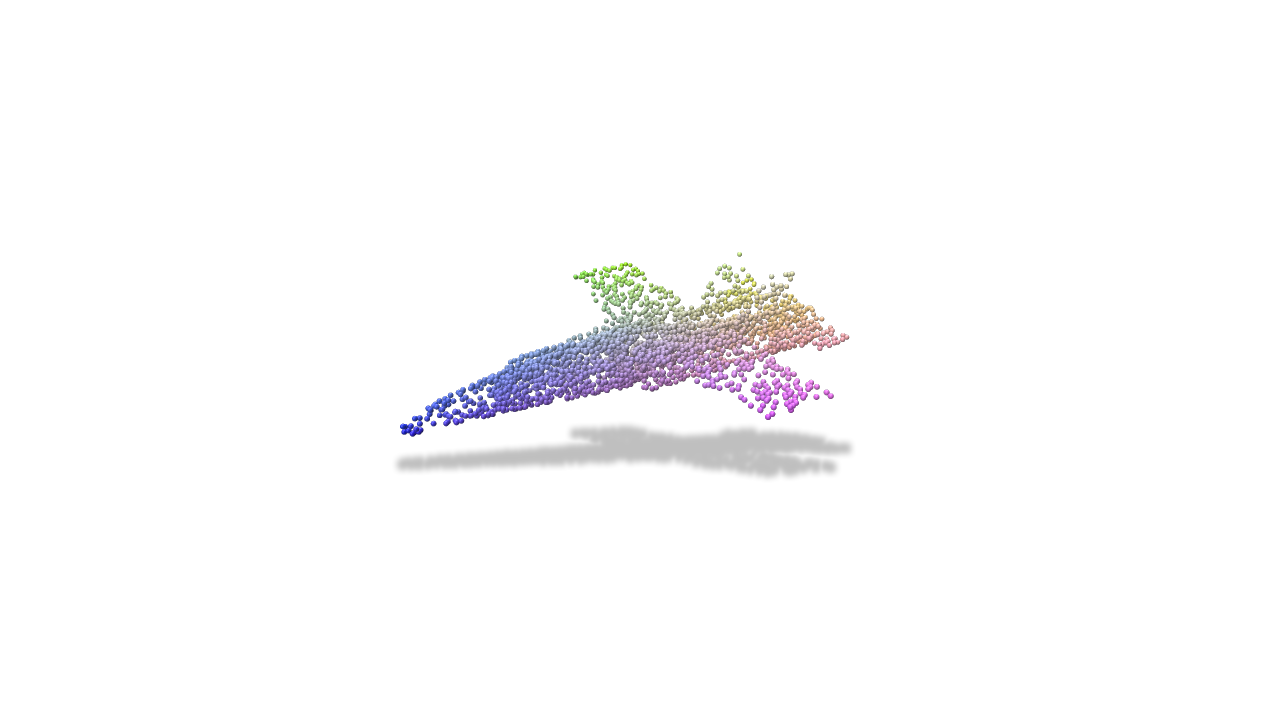}\hfill %
    \adjincludegraphics[height=\aalh,trim={ {\ach\width} {\cuthch\height} {\ach\width}  {\cuthch\height}},clip]{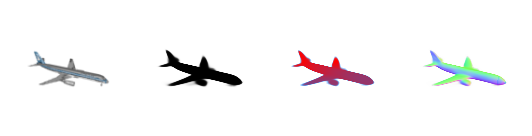}
    \adjincludegraphics[height=\alh,trim={ {\cch\width} {\cuthch\height} {\cch\width}  {\cuthch\height}},clip]{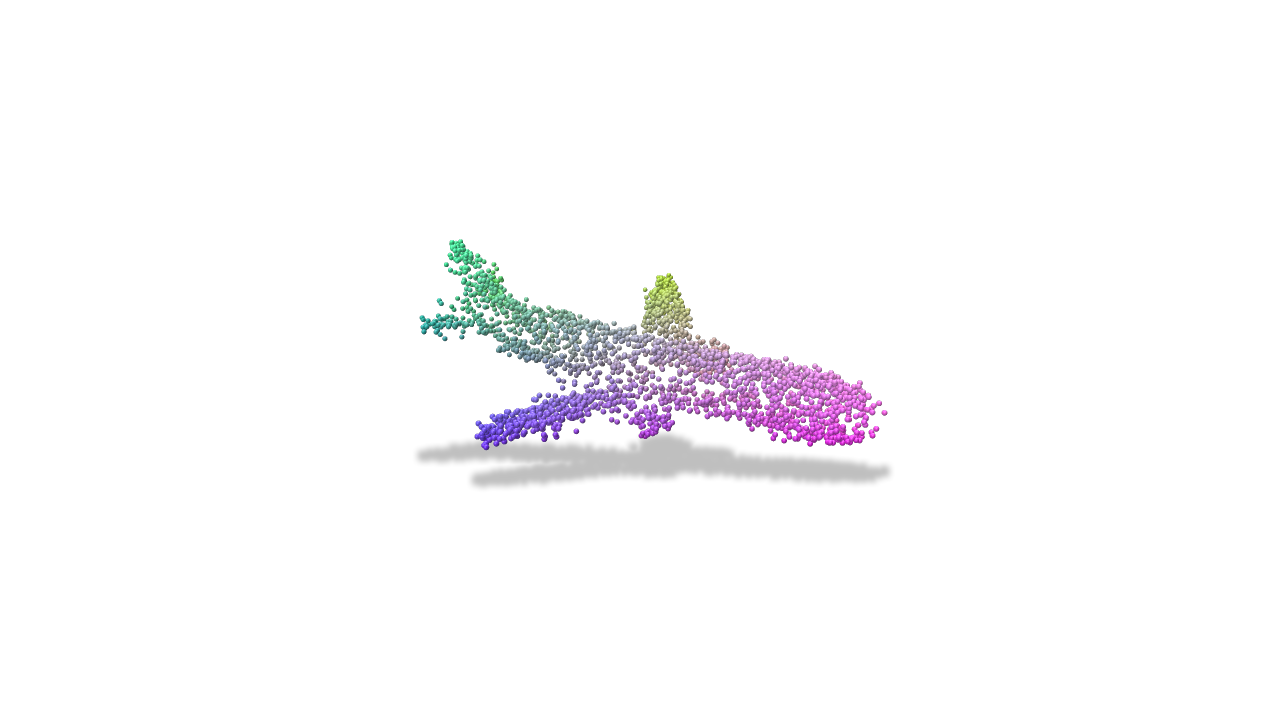}\hfill
    \adjincludegraphics[height=\alh,trim={ {\cch\width} {\cuthch\height} {\cch\width}  {\cuthch\height}},clip]{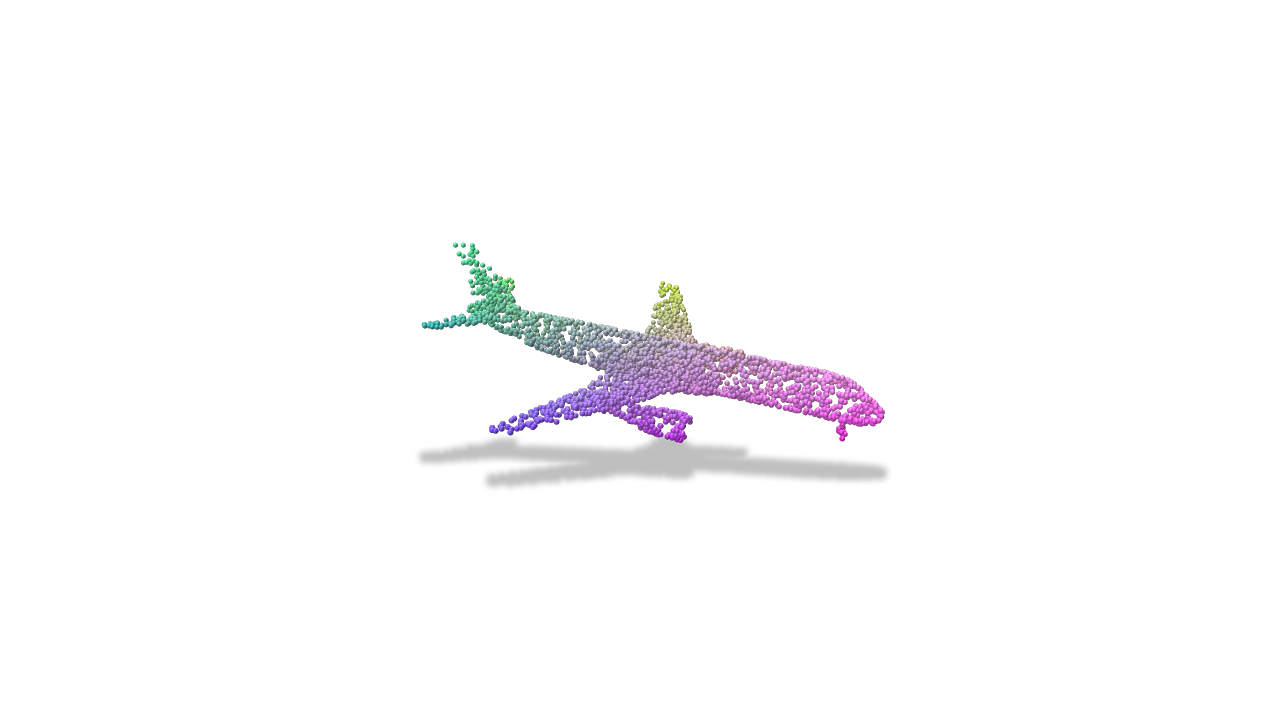} \\
    \adjincludegraphics[height=\aalh,trim={ {\ach\width} {\cuthch\height} {\ach\width}  {\cuthch\height}},clip]{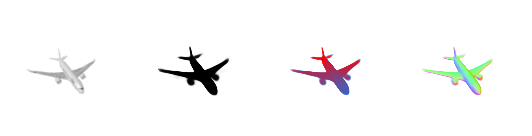}
    \adjincludegraphics[height=\alh,trim={ {\cch\width} {\cuthch\height} {\cch\width}  {\cuthch\height}},clip]{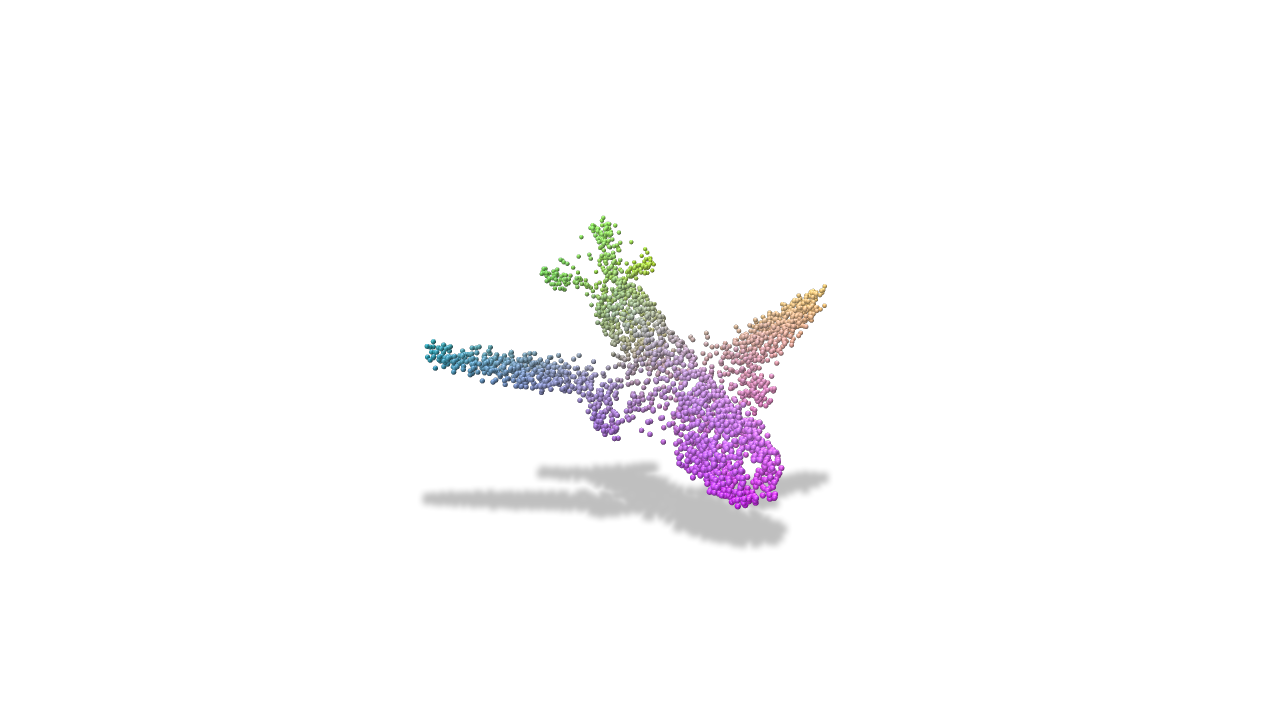}\hfill
    \adjincludegraphics[height=\alh,trim={ {\cch\width} {\cuthch\height} {\cch\width}  {\cuthch\height}},clip]{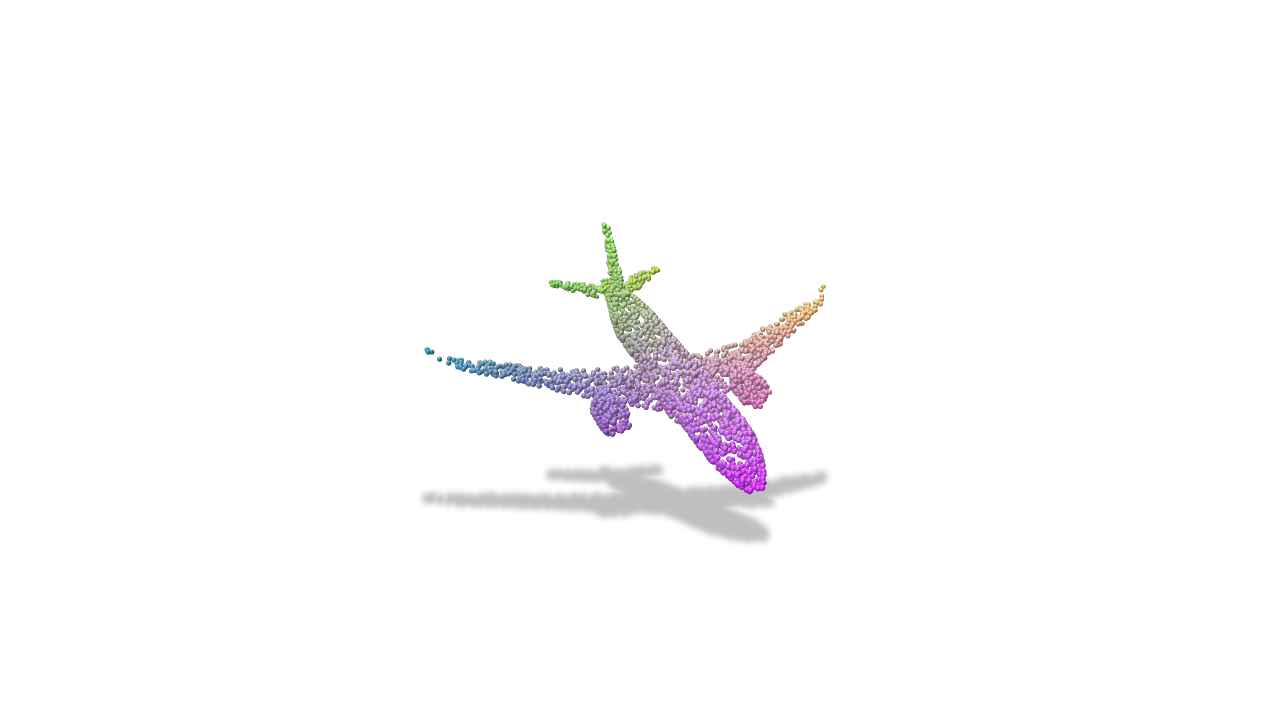}\hfill %
    \adjincludegraphics[height=\aalh,trim={ {\ach\width} {\cuthch\height} {\ach\width}  {\cuthch\height}},clip]{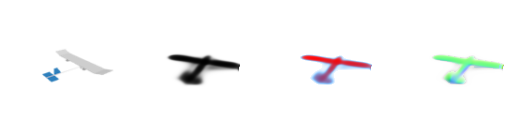}
    \adjincludegraphics[height=\alh,trim={ {\cch\width} {\cuthch\height} {\cch\width}  {\cuthch\height}},clip]{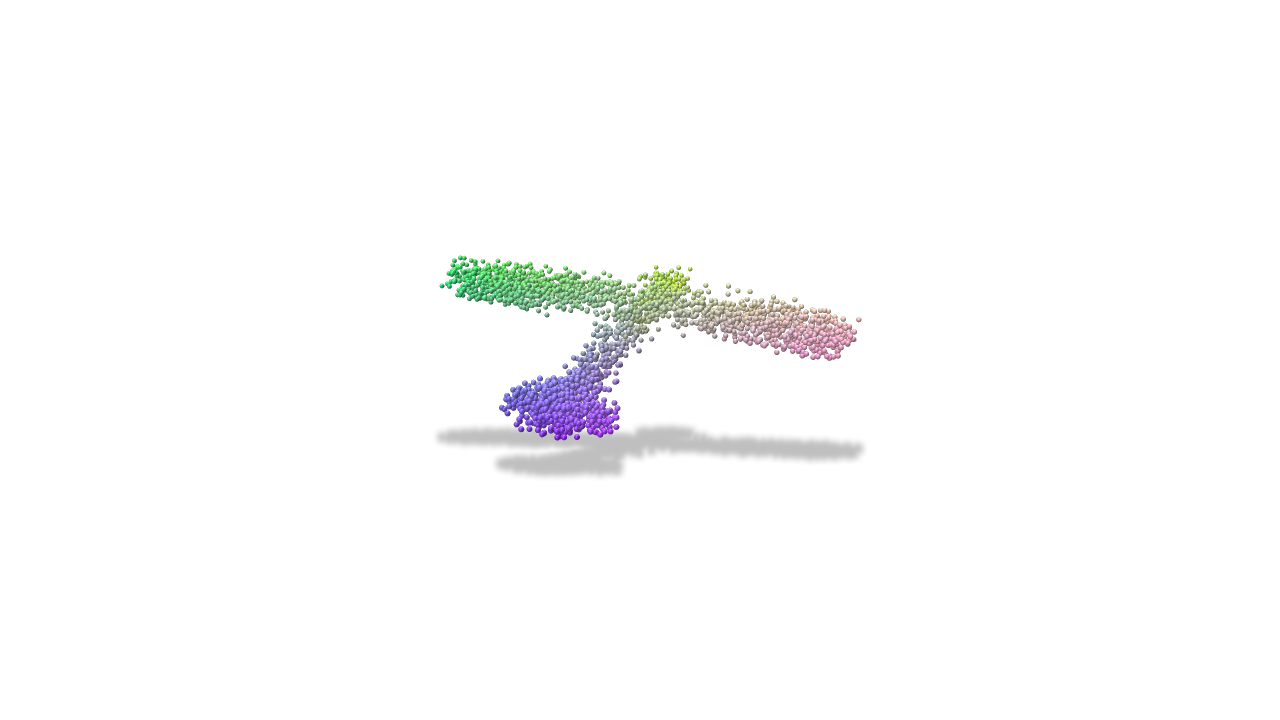}\hfill
    \adjincludegraphics[height=\alh,trim={ {\cch\width} {\cuthch\height} {\cch\width}  {\cuthch\height}},clip]{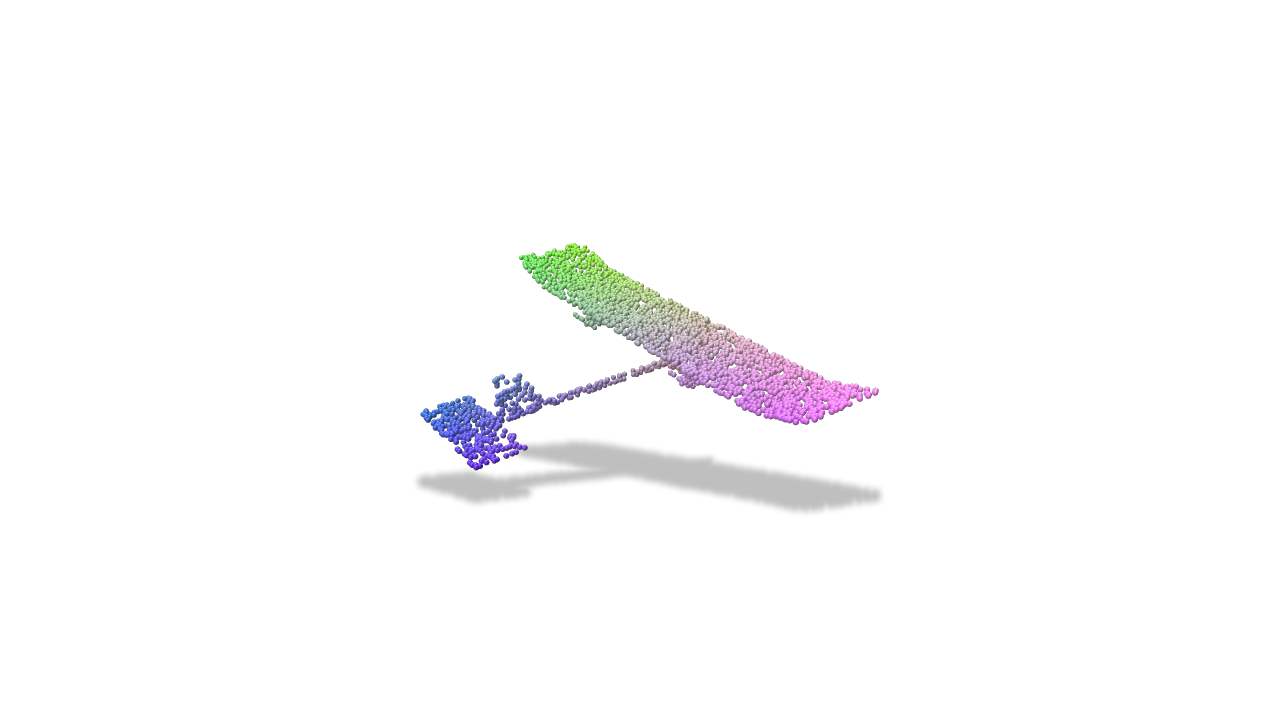} 
    \adjincludegraphics[height=\aalh,trim={ {\ach\width} {\cuthch\height} {\ach\width}  {\cuthch\height}},clip]{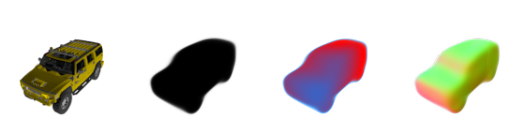}
    \adjincludegraphics[height=\alh,trim={ {\cch\width} {\cuthch\height} {\cch\width}  {\cuthch\height}},clip]{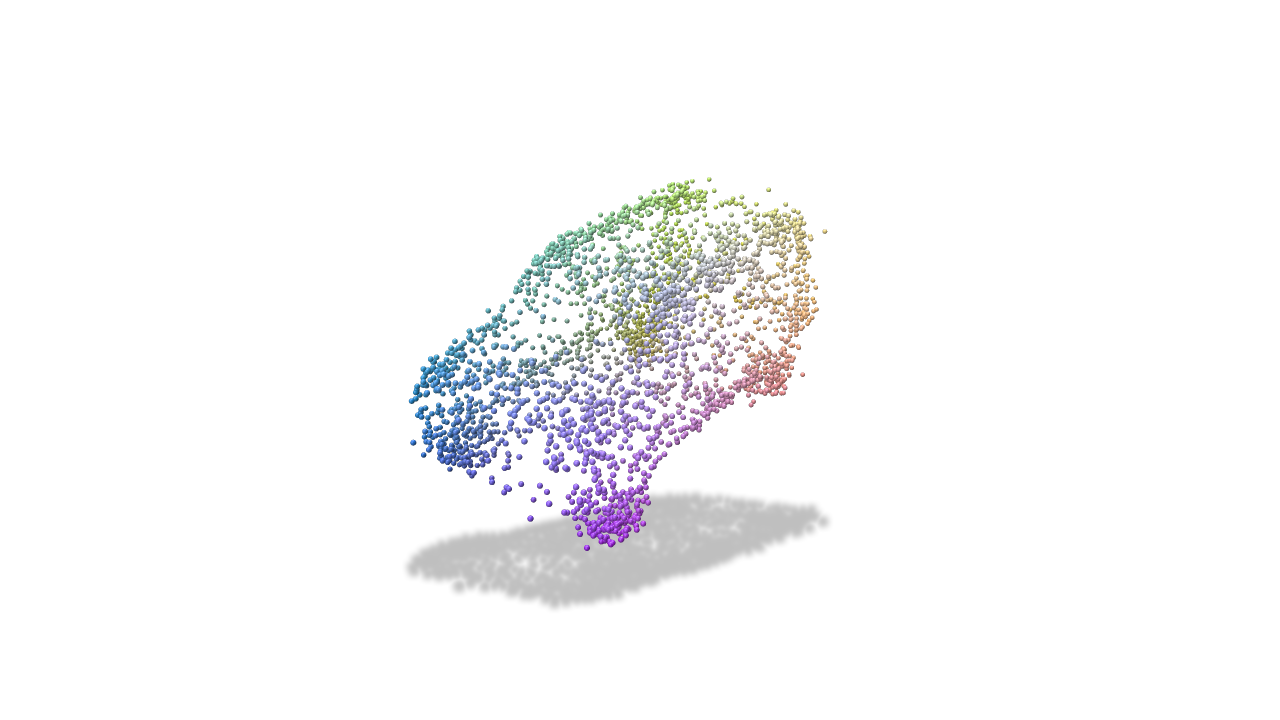}\hfill
    \adjincludegraphics[height=\alh,trim={ {\cch\width} {\cuthch\height} {\cch\width}  {\cuthch\height}},clip]{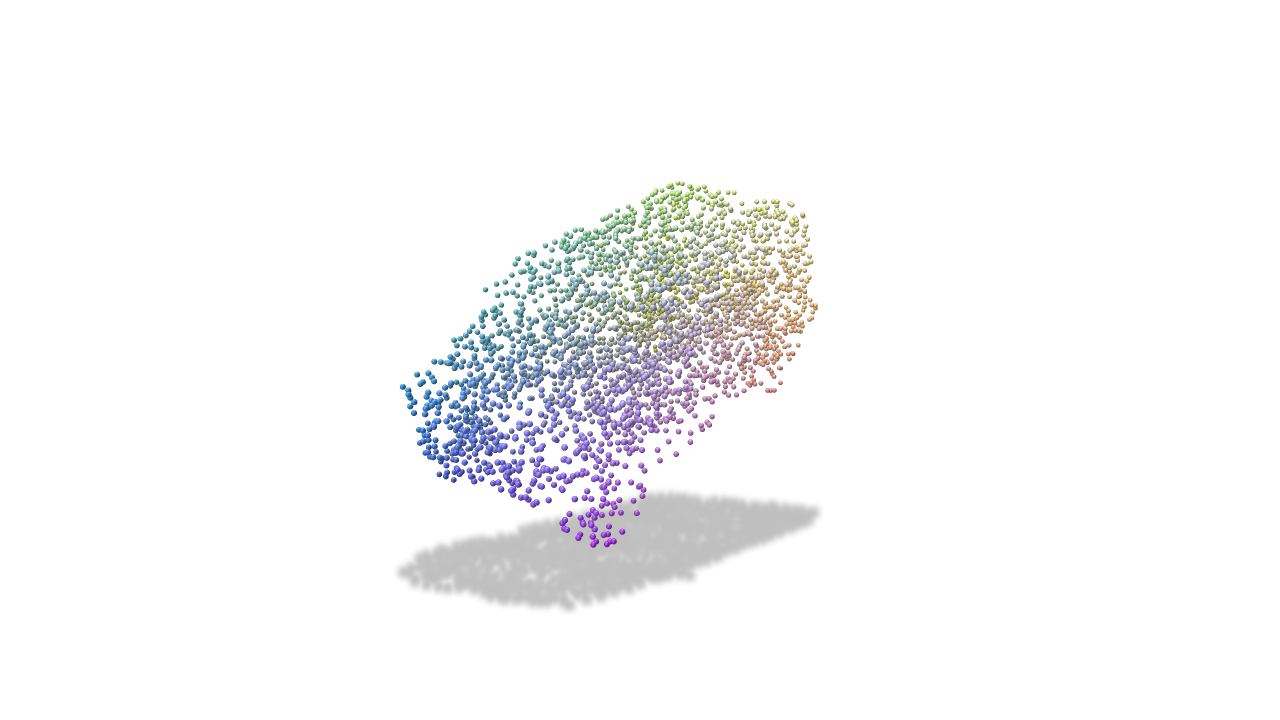}\hfill %
    \adjincludegraphics[height=\aalh,trim={ {\ach\width} {\cuthch\height} {\ach\width}  {\cuthch\height}},clip]{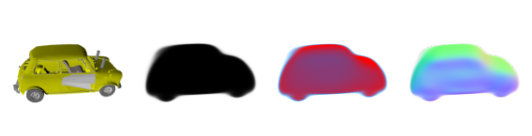}
    \adjincludegraphics[height=\alh,trim={ {\cch\width} {\cuthch\height} {\cch\width}  {\cuthch\height}},clip]{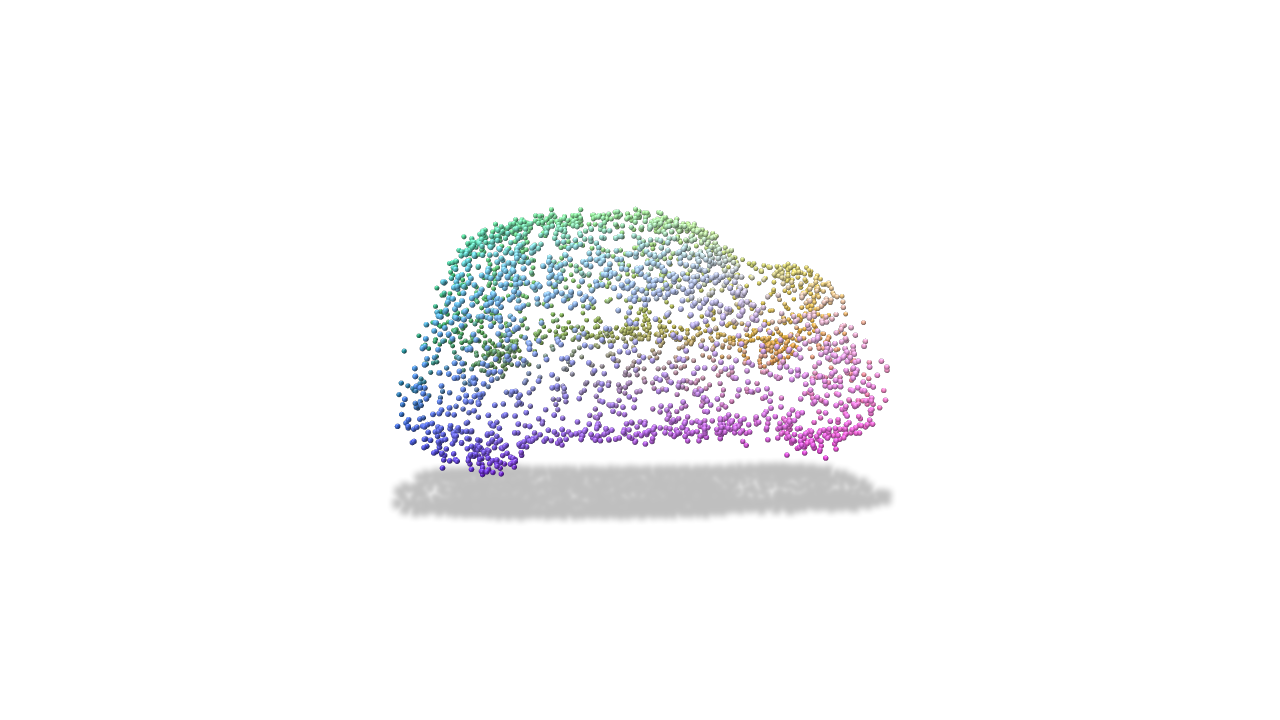}\hfill
    \adjincludegraphics[height=\alh,trim={ {\cch\width} {\cuthch\height} {\cch\width}  {\cuthch\height}},clip]{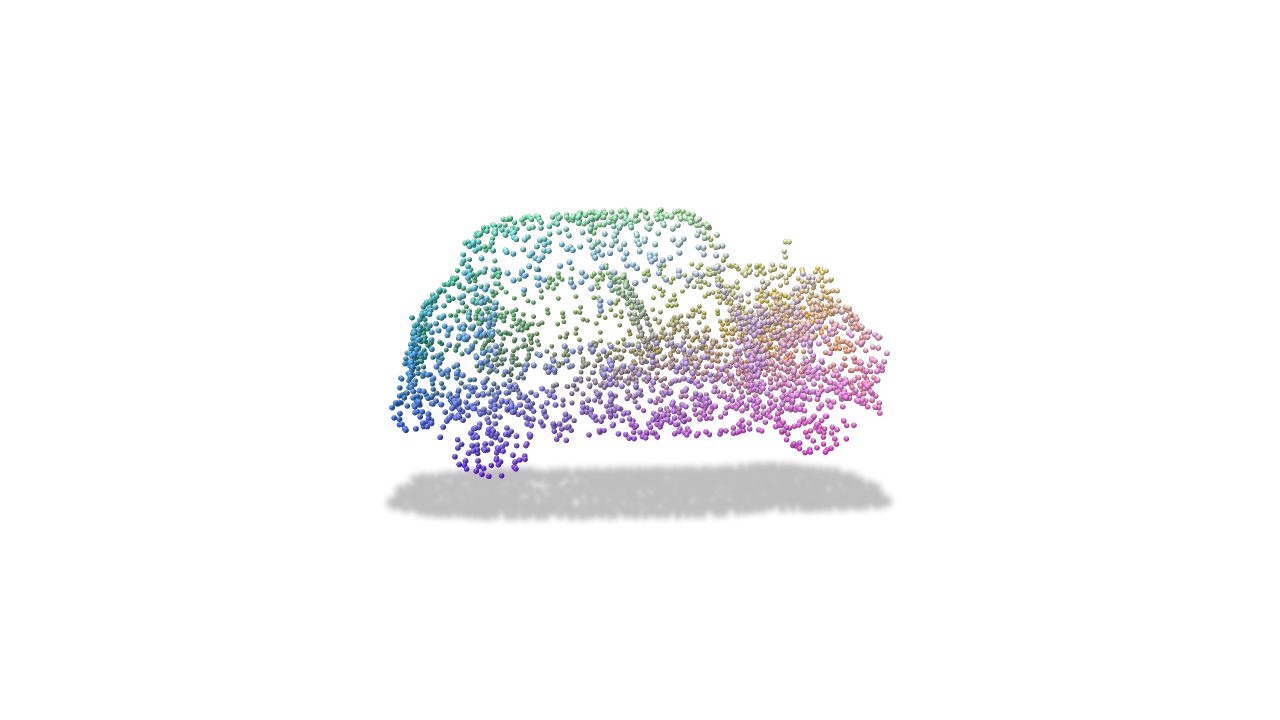} \\
    \adjincludegraphics[height=\aalh,trim={ {\ach\width} {\cuthch\height} {\ach\width}  {\cuthch\height}},clip]{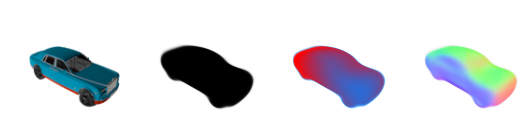}
    \adjincludegraphics[height=\alh,trim={ {\cch\width} {\cuthch\height} {\cch\width}  {\cuthch\height}},clip]{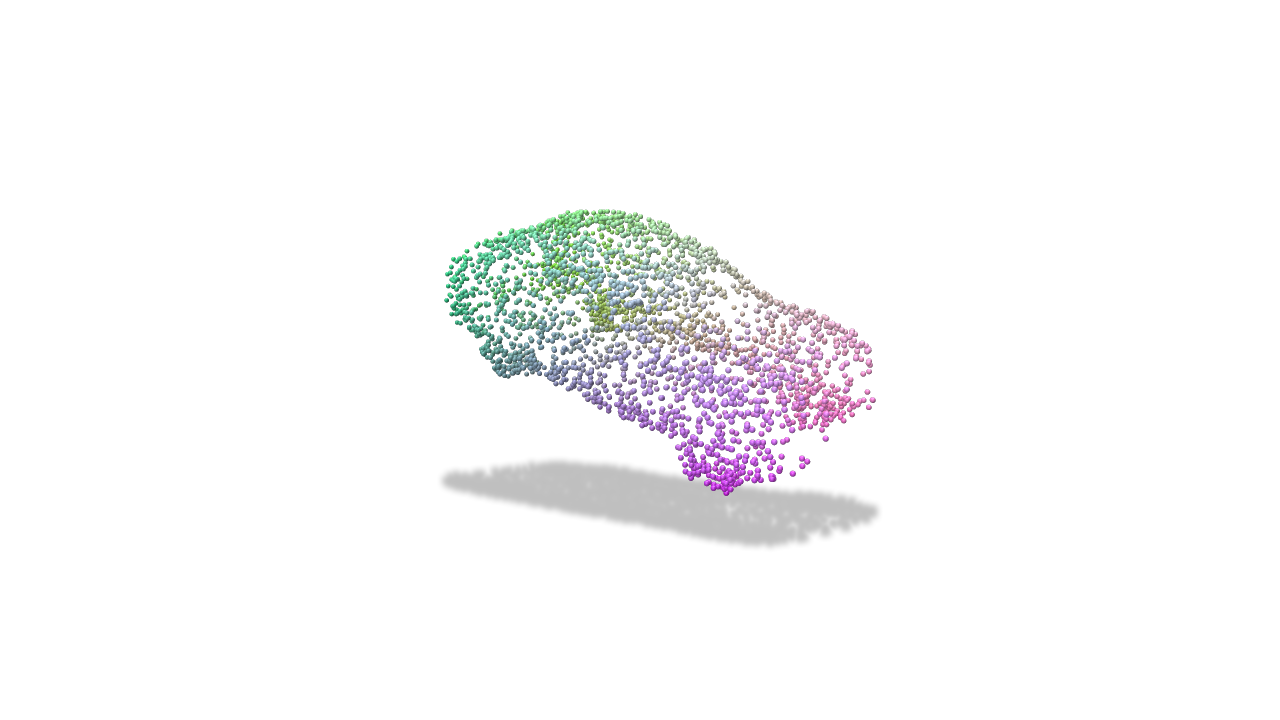}\hfill
    \adjincludegraphics[height=\alh,trim={ {\cch\width} {\cuthch\height} {\cch\width}  {\cuthch\height}},clip]{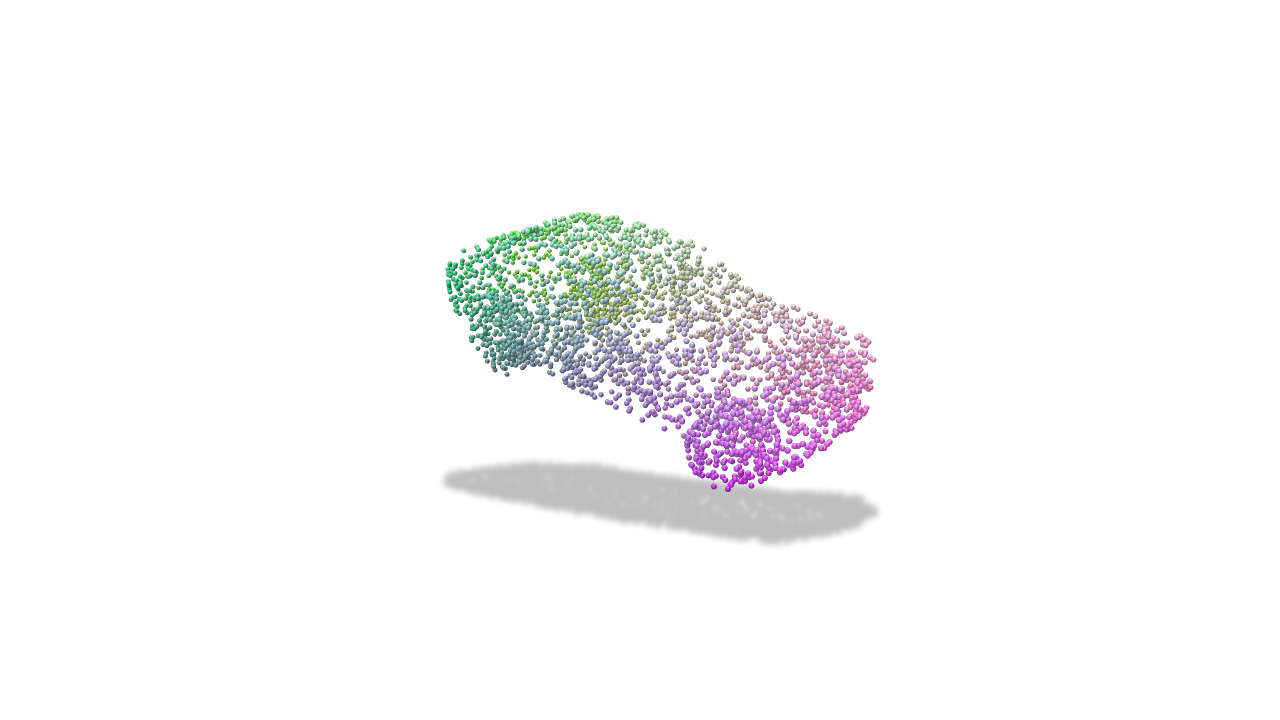}\hfill %
    \adjincludegraphics[height=\aalh,trim={ {\ach\width} {\cuthch\height} {\ach\width}  {\cuthch\height}},clip]{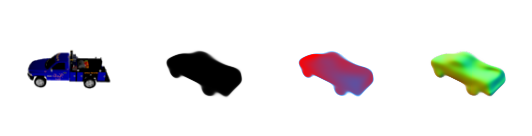}
    \adjincludegraphics[height=\alh,trim={ {\cch\width} {\cuthch\height} {\cch\width}  {\cuthch\height}},clip]{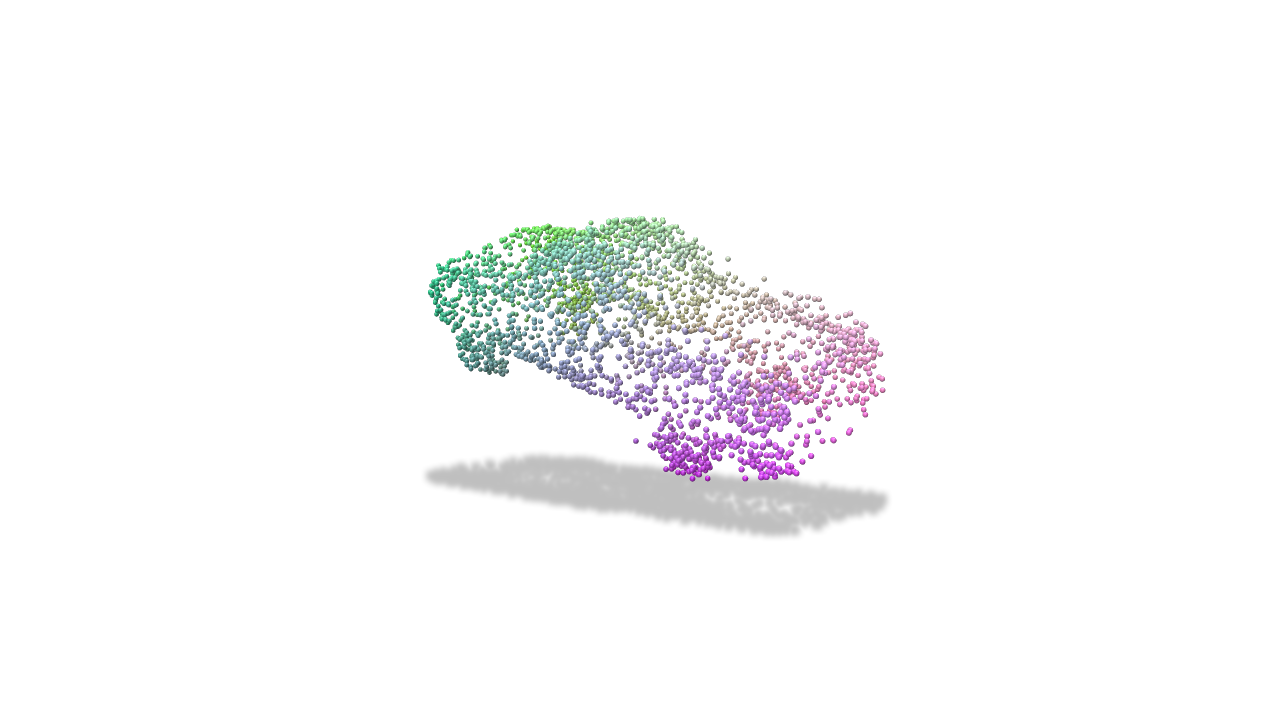}\hfill
    \adjincludegraphics[height=\alh,trim={ {\cch\width} {\cuthch\height} {\cch\width}  {\cuthch\height}},clip]{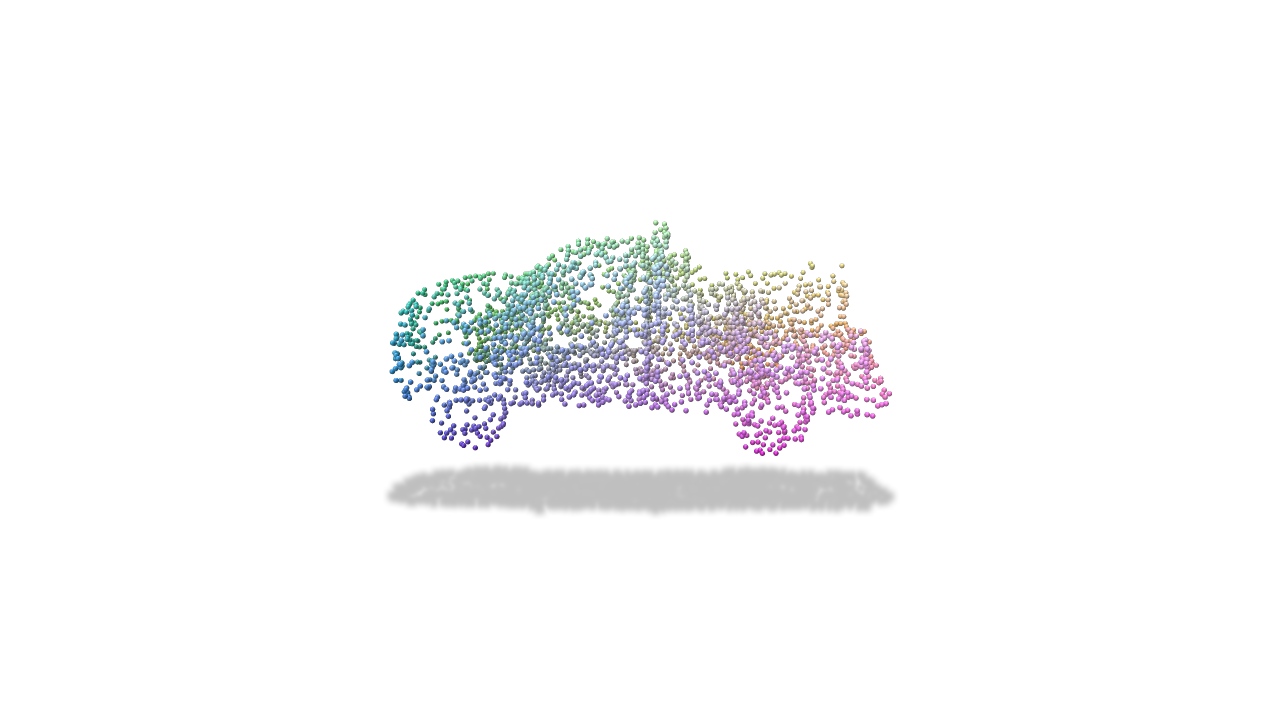}
    \caption{
    Single-image 3D reconstruction visualizations on held-out test data. 
    Per inset, columns represent 
    (i) the input RGB image, 
    (ii) the visibility $\widehat{\xi}$, 
    (iii) the depth $\widehat{d}$, 
    (iv) the normals $\widehat{n}$, 
    (v) the sampled point cloud (PC) from the DDF, and 
    (vi) a sample from the ground-truth PC.
    Quantities (ii-v) are all differentiably computed directly from the CPDDF and $\widehat{\Pi}$, per point or pixel (i.e., no post-processing needed).
    PC colours denote 3D coordinates.
    A high-error example is in the lower-right of each category.
    See Appendix \ref{appendix:si3dr} for more examples.
    }
    \label{fig:si3drvis}
\end{figure*}

\textbf{Results.}
We consider cars, planes, and chairs from ShapeNet \cite{shapenet2015}, using the data from %
\cite{choy20163d} (as in \cite{wang2018pixel2mesh}).
In Table \ref{tab:si3dr},
we show DDFs perform comparably to the architecture-matched PC-SIREN baseline (a little worse on chairs, but slightly better on cars).
See Fig.\ \ref{fig:si3drvis} for visualizations, as well as Appendix Fig.\ \ref{app:fig:si3drvis}.
Generally, the inferred DDF shapes correctly reconstruct most inputs, including thin structures like chair legs, and regardless of topology.
The most obvious errors are in pose estimation, but
the DDF can also sometimes output ``blurry'' shape parts when it is uncertain (e.g., for shapes far from the majority of training examples).
    
However, results can be improved 
(especially for $D_C$)
by correcting $\widehat{\Pi}$ to $\widehat{\Pi}_\nabla = \argmin_{\Pi} \mathcal{L}_M $, via gradient descent 
(starting from $\widehat{\Pi}$)
on the test image alpha channel.
While explicit modalities, like PCs, 
can be differentiably rendered 
(e.g., \cite{kato2018neural,tulsiani2017multi,eldar}), 
DDFs can do so by construction,
without additional heuristics or learning.
Further, note that %
(i) the DDF sampling procedure is not learned,
(ii) our model is not trained with $D_C$ (on which it is evaluated),
and 
(iii) DDFs are a richer representation than PCs, 
capable of representing higher-order geometry, built-in rendering, and even PC extraction.
Thus, for reconstruction, changing from PCs to DDFs can enrich the representation without quality loss.

Compared to the other baselines,
DDFs with predicted $\widehat{\Pi}$ underperform P2M, but outperform 3DR.
Results obtained with the ground-truth $\Pi_g$ indicate that much of this error is due to (imperfect) camera prediction, 
though this case is not directly comparable to P2M or 3DR. %
Indeed, with $\Pi_g$, 
the task becomes prediction in canonical object -- rather than camera -- coordinates.
While each frame %
has benefits and downsides \cite{shin2018pixels,tatarchenko2019single},
in our case it is useful to 
separate shape vs.\ camera error.
Our scores with $\Pi_g$ suggest DDFs can infer shape at similar quality levels to existing work, despite the naive architecture and sampling strategy. 
We remark that we do not expect DDFs to directly compare to specialized, highly tuned models achieving state-of-the-art performance. 
Instead, we show that our {representation} 
allows one to achieve good performance (especially when considering shape alone), 
even using simple off-the-shelf components (ResNets and SIREN MLPs), without sacrificing versatility. %

\begin{figure*}[ht] %
    \centering
    \begin{minipage}{0.50\textwidth}
    \adjincludegraphics[width=0.99\textwidth,trim={{.14\width} {.09\height} {.14\width}  {.09\height}},clip]{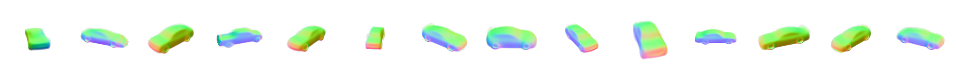}
    \adjincludegraphics[width=0.99\textwidth,trim={ {.14\width} {.09\height} {.14\width}  {.09\height}},clip]{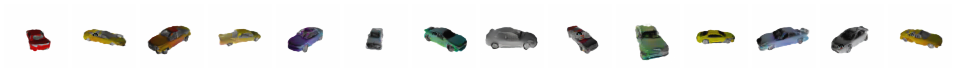}
    \adjincludegraphics[width=0.99\textwidth,trim={ {.07\width} {.09\height} {.21\width}  {.09\height}},clip]{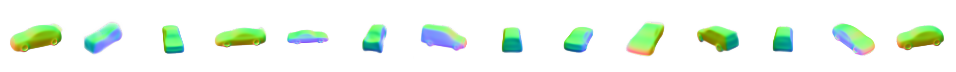}
    \adjincludegraphics[width=0.99\textwidth,trim={ {.07\width} {.09\height} {.21\width}  {.09\height}},clip]{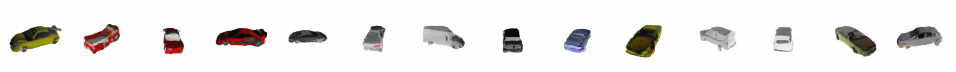}
    \end{minipage}\hfill\vline\hfill
    \begin{minipage}{0.30\textwidth}
    \adjincludegraphics[width=0.99\textwidth,trim={ {.04\width} {.09\height} {.02\width}  {.09\height}},clip]{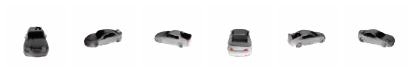}
    \adjincludegraphics[width=0.99\textwidth,trim={ {.04\width} {.09\height} {.02\width}  {.09\height}},clip]{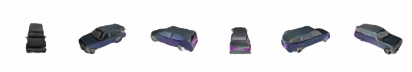}
    \adjincludegraphics[width=0.99\textwidth,trim={ {.04\width} {.09\height} {.02\width}  {.09\height}},clip]{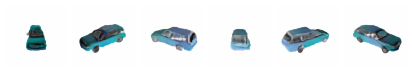}
    \adjincludegraphics[width=0.99\textwidth,trim={ {.04\width} {.09\height} {.02\width}  {.09\height}},clip]{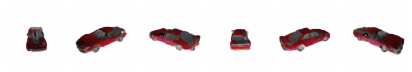}
    \end{minipage}\hfill\vline\hfill
    \begin{minipage}{0.19\textwidth}
    \adjincludegraphics[width=0.99\textwidth,trim={ {.0\width} {.08\height} {.0\width}  {.08\height}},clip]{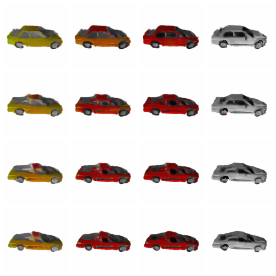}
    \end{minipage} %
\caption{
Left inset: random ShapeVAE (shown as surface normal renders) and translational image GAN samples. 
Middle inset: fixed latent texture and shape across six views.
Right inset: latent interpolations (horizontal: texture changes; vertical: shape changes).
}
\label{fig:gangens}
\end{figure*}

\subsection{Generative Modelling with Unpaired Data} %
\label{sec:genmodel}
Finally, we apply CPDDFs to 3D-aware generative modelling,
using 2D-3D unpaired data 
(see, e.g., \cite{aumentado2020cycle,kaya2020self,miyauchi2018shape,zhu2018visual}). 
This takes advantage of 3D model data, 
yet avoids requiring paired data.
We utilize a two-stage approach: %
(i) a CPDDF-based variational autoencoder (VAE) \cite{kingma2013auto,rezende2014stochastic} on 3D shapes,
then
(ii) a generative adversarial network (GAN) \cite{goodfellow2014generative}, 
which convolutionally translates CPDDF-derived surface normal renders into colour images. %
Briefly, the VAE trains a PointNet encoder \cite{qi2017pointnet} and 
CPDDF decoder,
via a $\beta$-VAE \cite{higgins2016beta} loss, $\mathfrak{L}_{\mathrm{VAE}} = \mathfrak{L}_S + \beta \mathcal{L}_\mathrm{KL}$.
The GAN performs image-to-image translation (from normals to RGB), with %
cyclic
consistency losses \cite{CycleGAN2017}. 
Fig.\ \ref{fig:gangens} displays results on 
ShapeNet cars \cite{shapenet2015,choy20163d},
including disentanglement of shape, viewpoint, and appearance 
(see also, e.g., \cite{schwarz2020graf,kaya2020self,henderson2020leveraging}).
While this underperforms a 2D image GAN 
(15 versus 27 FID \cite{heusel2017gans,obukhov2020torchfidelity}),
it still outperforms samples from image VAEs or GAN-based textured low-poly 3D mesh renders (${>}100$ FID \cite{aumentado2020cycle}) in image quality.
See Appendix \ref{appendix:genmodel} for details.

\section{Discussion}

We have devised \textit{directed distance fields} (DDFs), a novel shape representation, which maps oriented points to depth and visibility values. 
We have examined several useful theoretical properties (including a probabilistic extension for handling discontinuities), 
illustrated the fitting process for single objects 
(as well as composition and UDF extraction)
and applied it to single-image reconstruction and generative modelling.
DDFs are easily differentiably rendered to a surface normal image, which is non-trivial for voxels, NeRFs, or occupancy fields.
Unlike meshes, DDFs are topologically unconstrained, and rendering is independent of shape complexity. 
In contrast to NeRFs or SDFs, we require just a single forward pass per pixel for depth (plus a single backward pass to obtain normals).
One limitation of DDFs is the higher input dimensionality, leading to more complex data and algorithmic requirements, particularly when scaling up to complex scenes. 
Our simplistic approaches to geometric property enforcement and UDF/PC extraction can also be improved, which we leave for future work, along with 
extending the representation to
allow translucency or transmittance, and using a better model of materials, appearance, and lighting.
We also hope to apply DDFs to other tasks, such as haptics and navigation.

{\small
\bibliographystyle{ieee_fullname}
\bibliography{defs,egbib}
}

\include{supp}

\end{document}

%% file: supp.tex
\appendix

\section{Proofs of Geometric Properties}
\label{appendix:proofs}

For a shape $S$, 
    we consider visible oriented points $(p,v)$ 
    (i.e., $\xi(p,v) = 1$), 
    such that $q(p,v) = p + d(p,v) v \in S$, unless otherwise specified.

\subsection{Property I: Directed Eikonal Equation}
\label{appendix:proofs:prop1}

First, note that for any visible $(p,v)$ 
(with $p\notin S$),
there exists an %
$\epsilon > 0$
such that any
$ \delta \in (\delta_0 - \epsilon, \delta_0 + \epsilon) $
satisfies 
$q(p,v) = q(p+\delta v,v) \in S$.
In such a case, by definition of the directed distance field (DDF),
$
d(p + \delta v, v) = d(p,v) - \delta.
$
Restrict $\delta$ to this open interval.
The directional derivative along $v$ with respect to position $p$ is then
\begin{align*}
\nabla_p d(p,v)^T v
&=
\lim_{\delta\rightarrow 0}
\frac{d(p + \delta v, v) - d(p,v)}{ \delta } \\
&=
\lim_{\delta\rightarrow 0}
\frac{d(p, v) - \delta - d(p,v)}{ \delta } \\
&= 
-1,
\end{align*}
as required.

\subsubsection{Gradient Norm Lower Bound}

For any visible $(p,v)$, 
since $|\nabla_p d v| = ||\nabla_p d||_2 ||v||_2\, |\cos(\theta(\nabla_p d, v))| = 1$, 
where $\theta(u_1,u_2)$ denotes the angle between vectors $u_1$ and $u_2$,
we also see that 
$||\nabla_p d||_2 = 1 / |\cos(\theta(\nabla_p d, v))|$,
and thus 
$||\nabla_p d||_2 \geq 1$.

\subsubsection{Visibility Gradient}
Consider the same setup as in \ref{appendix:proofs:prop1}.
The visibility field satisfies a similar property: $\xi(p + \delta v, v) = \xi(p, v)$, as the visibility cannot change when  moving along the same view line (away from $S$).
Thus, we get $\nabla_p \xi(p,v)^Tv = 0$.

\subsection{Property II: Surface Normals}
\label{appendix:proofs:prop2}

Consider a coordinate system with origin $q_0\in S$,
with a frame given by
$(\widehat{i},\widehat{j},\widehat{k})$
where $\widehat{k} = n(q_0)$ and $\widehat{i},\widehat{j}\in\mathcal{T}_{q_0}(S)$ spans the tangent space at $q_0$.
Locally, near $q_0$, reparameterize $S$ in this coordinate system via
\begin{equation}
    S(x,y) = (x, y, f_S(x,y)), \label{app:eq:n0}
\end{equation}
where $f_S$ controls the extension of the surface in the normal direction.
Notice that 
\begin{equation}
    \partial_\alpha S|_{q_0} 
    = (\delta_{x\alpha}, \delta_{y\alpha}, \partial_\alpha f_S|_{q_0}),
\end{equation}
where $\alpha\in\{x,y\}$ and $\delta_{x\alpha}$ is the Kronecker delta,
but since
$\partial_x S|_{q_0} = \widehat{i}$ %
and
$\partial_y S|_{q_0} = \widehat{j}$ %
are in the tangent plane,
\begin{equation}
    \partial_\alpha f_S|_{q_0} = 0. \label{app:eq:n1}
\end{equation}

Consider any oriented position $(p,v)$ that points to $q(p,v)\in S$ on the surface.
Locally, the surface can be reparameterized in terms of $(p,v)$:
\begin{equation}
    q(p,v) = p + d(p,v)v \in S. \label{app:eq:n2}
\end{equation}
Yet, using $q = (q_x,q_y,q_z)$,
we can write this via Eq.\ \ref{app:eq:n0} as 
\begin{equation}
    S(q_x,q_y) = (q_x, q_y, f_S(q_x,q_y)),
\end{equation}
where $q_x$ is the component along the $x$ direction in local coordinates, 
which depends on $p$ and $v$.
In other words, the $z$ component of $q$ depends on $p$ via:
\begin{equation}
    q_z(p,v) = f_S(q_x(p,v), q_y(p,v)). \label{app:eq:n3}
\end{equation}

Let $(p_0,v_0)$ point to $q_0$ (i.e., $q_0 = q(p_0,v_0)$).
Then:
\begin{align*}
    \partial_{p_i} f_S|_{p_0}
    &=
    \underbrace{
    \partial_{q_x} f_S|_{q_0}
    }_0
    \partial_{p_i} q_x|_{p_0}
    +
    \underbrace{
    \partial_{q_y} f_S|_{q_0}
    }_0
    \partial_{p_i} q_y|_{p_0} \\
    &=
    0 \;\, \forall\; i\in\{x,y,z\},
\end{align*}
since $\partial_{q_\alpha} f_S|_{q_0} = 0$ using Eq.\ \ref{app:eq:n1}.

Derivatives with respect to position are then given by
\begin{align}
    \partial_{p_z} q_z|_{p_0}
    &= 1 + \partial_{p_z} d(p,v)|_{p_0} v_z
    = \partial_{p_z} f_S|_{p_0}
    = 0  \label{app:eq:nt1} \\
    \partial_{p_\alpha} q_z|_{p_0}
    &= \partial_{p_\alpha} d(p,v)|_{p_0} v_z
    = \partial_{p_\alpha} f_S|_{p_0}
    = 0,  \label{app:eq:nt2}
\end{align}
using Eq.\ \ref{app:eq:n2} and Eq.\ \ref{app:eq:n3},
with $\alpha\in\{x,y\}$.
Thus, using Eq.\ \ref{app:eq:nt1} and Eq.\ \ref{app:eq:nt2},
\begin{equation}
    \frac{\partial}{\partial(p_x,p_y,p_z)} d(p,v)|_{p_0} = (0, 0, -1/v_z),
\end{equation}
in local coordinates. 
Since the $z$ component here is along the direction of the surface normal, 
$n(q_0) = n(p_0,v_0)$,
this can be rewritten as 
\begin{equation}
    \nabla_p d(p,v)|_{p_0,v_0} = \frac{-1}{v_z} n(q_0)^T = \frac{-n(q_0)^T}{n(q_0)^T v} .
\end{equation}
For any $C^1$ surface $S$, and point $q_0\in S$ (that is intersected by visible oriented point $(p,v)$), we can always construct such a coordinate system,
so we can more generally write:
\begin{equation}
    \nabla_p d(p,v) = \frac{-n^T}{n^T v}.
\end{equation}

\subsection{Property III: Gradient Consistency}
\label{appendix:proofs:prop3}

Consider the same setup (coordinate system) and notation as in the Proof of Property II above (Appendix \ref{appendix:proofs:prop2}).
Since $v\in\mathbb{S}^2$, it suffices to consider an infinitesimal rotational perturbation of some initial view direction $v_0$:
\begin{equation}
    dR(t) = [\omega]_\times dt,
\end{equation}
where 
$[\omega]_\times$ is a skew-symmetric cross-product matrix 
    of some angular velocity vector $\omega$,
so that
$\widetilde{v} = (I + dR(t))v_0$ 
and 
$dv = \widetilde{v} - v_0 =[\omega]_\times\, dt\, v_0 $
is the change in the view (given $dt$) and a velocity of 
$u := \partial_t v = [\omega]_\times v_0$.
The change in surface position,
$q(p,v(t)) = p + d(p,v(t)) v(t)$,
with respect to $t$ is then
\begin{align*}
    \partial_t q 
    &= v\partial_t d + d \partial_t v \\
    &= (\partial_v d \partial_t v)v + d\partial_t v \\
    &= (\partial_v d u)v + d u .
\end{align*}

The $z$ component, in local coordinates, is then
\begin{align*}
    \partial_t q_z|_{q_0} 
    &= \partial_t f_S(q_x,q_y)|_{q_0} \\
    &= \partial_{q_x} f_S|_{q_0} \partial_t q_x + 
       \partial_{q_y} f_S|_{q_0} \partial_t q_y \\
    &= 0,
\end{align*}
via Eq.\ \ref{app:eq:n1}.
Thus, $\partial_t q_z|_{q_0} 
    = (\partial_v d u)v_z + d u_z
    = 0$, meaning
\begin{equation}
    d u^Tn = -(\partial_v d u) v^Tn
\end{equation}
using 
$u_z = u^Tn$
and 
$v_z = v^Tn$ at $q_0$.
Recalling that $\partial_p d = -n^T / (n^Tv)$ and rearranging,
we get 
\begin{equation}
    \partial_v d u = -d \frac{n^T}{v^Tn}u = d \partial_p d u.
\end{equation}
Notice that $u = \partial_t v = [\omega]_\times v_0$ is orthogonal to $v_0$ and the arbitrary vector $\omega$. 
Hence, for any visible oriented point $(p,v)$ away from viewing the boundary of $S$ (i.e., where $n^Tv=0$) and for all $\omega\in\mathbb{R}^3$, we have
\begin{equation}
    \label{app:eq:crossprod}
    \nabla_v d [\omega]_\times v = d\, \nabla_v d [\omega]_\times v.
\end{equation}
This constrains the derivatives with respect to the view vector to be closely related to those with respect to position, along any directions not parallel to $v$.

\subsubsection{Alternative Expression}
Note that the inner product with $\delta_v$ primarily serves to restrict the directional derivative in valid directions of $v$.
A cleaner expression can be obtained with a projection operator, which removes directional components parallel to $v$.

First, let us define $d$ to normalize $v$: 
\begin{equation}
\label{app:eq:altdv:normalized}
d(p,v) := d\left( p, \frac{v}{||v||_2} \right), 
\end{equation}
for all $v\in\mathbb{R}^3 \setminus \{0\}$. 
Then, consider a perturbation along the view direction of size $|\delta| < 1$:
\begin{equation}
    \label{app:eq:altdv:zerochangea}
    d(p_0, v_0 + \delta v_0)
    = d(p_0, v_0(1+\delta))
    =d(p_0,v_0),
\end{equation}
with $||v_0|| = 1$ and
using Eq.\ \ref{app:eq:altdv:normalized} for the last step.
This means the directional derivative along $v$ must satisfy \begin{equation}
    \label{app:eq:altdv:zerochangeb}
    \partial_v d(p,v) v = 0.
\end{equation}
In the previous section, we showed that
$\partial_v d u = d\partial_p d u$ for all $u \perp v$.
Let $\mathcal{P}_v = I - vv^T$ be the orthogonal projection removing components parallel to $v$.
Then we can rewrite the result of the previous section
as $\partial_v d \mathcal{P}_v = d\partial_p d \mathcal{P}_v$.
But $\partial_v d \mathcal{P}_v = \partial_v d - \partial_v d vv^T = \partial_v d$ by Eq.\ \ref{app:eq:altdv:zerochangeb}.
Thus, we may write
\begin{equation}
    \partial_v d = d\, \partial_p d\, \mathcal{P}_v,
\end{equation}
which agrees with Eq.\ \ref{app:eq:crossprod}.

\subsection{Property V: Local Differential Geometry}
\label{appendix:proofs:prop5}

Given a visible oriented point $(p,v)$, 
    we have shown that the surface normal $n(p,v)$ on $S$
    (at $q = p + d(p,v)v$)
    is computable from $\nabla_p d$ (see Appendix \ref{appendix:proofs:prop2}).
Curvatures, however, require second-order information.

We first construct a local coordinate system via $n$, 
    by choosing two tangent vectors at $q$: 
    $t_x,t_y\in\mathcal{T}_q(S)$, 
    where $||t_\alpha||_2=1$ and $t_x^T t_y=0$.
In practice, this can be done by sampling Gaussian vectors, 
    and extracting an orthogonal tangent basis from them.
We can then reparameterize the surface near $q_0 = q(p_0,v_0)$ (with surface normal $n_0 = n(q_0)$) via
$S(x,y) = q_0 + xt_x + yt_y + f_S(x,y)n_0$,
where $x$ and $y$ effectively control the position on the tangent plane.
Alternatively, 
we can write $S(x,y) = q(p(u),v_0)$,
which parameterizes $S$ about the oriented point 
(or viewpoint) $p(u)$, 
where $u=(x,y)$, and $p(u) = p_0 + xt_x + yt_y$.
Notice that $p(u)$ is essentially a local movement of $p$ parallel to the tangent plane at $q_0\in S$ (which is ``pointed to'' by $(p_0,v_0)$). Note that $p_0$, $v_0$, and $q_0$ are fixed; only $u$ and $p(u)$ are varied.
Further, notice that the plane defined by $p(u)$ (in the normalized tangent directions $t_x$ and $t_y$) is parallel to the surface tangent plane at $q_0$ (and thus orthogonal to $n_0$), but \textit{not} necessarily perpendicular to $v$.

Using this local frame on $\mathcal{T}_{q_0}(S)$,
the first-order derivatives of the surface are
\begin{align}
    \partial_{i} S|_{u=0}
    &= \partial_{i} \left( p(u) + v_0 d(p(u),v_0) \right)|_{u=0} \\
    &= t_i + \partial_{i} d(p(u),v_0) |_{u=0} v_0 \\
    &= t_i + (\nabla_p d(p,v_0) t_i) v_0
\end{align}
where $i\in\{x,y\}$ and $p_j$ is the $j$th component of $p$.
Then the metric tensor (first fundamental form) is given by
\begin{align}
    g_{ij} 
    &= \partial_{i} S|_{u=0}^T \, \partial_j S|_{u=0} \\
    &= \delta_{ij} + c_ic_j + c_i v_0^Tt_j + c_j v_0^T t_i
    \label{appendix:eq:gfail},
\end{align}
where $i,j\in\{x,y\}$, $u=(x,y)$, 
$\delta_{ij}$ is the Kronecker delta function, and $c_k = \nabla_p d t_k$.
Notice that $\nabla_p d$ is parallel to $n$ 
    (see Property \hyperref[property2]{II}), 
    and thus orthogonal to $t_1$ and $t_2$.
Thus, $c_x = c_y = 0$, and so, 
    in theory, $g_{ij} = \delta_{ij}$, 
    but this may not hold for a learned DDF,
    if one computes directly with Eq.\ \ref{appendix:eq:gfail}.
However, in practice, 
    we use the theoretical value (where $g = I_2$ in local coordinates),
    as deviations from it are due to error only (assuming correct normals).

Next, the shape tensor (second fundamental form) $\RomanII_{ij}$ can be computed from the second-order derivatives of the surface \cite{kreyszigdg} via:
\begin{align}
\RomanII_{ij}
    &= n_0^T \partial_{ij} S|_{u=0} \\
    &= n_0^T \partial_{ij} ( p(u) + v_0 d(p(u),v_0) )|_{u=0} \nonumber\\
    &= n_0^T v_0 \partial_{ij}  d(p(u),v_0) |_{u=0}  \nonumber\\ 
    &= n_0^T v_0 \partial_i \sum_k \partial_{p_k} d(p(u),v_0) 
        \underbrace{[\partial_j p(u)]_k}_{[t_j]_k} |_{u=0}  \nonumber\\
    &= n_0^T v_0 \sum_k \partial_i \partial_{p_k} d(p(u),v_0) 
        [t_j]_k |_{u=0}  \nonumber\\
    &= n_0^T v_0 \sum_k [t_j]_k 
            \sum_\ell 
                \partial_{p_\ell} \partial_{p_k} d(p(u),v_0)
                [\partial_i p(u)]_\ell
         |_{u=0}  \nonumber\\
    &= n_0^T v_0 \sum_k [t_j]_k 
            \sum_\ell 
                \underbrace{
                \partial_{p_\ell,p_k} d(p(u),v_0)
                }_{ \mathcal{H}_p[d]_{k \ell} }
                [t_i]_\ell
         |_{u=0}  \nonumber\\
    &= \left( t_j^T \mathcal{H}_p[d] t_i \right) n_0^T v_0,
        \label{app:eq:curvderiv} 
\end{align}
where 
$i,j\in\{x,y\}$, 
$u=(x,y)$, 
$n_0 = n(q_0)$,
$[t_i]_k$ is the $k$th component of $t_i$,
$\mathcal{H}_p[d]$ is the Hessian of $d$ with respect to $p$ 
    (at $u=0$),
and 
$p(u) = p_0 + x t_x + y t_y$.
Note that our parameterization of the surface using the DDF (i.e., $q(p(u),v_0)$) defines surface deviations in terms of $v_0$; the $n_0^T v_0$ term undoes this effect, rewriting the deviation in terms of $n$ instead.

Curvatures can then be computed via the shape tensor (see, e.g., \cite{kreyszigdg}).
Gaussian curvature is given by
$\mathcal{C}_K = \det(\RomanII)/\det(g)$, 
while mean curvature is written
$\mathcal{C}_H = \mathrm{tr}(\RomanII g^{-1})$.
Notice that these quantities can be computed for any visible $(p,v)$, 
    using only $d(p,v)$ and derivatives of $d$ with respect to $p$ (e.g., with auto-differentiation).
Thus, the curvatures of any visible surface point can be computed using only local field information at that oriented point.

\subsection{View Consistency}

\label{appendix:viewconsis}

DDFs ideally satisfy a form of view consistency,
    which demands that an opaque position viewed from one direction 
    must be opaque from all directions, 
    with depth lower-bounded by that known surface position.
Consider two oriented points $(p_1,v_1)$ and $(p_2,v_2)$.
Assume 
(i) that $(p_1,v_1)$ is visible, 
such that $q_1 = p_1 + d(p_1,v_1)v_1 \in S$,
and 
(ii) that there exists $t>0$ 
such that $\ell_{p_2,v_2}(t) = p_2 + t v_2 = q_1$.
That is, 
    if viewing the scene via $(p_2,v_2)$,
    we assume our line of sight intersects that of $(p_1,v_1)$, at the surface point $q_1$.
    
Then, in this case, we have that: 
(1) $(p_2,v_2)$ is visible, meaning $\xi(p_2,v_2) = 1$,
and
(2) $d(p_2,v_2) \leq ||p_2 - q_1||_2 = t$.
These can be seen by the definition of $\xi$ and $d$.
For (1), since there exists a surface intersection (at $q_1$) along $\ell_{p_2,v_2}$, the oriented point $(p_2,v_2)$ must be visible (i.e., $\xi(p_2,v_2) = 1$).
For (2), $d(p_2,v_2)$ can be no further than $t$, since DDFs return the minimum distance  to a point on $S$ along the line $\ell_{p_2,v_2}(t)$, and hence its output can be no greater than the known (assumed) distance $t$.

\subsection{Neural Depth Renderers as DDFs}
\label{appendix:neurren}
A natural question is how to parallelize a DDF.
Consider a function $f : \Lambda_\Pi \times \Lambda_S \rightarrow \mathcal{I}_{d,\xi}^m$, where $ \mathcal{I}_{d,\xi} = \mathbb{R}_+ \times [0,1]$, from continuous camera parameters (i.e., elements of $\Lambda_\Pi$) and some space of shapes $\Lambda_S$ to an $m$-pixel depth and visibility image.
Then $f$ is implicitly a conditional DDF, as each pixel value can be computed as $(d(p,v), \xi(p,v))$, where $p$ and $v$ are a fixed function of $\Pi\in\Lambda_\Pi$.
In other words, the camera intrinsics and extrinsics determine the $p$ and $v$ value for each pixel, with $p$ determined by the centre of projection (camera position) and $v$ determined by the direction corresponding to that pixel, from the centre of projection out into the scene.
We may thus regard the depth pixel value as a function of $p$ and $v$, via the camera parameters.
This holds regardless of the architecture of $f$. %
Thus, all the properties of DDFs hold in these cases as well: for example, for a depth image $I_d$, camera position $C_p$ and viewpoint $C_v$,  Property \hyperref[property2]{II} relates $\partial_{C_p} I_d$ to the surface normals image, while Property \hyperref[property3]{III} constrains $\partial_{C_v} I_d$.
We believe this point of view can improve the framework for differentiable neural rendering of geometric quantities (e.g., \cite{yan2016perspective,wu2017marrnet,tulsiani2017multi,nguyen2018rendernet}).

\newcommand{\trw}{0.143\textwidth}

\begin{figure*}[ht]
    \centering
        \includegraphics[clip,trim={0.0cm 0.0cm 0.0cm 0.0cm},width=0.11\textwidth]{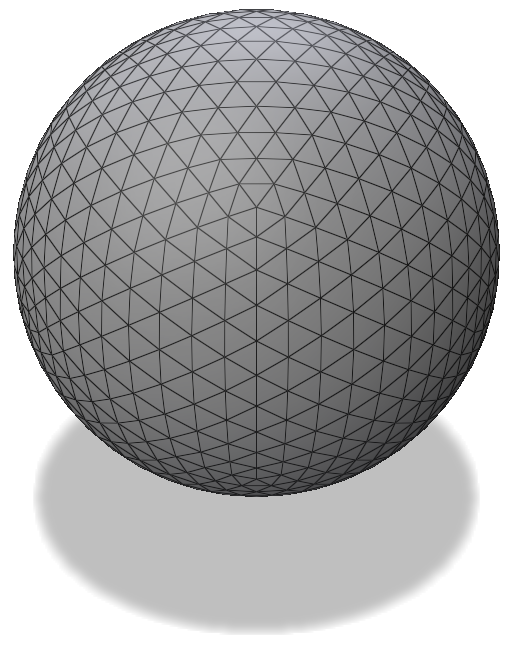} \hfill
        \includegraphics[clip,trim={12.2cm 1cm 12.2cm 1cm},width=\trw]{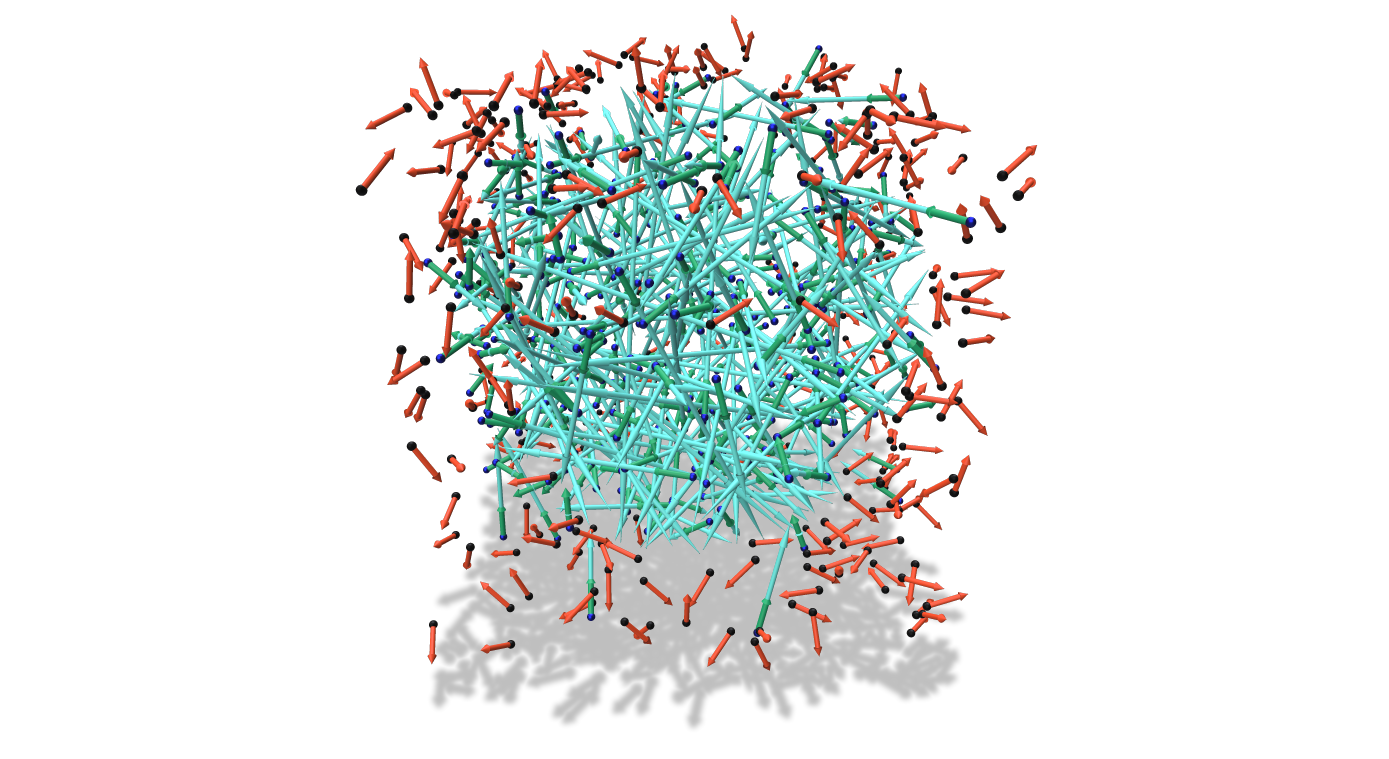} \hfill
        \includegraphics[clip,trim={12.2cm 1cm 12.2cm 1cm},width=\trw]{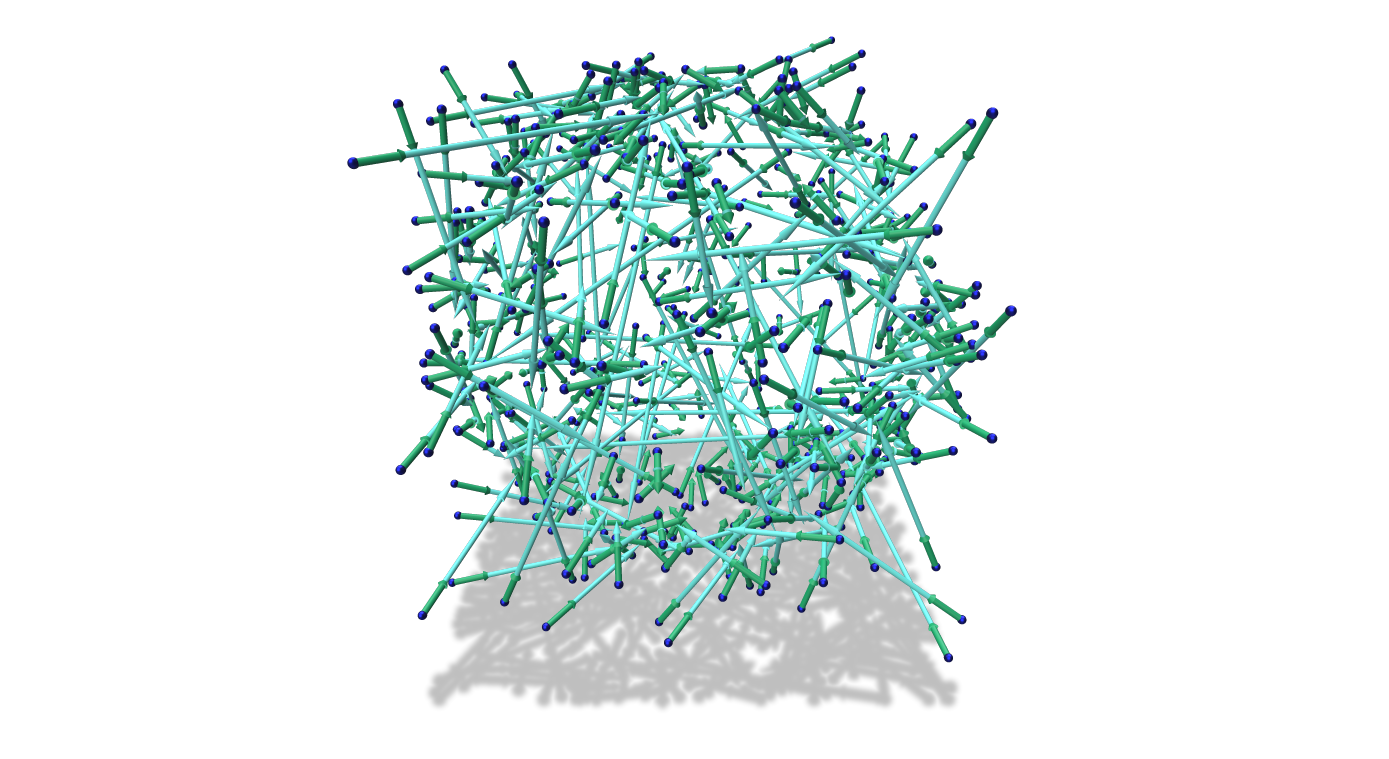} \hfill
        \includegraphics[clip,trim={12.2cm 1cm 12.2cm 1cm},width=\trw]{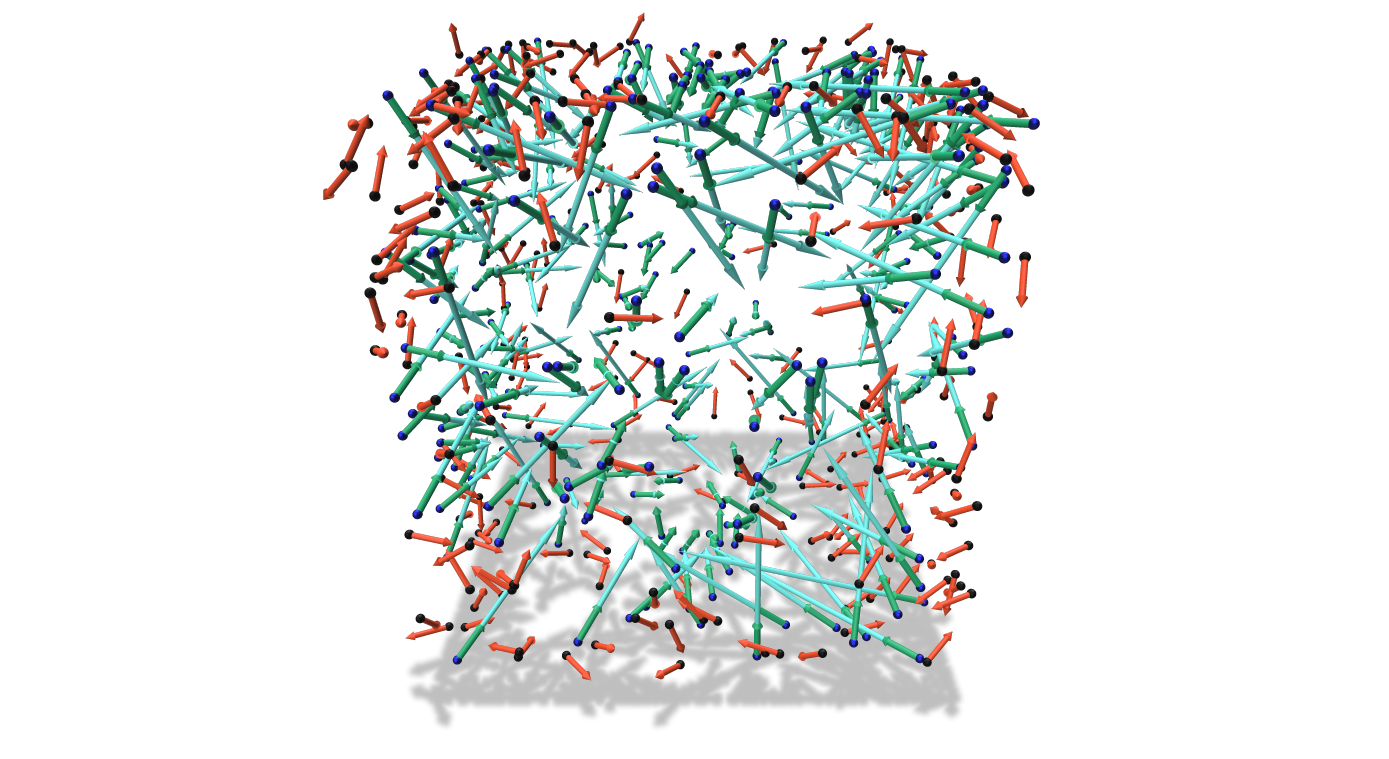} \hfill
        \includegraphics[clip,trim={12.2cm 1cm 12.2cm 1cm},width=\trw]{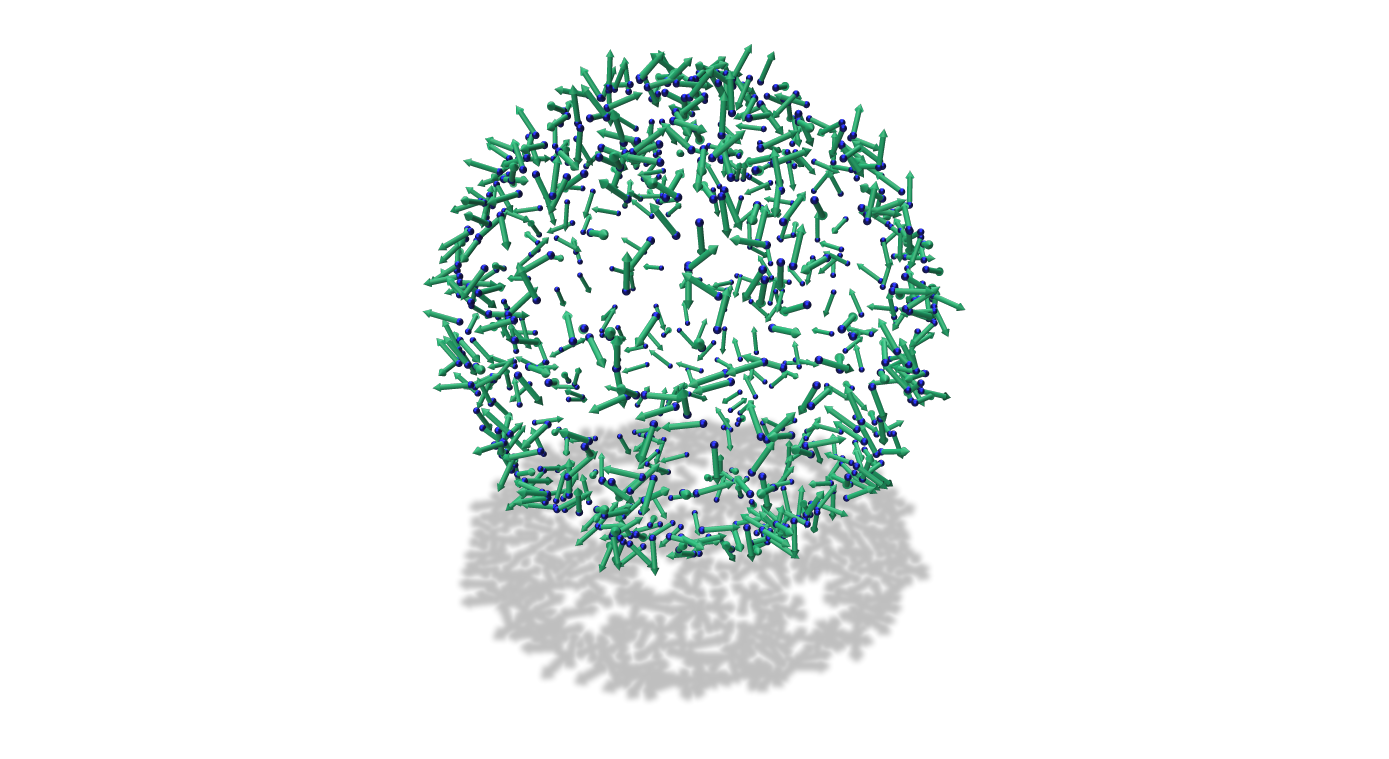} \hfill
        \includegraphics[clip,trim={12.2cm 1cm 12.2cm 1cm},width=\trw]{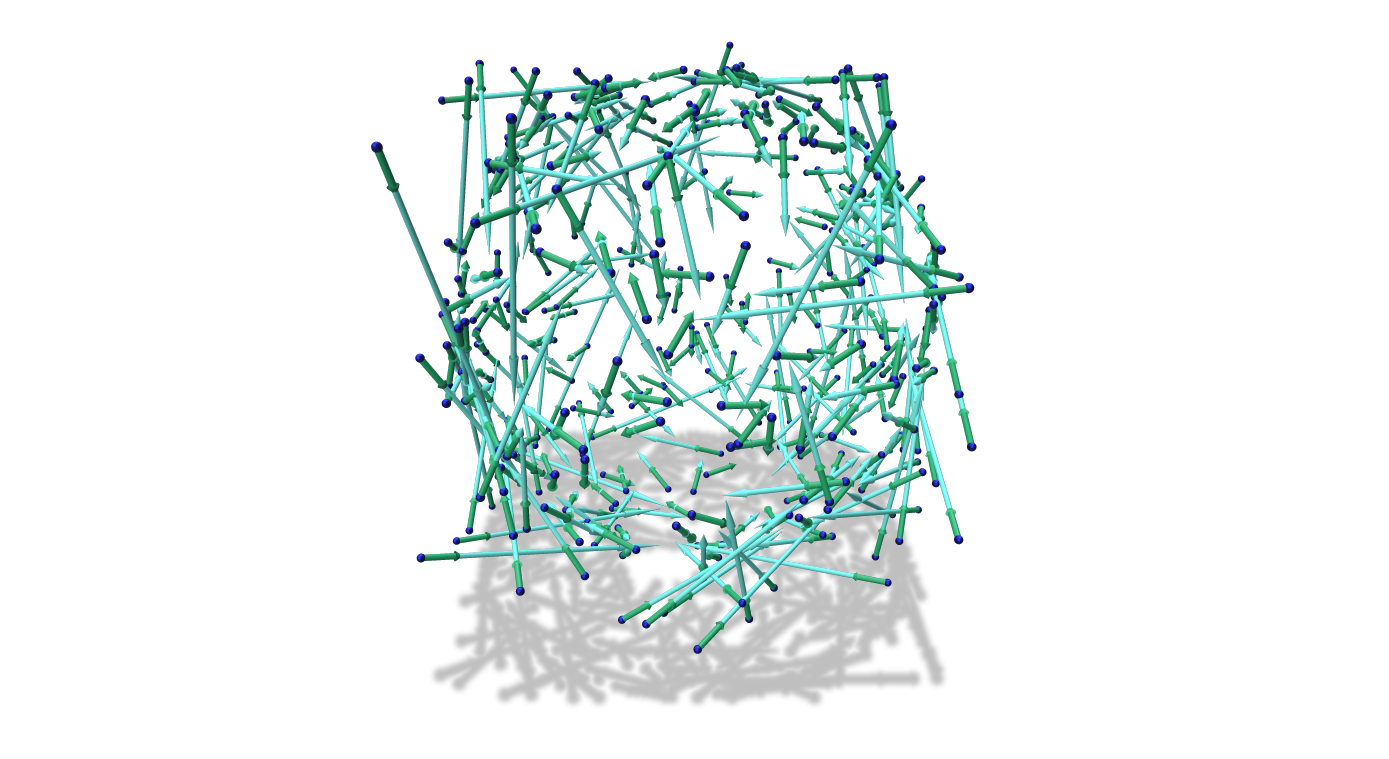} \hfill
        \includegraphics[clip,trim={12.2cm 1cm 12.2cm 1cm},width=\trw]{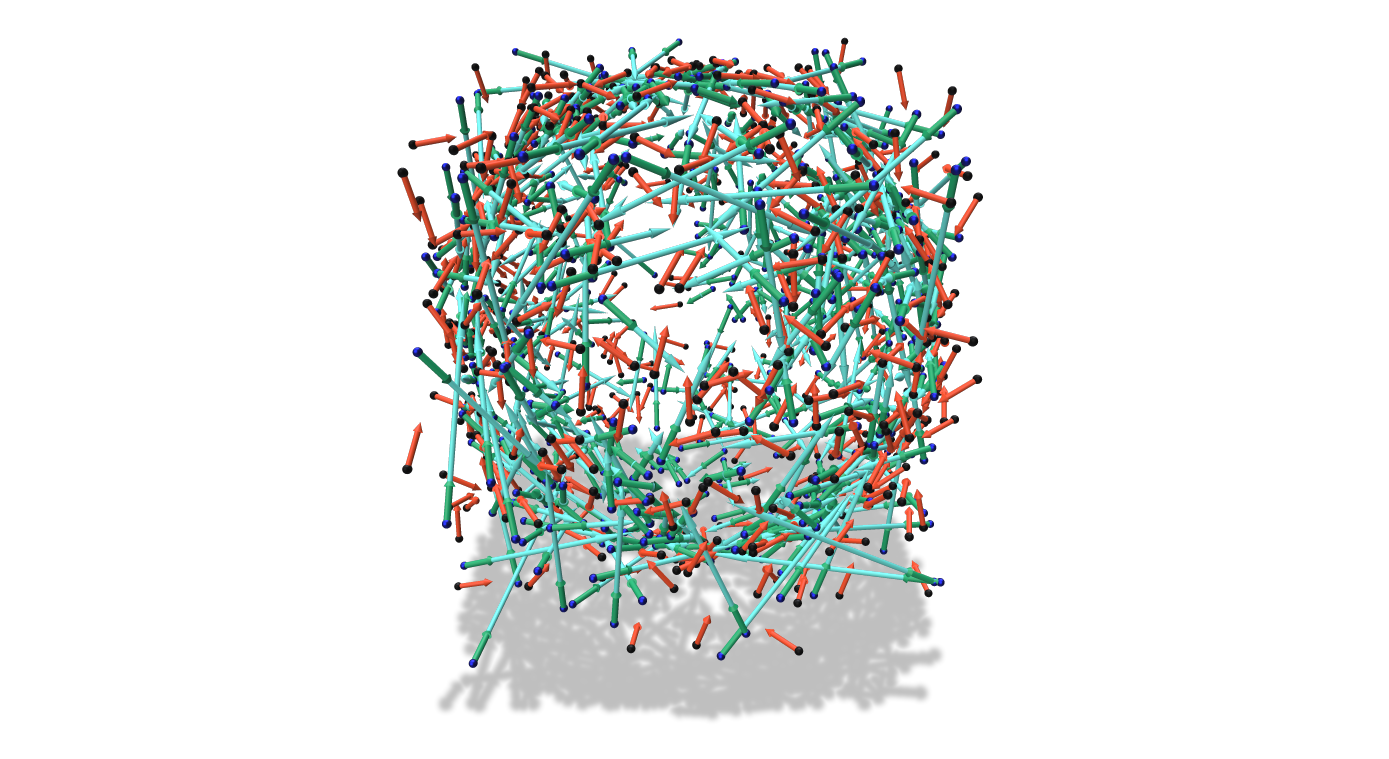}
    \makebox[0pt][r]{%
    \begin{tikzpicture}[overlay]%
    \node[xshift=0cm] (hidden) at (0,0) {};
    \node[xshift=0cm,yshift=0cm] (M) at (-7.20,0.60) {\footnotesize Mesh};
    \node[xshift=0cm,yshift=0cm] (U) [right=0.8in of M] {\footnotesize U};
    \node[xshift=0cm,yshift=0cm] (A) [right=0.85in of U] {\footnotesize A};
    \node[xshift=0cm,yshift=0cm] (B) [right=0.85in of A] {\footnotesize B};
    \node[xshift=0cm,yshift=0cm] (S) [right=0.85in of B] {\footnotesize S};
    \node[xshift=0cm,yshift=0cm] (T) [right=0.85in of S] {\footnotesize T};
    \node[xshift=0cm,yshift=0cm] (O) [right=0.85in of T] {\footnotesize O};
    \end{tikzpicture}%
    }%
    \vspace{-0.1in}
    \caption{Illustration \cite{polyscope} of data types. Left to right: input sphere mesh, U, A, B, S, T, and O data. Visible points depict $p$ in blue, $v$ in green, and a line from $p$ to $q$ in turquoise; non-visible points depict $p$ in black and $v$ in red. 
    See \S\ref{sec:app:data} for details and \S\ref{app:sec:dataablation} for an ablation study of the data types. 
    }
    \label{fig:datatypes}
\end{figure*}
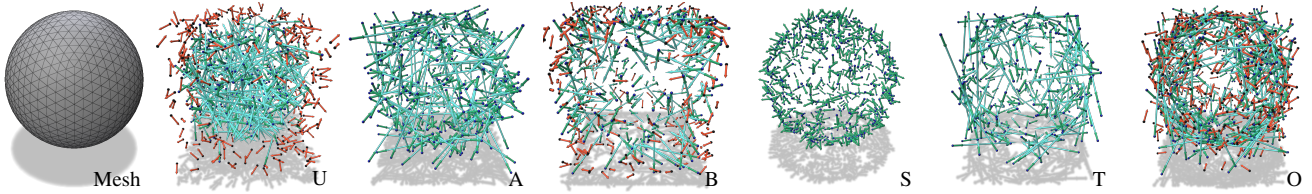

\section{Mesh Data}
\label{sec:datatypes}

We display visualizations of each data type in Fig. \ref{fig:datatypes}.
We briefly describe how each is computed (note that we only need to obtain $(p,v)$, after which $(\xi,d)$ can be obtained via ray-triangle intersection):
\begin{itemize}
    \item Uniform (U): simply sample $p\sim\mathcal{U}[\mathcal{B}]$ and $v\sim\mathcal{U}[\mathbb{S}^2]$.
    \item At-surface (A): start with $q_0\sim\mathcal{U}[S]$ (obtained via area-weighted sampling) and $v_0\sim\mathcal{U}[\mathbb{S}^2]$. Sample $p$ on the line between $q_0$ and its intersection with $\mathcal{B}$ along $v_0$. Set $v = -v_0$. Note that the final output data may not actually intersect $q_0$ (since one may pass through a surface when sampling $p$).
    \item Bounding (B): sample $p\sim\mathcal{U}[\partial \mathcal{B}]$ and $v\sim\mathcal{U}[\mathbb{S}^2]$, but restrict $v$ to point to the interior of $\mathcal{B}$. 
    \item Surface (S): simply use $p\sim\mathcal{U}[S]$ and then take $v\sim\mathcal{U}[\mathbb{S}^2]$.
    \item Tangent (T): the procedure is the same as for A-type data, except we enforce $v_0$ to lie in the tangent plane $\mathcal{T}_{q_0}(S)$.
    \item Offset (O): take a T-type $(p,v)$ and simply do $p\leftarrow p + \varsigma_O \epsilon_O n_0$, where $n_0$ is the normal at $q_0$, and we set $\epsilon_O = 0.05$ and sample $\varsigma_O \sim \mathcal{U}[\{-1,1\}]$.
\end{itemize}
Since we assume $\mathcal{B}$ is an axis aligned box (with maximum length of 2; i.e., at least one dimension is $[-1,1]$), sampling positions on it, or directions with respect to it, is straightforward. See also \S\ref{app:sec:dataablation}, which examines the effect of ablating each data type.

\section{Single Entity Fitting Details}
\label{appendix:singlefits}

Given a mesh, we first extract data of each type 
(see \S\ref{sec:app:data}), obtaining $(p,v)$ tuples in the following amounts:
250K (A and U) and 125K (B, T, O, and S). 
Since rendering outside $\mathcal{B}$ uses query points on $\partial\mathcal{B}$, we bias the sampling procedure for T, A, and O data, 
    such that 10\% of the $p$-values are sampled from $\partial\mathcal{B}$.
For each minibatch, we sample
6K (A and U) and 3K (B, T, O, and S) points, across data types. 
In addition to these, we sample an additional 1K 
uniformly random oriented points per minibatch,
on which we compute only regularization losses 
($\mathcal{L}_V$ and $\mathcal{L}_\mathrm{DE}$), 
for which ground truth values are not needed.

We note that not all loss terms are applied to all data types. As discussed in \S\ref{sec:app:learning}, the transition loss $\mathcal{L}_T$ is only applied to S and T type data (since we do not want spurious field discontinuities, due to the field switching between components, except when necessary).
As briefly noted in Property \hyperref[property2]{II}, the DDF gradient $\nabla_p d$ (and hence field-derived surface normals) are not well-defined when $n$ and $v$ are orthogonal.
The gradients are also not well-defined on S-type data (since $d$ is explicitly discontinuous for $p\in S$, and thus $\nabla_p d$ does not exist there). 
Hence, we do not apply the directed Eikonal regularization $\mathcal{L}_\text{DE}$ or the normals loss $\mathcal{L}_n$ to S, T, or O data.
Similarly, the weight variance loss $\mathcal{L}_V$ (designed to reduce weight field entropy) should not be applied to S, T, or O samples, as the weight field should be transitioning near those samples (and thus have a higher pointwise Bernoulli variance).
The remaining losses are applied on all data types.

We then optimize Eq.\ \ref{eq:singlefitall} via Adam 
    (learning rate: $10^{-4}$; $\beta_1 = 0.9$, $\beta_2 = 0.999$), 
    using a reduction-on-plateau learning rate schedule 
    (reduction by 0.9, minimum of 5K iterations between reductions).
We run for 100K iterations,
using 
$\gamma_d = 5$,
$\gamma_\xi = 1$,
$\gamma_n = 10$,
$\gamma_V = 1$,
$\gamma_\mathrm{E,d} = 0.05$, 
$\gamma_\mathrm{E,\xi} = 0.01$, and
$\gamma_T = 0.25$.
Note that we double $\gamma_d$ on A and U data.
The field itself is implemented as a SIREN with seven hidden layers, initialized with $\omega_0 = 1$, each of size 512, mapping $(p,v)\in\mathbb{R}^6 \rightarrow (\{d_i\}_{i=1}^K,\{w_i\}_{i=1}^K,\xi)\in\mathbb{R}^{2K+1}$ 
(note that the set of $w_i$'s has $K-1$ degrees of freedom).
We use $K=2$ delta components.
Since $w_2 = 1 - w_1$, we output only $w_1$ and pass it through a sigmoid non-linearity. 
We apply ReLU to all $d_j$ outputs, to enforce positive values, and sigmoid to $\xi$.
We use the implementation from \cite{aman_dalmia_2020_3902941} for SIREN.
In terms of data, we test on shapes from the Stanford Scanning Repository
\cite{stanfordscanningrepo} (data specifically from
\cite{turk1994zippered,curless1996volumetric,krishnamurthy1996fitting}).

\subsection{Non-Compositional Fitting Example}
\label{app:noncompfiteg}

To show the improved scaling of composing DDFs, we also attempted to fit the same scene using the single-object fitting procedure above. 
For fairness, we doubled the number of data samples extracted, as well as the size of each hidden layer.
In comparison to Fig.\ \ref{fig:simpleroom}, this naive approach struggles to capture some high frequency details, though we suspect this could be mitigated by better sampling procedures.

\renewcommand{\ctwca}{0.09\textwidth}
\begin{figure}
    \centering
    \includegraphics[width=\linewidth]{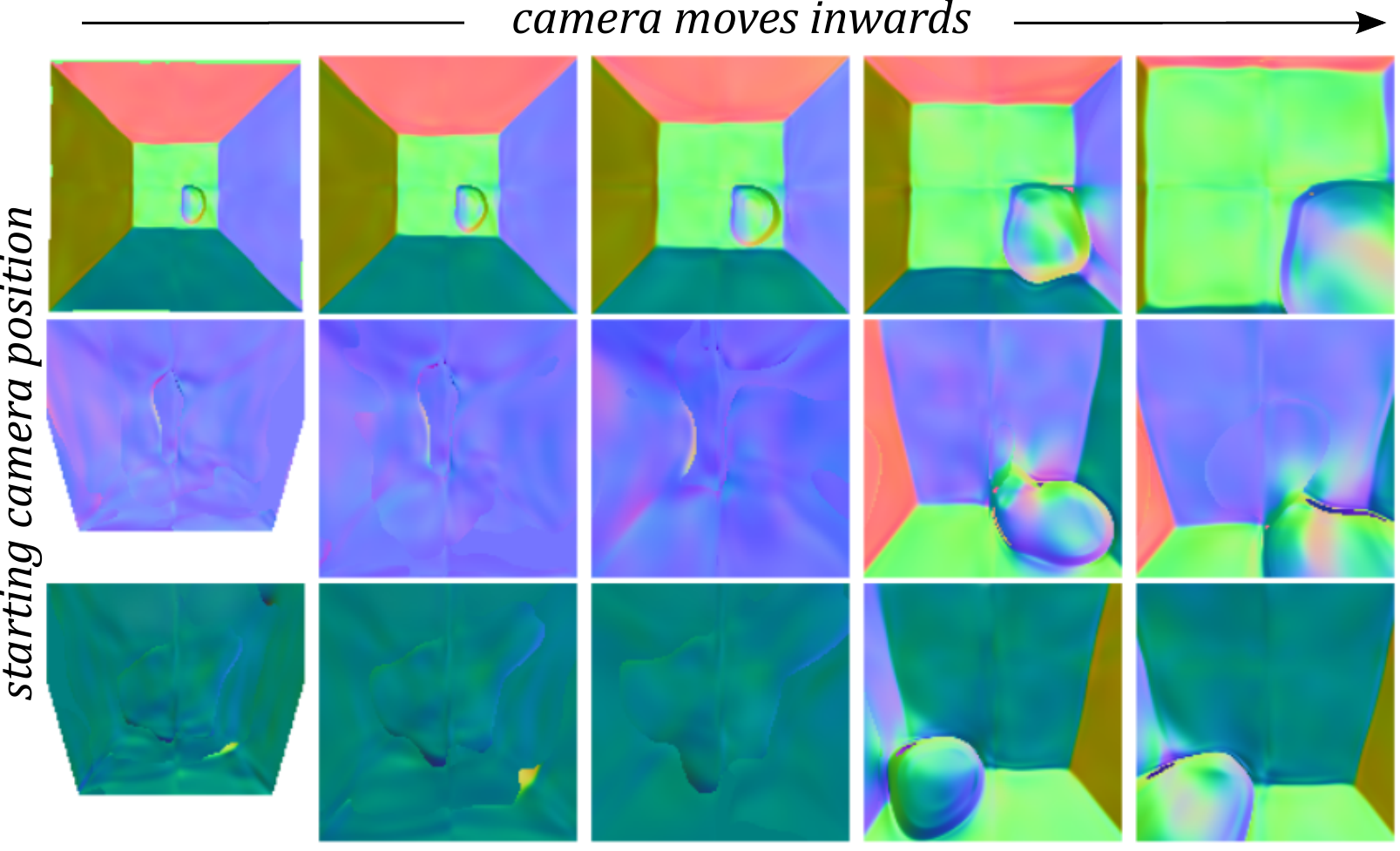}
    \caption{
        Example showing underfitting without compositional model (same scene as Fig.\ \ref{fig:simpleroom}). 
        Colours correspond to surface normals estimated via the DDF. 
        (See Appendix \ref{app:noncompfiteg}).
    }
    \label{app:fig:simpleroom2xsl}
\end{figure}

\subsection{Data Sample Type Ablation Experiment}
\label{app:sec:dataablation}

\definecolor{ultramarine}{RGB}{0,32,96}
\definecolor{tred}{RGB}{191,10,10}
\definecolor{tpink}{RGB}{232,114,114}
\newcommand{\tred}[1]{\textbf{\textcolor{tred}{#1}}}
\newcommand{\tpink}[1]{\textbf{\textcolor{tpink}{#1}}}

\begin{table*}
\centering
\begin{tabular}{c|cccccc||cccccc}
\multirow{2}{*}{Ablation} & 
    \multicolumn{6}{c||}{Minimum Distance Error ($L_1; \times 10$) $\downarrow$} &
    \multicolumn{6}{c}{Visibility Error (BCE) $\downarrow$} \\ 
    & $\mathcal{L}_{d,1}$-U & $\mathcal{L}_{d,1}$-A & $\mathcal{L}_{d,1}$-B & $\mathcal{L}_{d,1}$-O & $\mathcal{L}_{d,1}$-T & $\mathcal{L}_{d,1}$-S & %
    $\mathcal{L}_{\xi}$-U & $\mathcal{L}_{\xi}$-A & $\mathcal{L}_{\xi}$-B & $\mathcal{L}_{\xi}$-O & $\mathcal{L}_{\xi}$-T & $\mathcal{L}_{\xi}$-S \\\hline
    U & \tred{0.58} & 0.79 & 0.18 & 0.49 & 0.75 & 0.47 & 
        \tred{2.11} & 0.03 & 0.07 & \tpink{0.56} & 0.14 & 0.05
    \\
    A & 0.48 & \tpink{0.82} & \tpink{0.20} & 0.49 & 0.81 & 0.54 &  
        0.20 & 0.05 & 0.07 & 0.49 & \tpink{0.18} & \tpink{0.07}
    \\
    B & 0.37 & 0.67 & \tpink{0.20} & 0.46 & 0.73 & 0.58 &
        0.20 & 0.03 & \tpink{0.10} & 0.50 & 0.15 & 0.06
    \\
    O & 0.39 & 0.67 & 0.18 & \tred{0.56} & 0.70 & 0.59 &
        \tpink{0.28} & 0.01 & \tred{0.11} & \tred{1.71} & 0.03 & 0.03
    \\
    T & 0.39 & 0.70 & 0.19 & 0.45 & \tpink{0.84} & {0.65} & 
        0.19 & \tred{0.09} & 0.06 & 0.32 & \tred{0.48} & \tpink{0.07}
    \\
    S & \tpink{0.55} & \tred{0.95} & \tred{0.23} & \tpink{0.51} & \tred{0.89} & \tred{1.43} &
        0.14 & \tpink{0.06} & 0.06 & 0.42 & 0.17 & \tred{0.48}
    \\\hline
   -- & 0.45 & 0.75 & 0.19 & 0.50 & 0.77 & \tpink{0.67} &
        0.23 & 0.04 & 0.08 & \tpink{0.56} & 0.15 & \tpink{0.07}
\end{tabular}
\caption{
Data sample type ablation on the Stanford Bunny 
    (see \S\ref{app:sec:dataablation}).
Rows: type of data (i.e., $(p,v)$ sample type) ablated.
Columns: errors on held-out data (left: minimum distance loss, but computed with $L_1$ instead of $L_2$, for greater interpretability; right: binary cross-entropy-based visibility loss).
Each column computes the error on a different sample type (e.g., $\mathcal{L}_{d,1}$-A is the minimum distance error on A-type data).
Error computation is on 25K held-out samples, for each data type.
Per loss type, the \tred{red} numbers are the \textit{worst} (highest) error cases; the \tpink{pink} numbers are the \textit{second-worst} error cases.
Each ablation scenario uses 100K of each data type, except for the ablated one (of which it uses zero);
the ``--'' case uses 83,333 samples of all six data types (to control for the total number of points).
Several observations are notable:
(1) in most cases, performance on a data type is worst or second-worst when that type is ablated in training;
(2) ablating S appears to be damaging across other data types as well; and
(3) there is some interplay between data types 
(e.g., ablating T-type data is worse for A-type data error, than ablating A-type data itself).
}
\label{tab:dataablation}
\end{table*}

We perform a small-scale ablation experiment to discern the importance of each data sample type.
We perform a single-shape fit to the Stanford Bunny, with slight modifications to the algorithm discussed in \S\ref{appendix:singlefits} just above.
In particular, we consider six scenarios, in each of which we train with 100K samples of each type except one, which is removed/ablated.
We consider one other scenario that has the same number of total points, but no single type is ablated (i.e., 83,333 samples for each data type).
We then measure the depth and visibility prediction error on 25K held-out samples of each data type, including the ablated one.
All other aspects of the fitting process remain the same.

See Table \ref{tab:dataablation} for results.
We first notice that the worst errors are incurred for models trained where that data type was ablated in training. 
For the minimum distance error, the worse or second-worst result is always obtained for the model that ablates that data type; this is also true for the visibility error, except for ablating and testing on A-type data. 
This case may be due to the relation between A-type and T-type data, since T-type data is effectively the hardest subset of A-type data (excepting perhaps S-type data).
There is also significant overlap between U-type and B-type data with A-type data.
Hence, we speculate that ablating T-type data removes access to the most difficult (and most likely to have high error) of A-type samples, leading to high visibility error, whereas ablating A-type data itself can in part be accounted for by other data types (e.g., U-type).

One surprising scenario is the S-type data ablation, as it worsens performance across all other data types as well. This is potentially due to the greater difficulty in the network knowing when to transition between Dirac delta components in the weight field.

Finally, we also found that ablating B-type data has a relatively minor effect on the resulting errors.
We remark that, for \textit{rendering} applications, where the camera is more likely to be outside the box (and thus most field query positions $p$ will be on $\partial\mathcal{B}$; see \S\ref{sec:app:rendering}), B-type data will be much more important, and hence useful to concentrate on.
Further, we note that the data bias for T-, A-, and O-type data 
(ensuring there are positions $p\in\partial \mathcal{B}$) leads to increased overlap with B-type data, reducing the effect of ablating it.
Other effects, such as enforcing the directed Eikonal property, should help as well.
We remark that both A and B sample types are biased: specifically, caring more about the shape than the non-visible parts of the scene (for A), and focusing on rendering-oriented applications (for B).

Overall, though this is only on a single shape with a simplistic set of scenarios, it suggests that each data type has information that the other types cannot completely make up for, especially for U, O, T, and S data.

\section{UDF and $v^*$}
\label{appendix:udf}

\subsection{$v^*$ Extraction}
\label{appendix:udf:vstarextract}

We obtain $v^*$ via a fitting procedure, with a forward pass similar to composition 
    (see \S\ref{sec:results:singlefieldfitting}).
Starting from an already trained (P)DDF, 
    we define a small new SIREN network $g_v$
    (five hidden layers, each size 128),
    which maps position to a set of $K_c$ \textit{candidate directions}, 
    such that
    $g_v:\mathbb{R}^3\rightarrow\mathbb{S}^{2\times K_c}$
Given a position $p$ and candidates $\{v^*_i\}_{i=1}^{K_c} = g_v(p)$,
    we can compute 
    $ \zeta_{v^*} = \{ d(p,v^*_i(p)), \xi(p,v^*_i(p)) \}_{i=1}^{K_c} $.
To obtain a UDF depth estimate, we use 
\begin{equation}
    \widehat{\mathrm{UDF}}(p) = 
    \sum_i \omega_{\zeta_{v^*}}^{(i)}(p) \, d(p,v^*_i(p)),
\end{equation}
where the weights $\omega_{\zeta_{v^*}}$ are based on those from our explicit composition method:
\begin{equation}
    \omega_{\zeta_{v^*}}(p)
    =
    \mathrm{Softmax}
        \left(  
            \left\{
                \frac{ 
                    \eta_{T}^{-1} \xi(p,v^*_i(p))
                }{ 
                    \varepsilon_s + d(p,v^*_i(p)) 
                }
            \right\}_i\,
        \right).
\end{equation}
We use the same $\eta_T$ and $\varepsilon_s$ as for composition.
Note that this formulation essentially takes the candidate directions, computes their associated distances, and then (softly) chooses the one that has lowest distance value while still being visible. 
We can also obtain $v^*(p)$ via these weights, using
\begin{align}
    \widetilde{v}^*(p) &= 
    \sum_i \omega_{\zeta_{v^*}}^{(i)}(p) \, v^*_i(p) \\
    {v}^*(p) &= 
    \frac{\widetilde{v}^*(p)}{||\widetilde{v}^*(p)||_2} ,
\end{align}
Note that, in the ideal case, $v^*(p) = -\nabla_p \mathrm{UDF}(p)^T$, 
    meaning we could compute it with a backward pass,
    but we found such an approach was noisier in practice.

To train $g_v$ to obtain good candidates, we use the following loss: 
\begin{align}
    \mathcal{L}_{v^*}
    &= \frac{1}{K_c}\sum_i d(p,v^*_i(p)) - \xi(p,v^*_i(p)) \label{app:eq:vstartermone} \\
    &\;\;\; + \frac{2\tau_n}{K_c^2 - K_c} \sum_i \sum_{j \ne i} v^*_i(p)^T v^*_j(p) \\
    &\;\;\; + \frac{\tau_d}{K_c} \sum_i \left[ v^*_i(p)^T n(p,v^*_i(p)) + 1 \right]^2
\end{align}
where the 
first sum encourages obtaining the direction with minimum depth that is visible, 
the second prevents collapse of the candidates to a single direction (e.g., local minima), and 
the third encodes alignment with the local surface normals (similar to the directed Eikonal regularizer; see Appendix \ref{appendix:vstarsurfacenormals}).
Note that the first sum (Eq.\ \ref{app:eq:vstartermone}) includes a $d$ term, which linearly penalizes longer distances, and a $-\xi$ term, which pushes the visibility (i.e., probability of surface existence in the direction $v_i^*(p)$, from position $p$) to be high.
We use $K_c = 5$, $\tau_n = 5\times 10^{-3}$, and $\tau_d = 0.1$.
Optimization was run for $10^4$ iterations
using Adam 
(LR: $10^{-4}$; $\beta_1 = 0.9$, $\beta_2 = 0.999$).
Each update used 4096 points, 
    uniformly drawn from the bounding volume 
    ($p\sim\mathcal{U}[\mathcal{B}]$).

\subsection{Surface Normals of $v^*$}
\label{appendix:vstarsurfacenormals}
Let $v^*$ be defined as in \S\ref{sec:app:udfextract} and choose $p$ such that $v^*(p)$ is not multivalued (i.e., neither on $S$ nor on the medial surface of $S$).
Recall that, by definition, 
\begin{equation}
\mathrm{UDF}(p) = d(p,v^*(p)) \label{udf:1}
\end{equation}
and
\begin{equation}
n(p,v) = n(q(p,v)) = \frac{\nabla_p d(p,v)^T}{||\nabla_p d(p,v)||}
\end{equation}
where $n(p,v)^T v < 0$ (see Property \hyperref[property2]{II}).
Then, by Eq.\ \ref{udf:1}, 
\begin{align*}
\nabla_p \mathrm{UDF}(p) 
&= \nabla_p d(p,v^*(p)) \\
&= ||\nabla_p d(p,v^*(p))|| \,
    \frac{\nabla_p d(p,v^*(p))}{||\nabla_p d(p,v^*(p))||} \\
&= ||\nabla_p \mathrm{UDF}(p)||\; n(p,v^*(p))^T  \\
&= n(p,v^*(p))^T,
\end{align*}
via the S/UDF Eikonal equation in the last step.

Finally, by the directed Eikonal property (\hyperref[property1]{I}),
we have:
\begin{equation}
    \nabla_p \mathrm{UDF}(p) v^*(p) =
    \nabla_p d(p,v^*(p)) v^*(p) =
    -1,
\end{equation}
meaning $\nabla_p \mathrm{UDF}(p) = -v^*(p)^T$.
Indeed, recall that the gradient of a UDF has 
(1) unit norm (i.e., satisfies the Eikonal equation) and 
(2) points away from the closest point on $S$,
which is precisely the definition of $-v^*(p)$.

\begin{figure*}
    \adjincludegraphics[height=\aalh,trim={ {\ach\width} {\cuthch\height} {\ach\width}  {\cuthch\height}},clip]{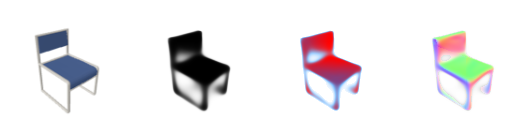}
    \adjincludegraphics[height=\alh,trim={ {\cch\width} {\cuthch\height} {\cch\width}  {\cuthch\height}},clip]{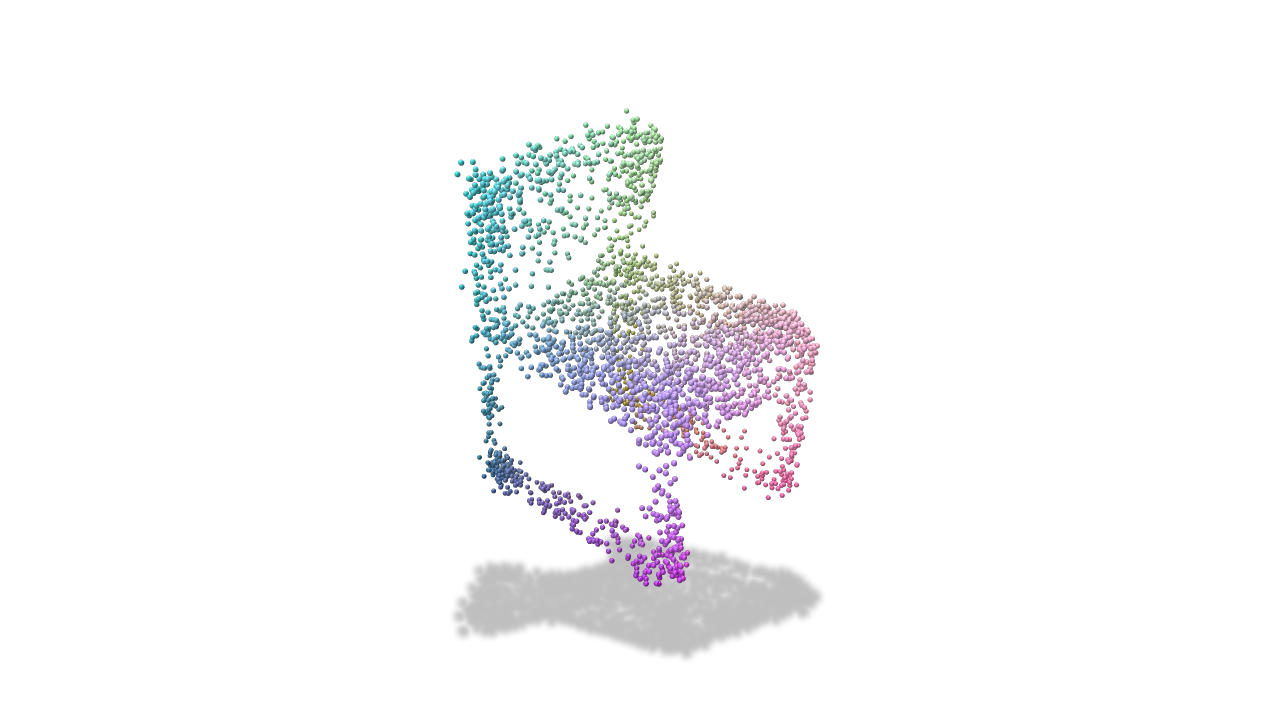}\hfill
    \adjincludegraphics[height=\alh,trim={ {\cch\width} {\cuthch\height} {\cch\width}  {\cuthch\height}},clip]{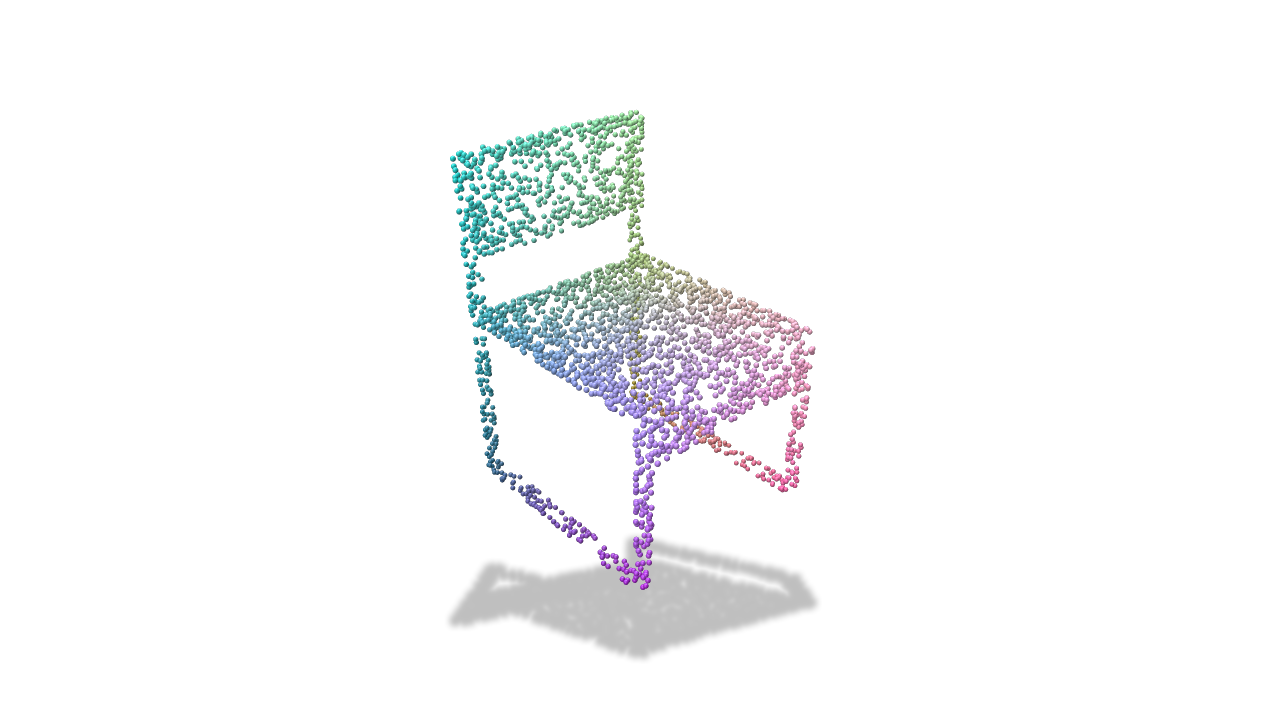}\hfill %
    \adjincludegraphics[height=\aalh,trim={ {\ach\width} {\cuthch\height} {\ach\width}  {\cuthch\height}},clip]{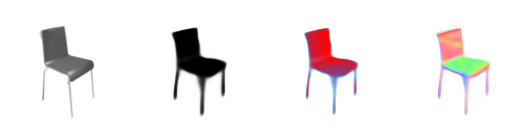}
    \adjincludegraphics[height=\alh,trim={ {\cch\width} {\cuthch\height} {\cch\width}  {\cuthch\height}},clip]{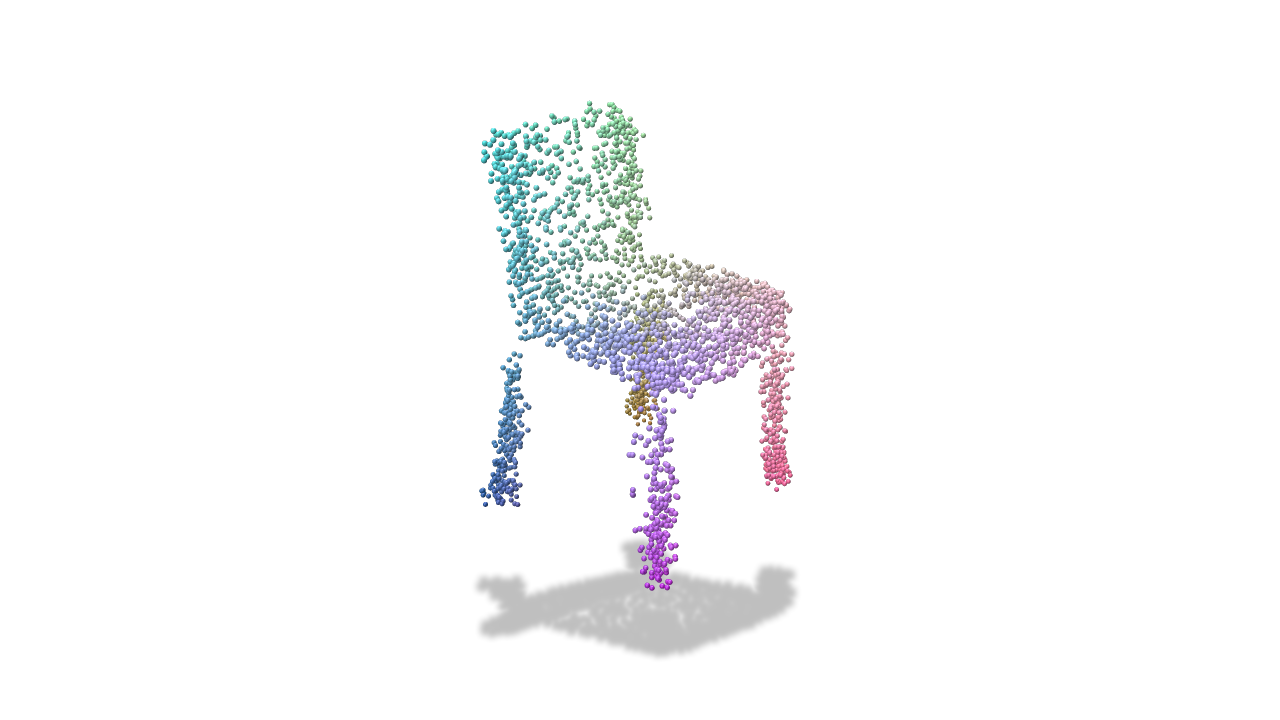}\hfill
    \adjincludegraphics[height=\alh,trim={ {\cch\width} {\cuthch\height} {\cch\width}  {\cuthch\height}},clip]{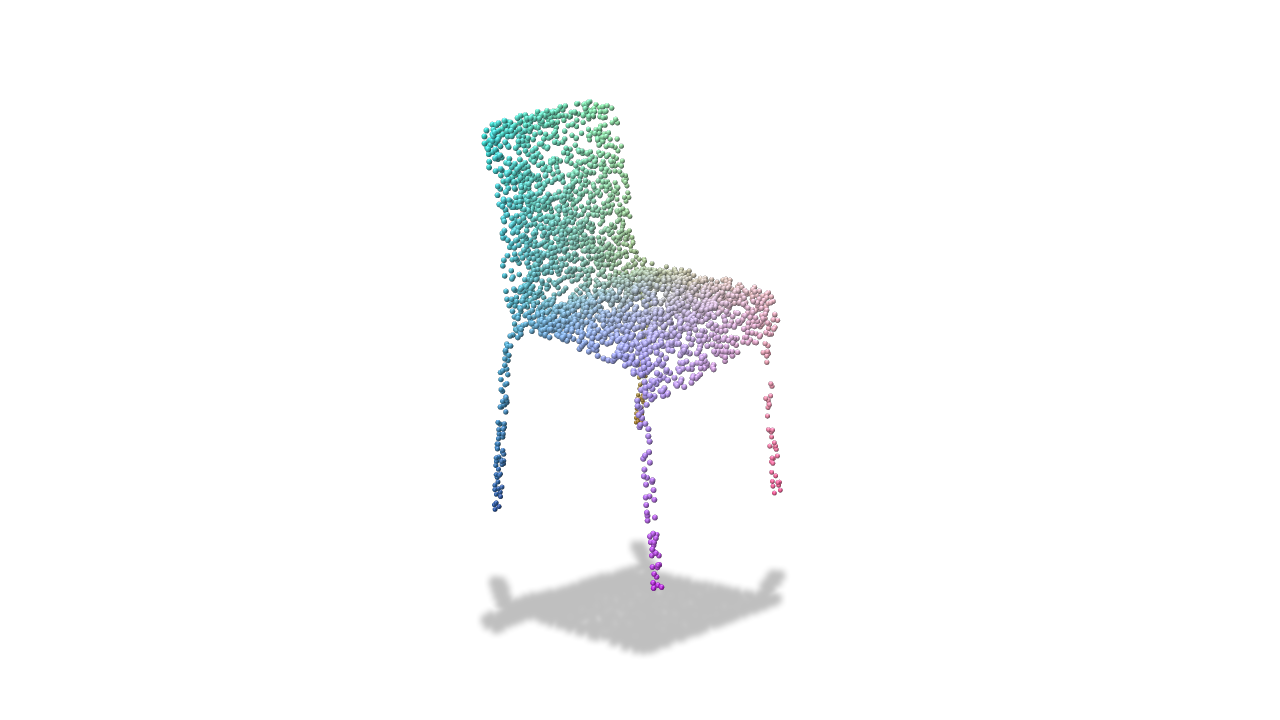} \\
    \adjincludegraphics[height=\aalh,trim={ {\ach\width} {\cuthch\height} {\ach\width}  {\cuthch\height}},clip]{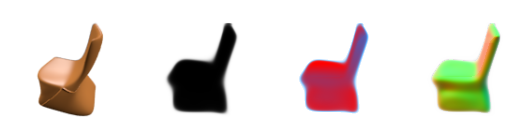}
    \adjincludegraphics[height=\alh,trim={ {\cch\width} {\cuthch\height} {\cch\width}  {\cuthch\height}},clip]{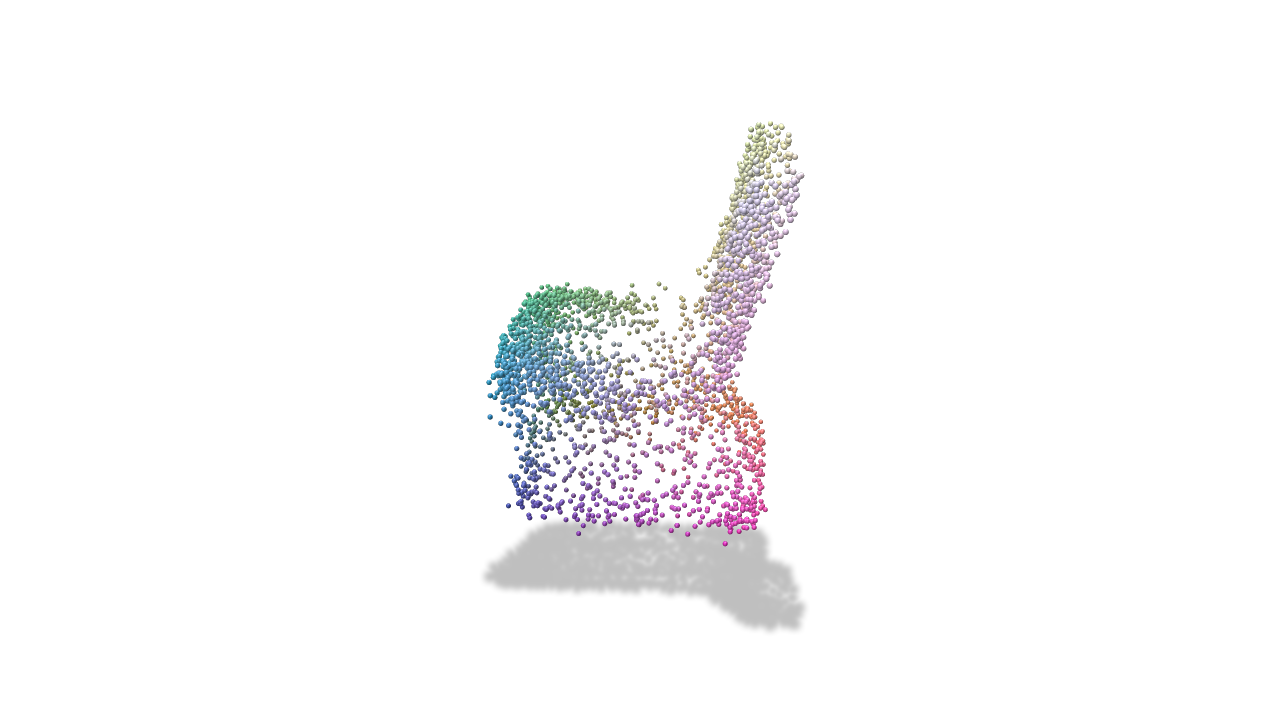}\hfill
    \adjincludegraphics[height=\alh,trim={ {\cch\width} {\cuthch\height} {\cch\width}  {\cuthch\height}},clip]{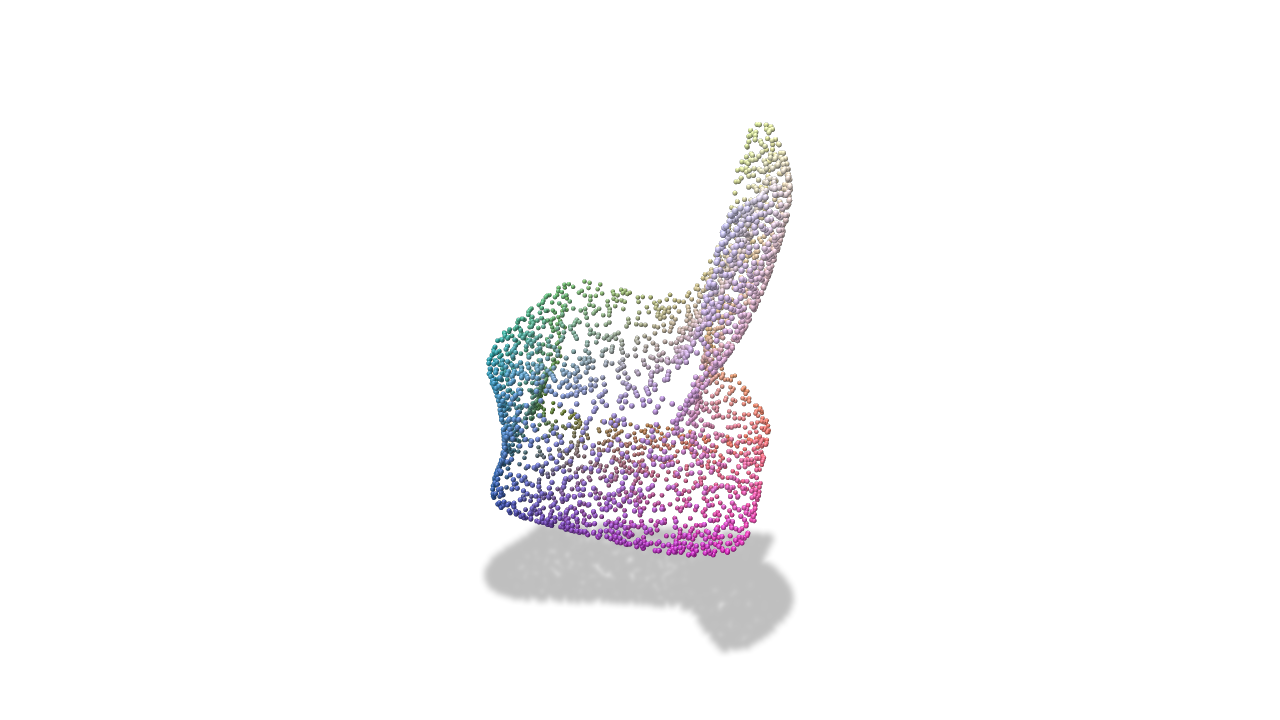}\hfill %
    \adjincludegraphics[height=\aalh,trim={ {\ach\width} {\cuthch\height} {\ach\width}  {\cuthch\height}},clip]{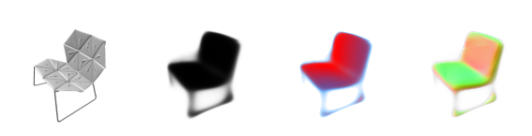}
    \adjincludegraphics[height=\alh,trim={ {\cch\width} {\cuthch\height} {\cch\width}  {\cuthch\height}},clip]{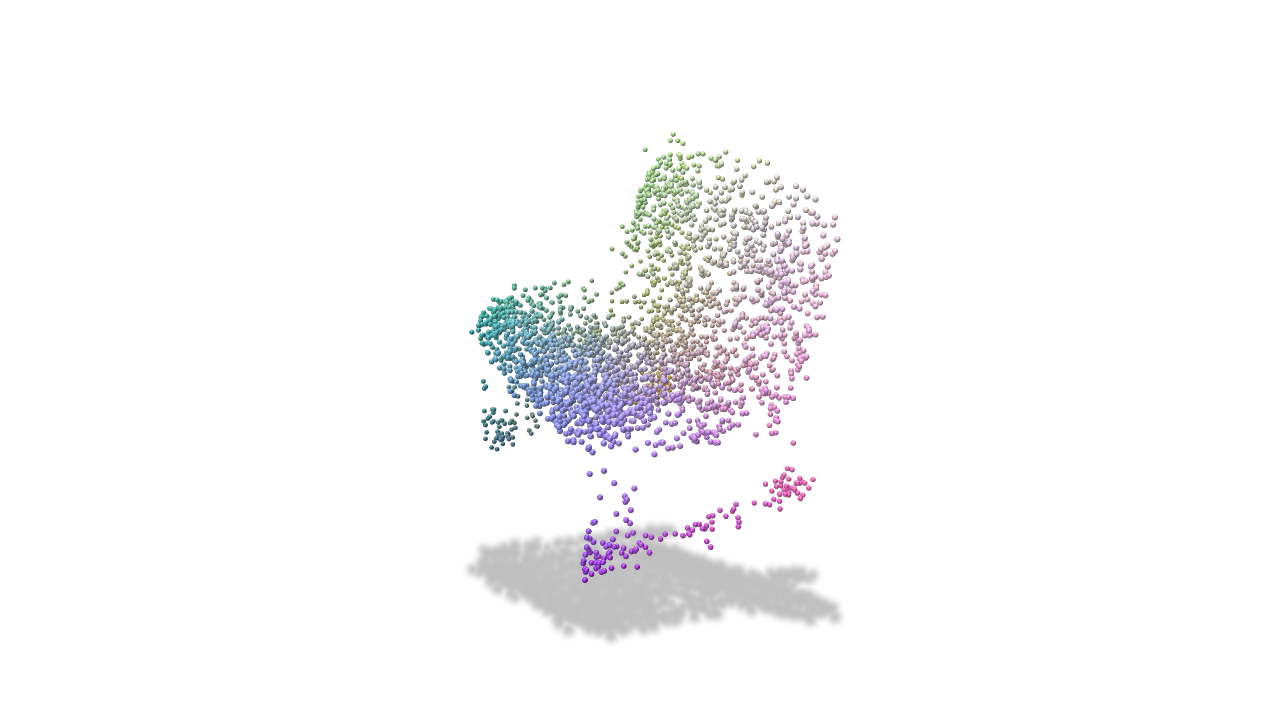}\hfill
    \adjincludegraphics[height=\alh,trim={ {\cch\width} {\cuthch\height} {\cch\width}  {\cuthch\height}},clip]{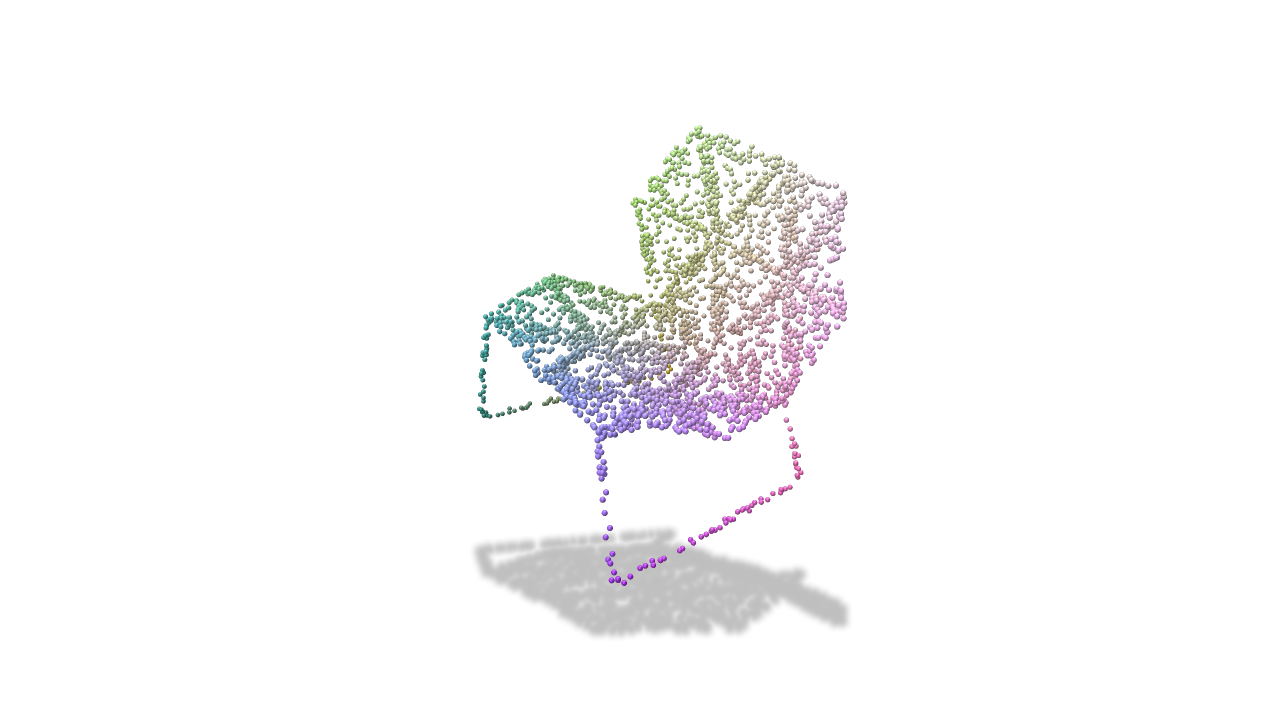} \\
    \adjincludegraphics[height=\aalh,trim={ {\ach\width} {\cuthch\height} {\ach\width}  {\cuthch\height}},clip]{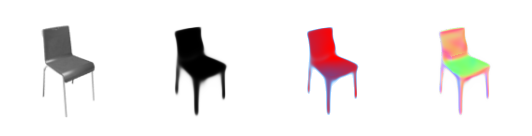}
    \adjincludegraphics[height=\alh,trim={ {\cch\width} {\cuthch\height} {\cch\width}  {\cuthch\height}},clip]{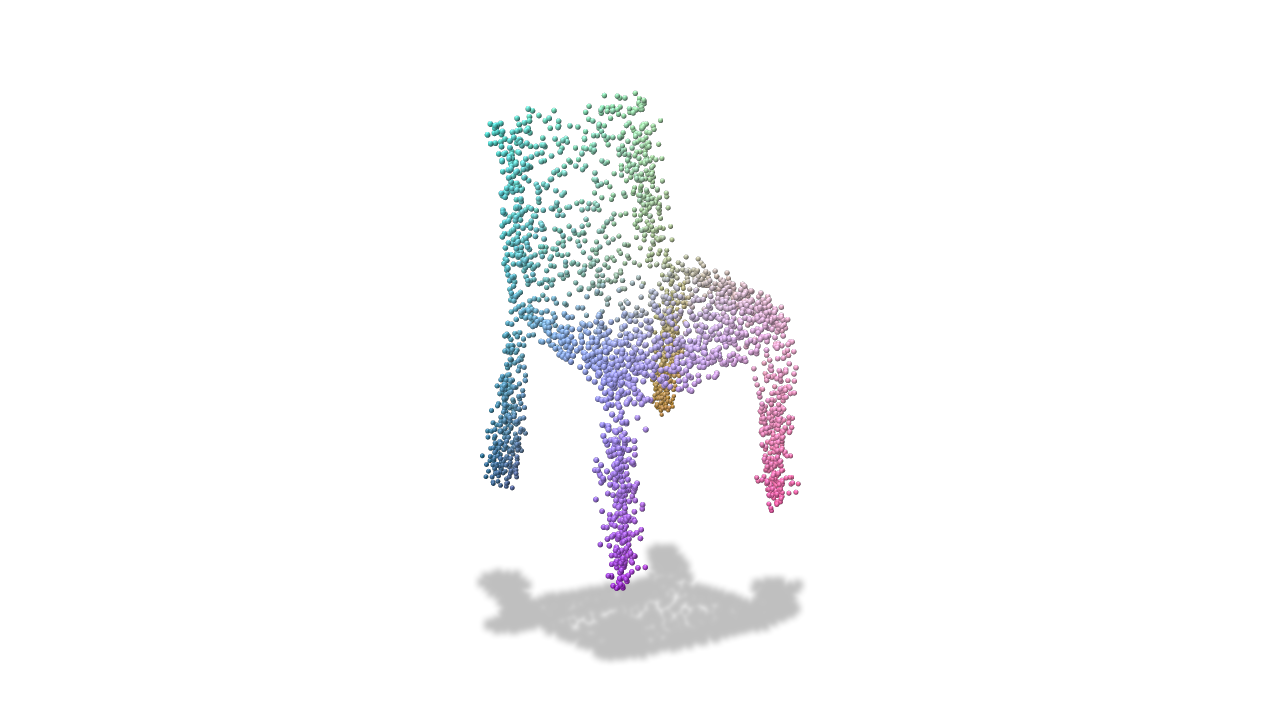}\hfill
    \adjincludegraphics[height=\alh,trim={ {\cch\width} {\cuthch\height} {\cch\width}  {\cuthch\height}},clip]{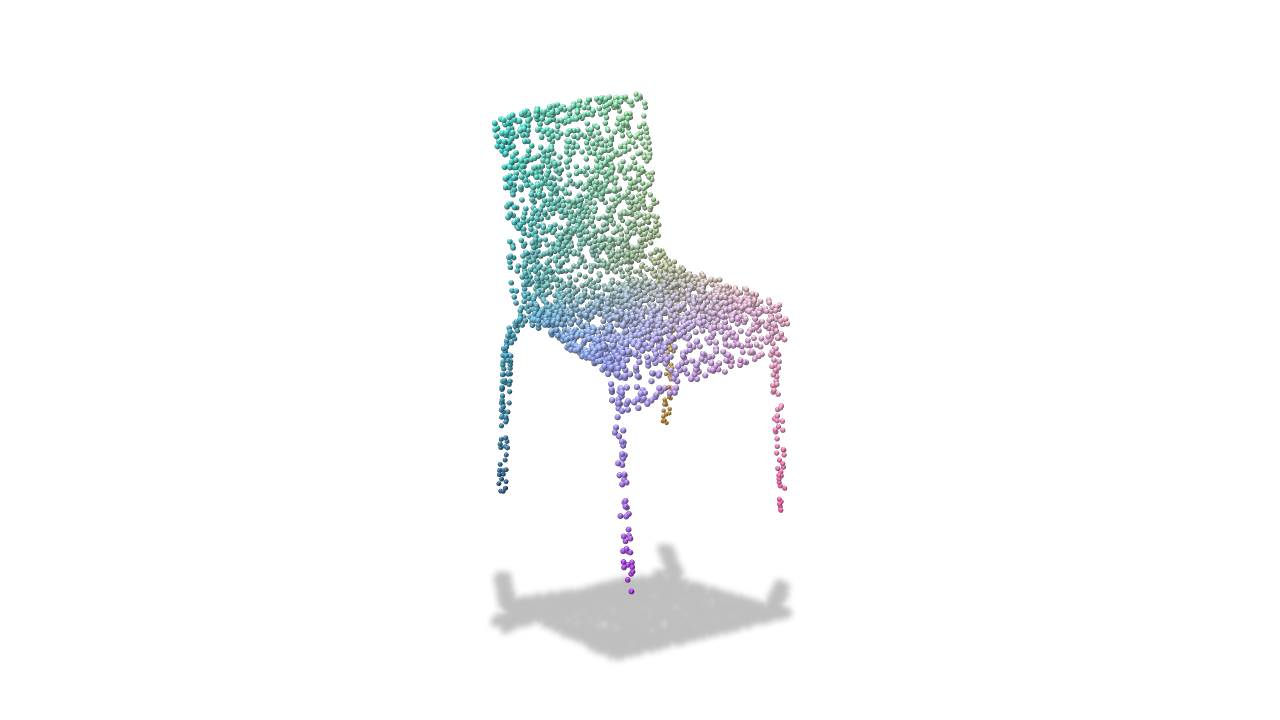}\hfill %
    \adjincludegraphics[height=\aalh,trim={ {\ach\width} {\cuthch\height} {\ach\width}  {\cuthch\height}},clip]{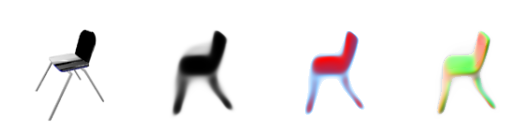}
    \adjincludegraphics[height=\alh,trim={ {\cch\width} {\cuthch\height} {\cch\width}  {\cuthch\height}},clip]{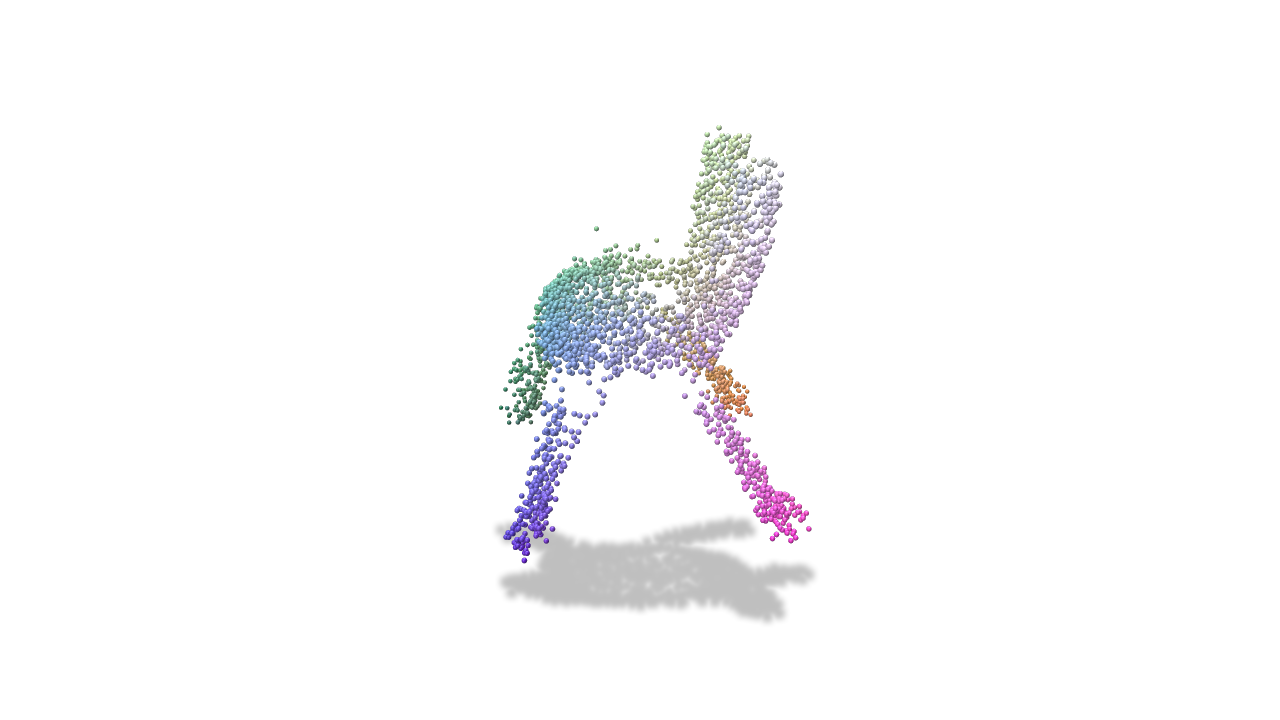}\hfill
    \adjincludegraphics[height=\alh,trim={ {\cch\width} {\cuthch\height} {\cch\width}  {\cuthch\height}},clip]{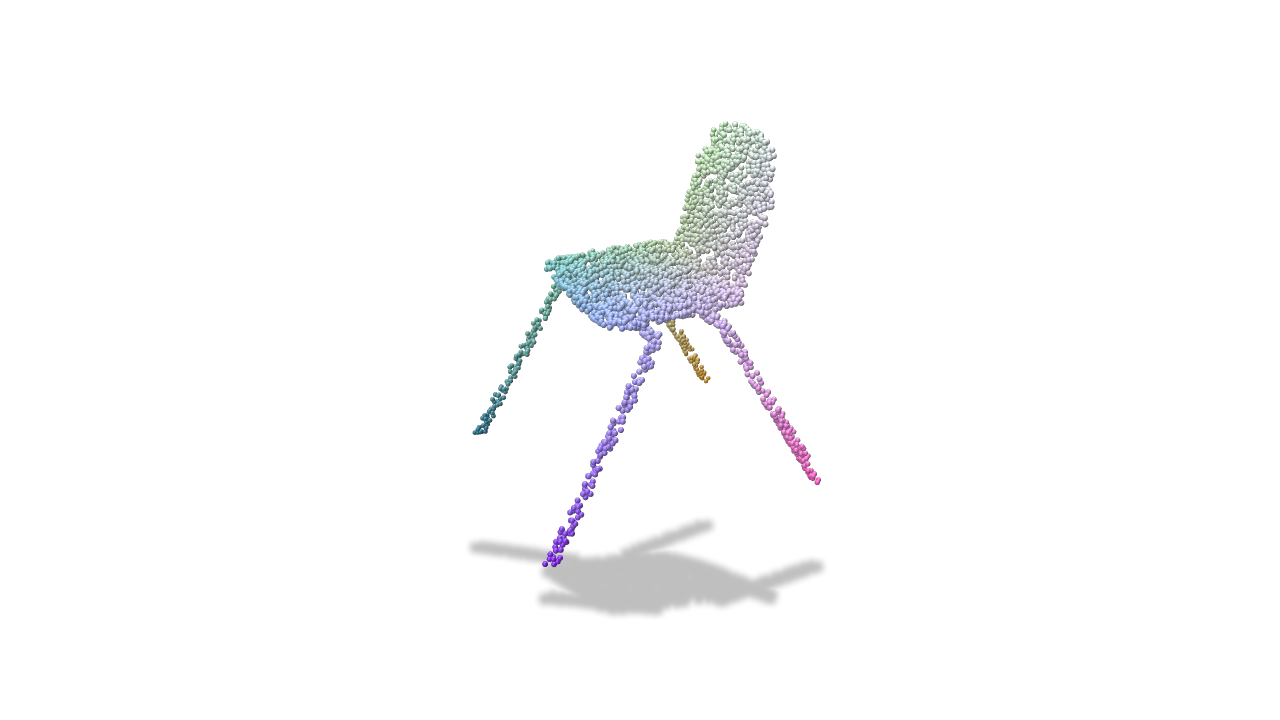} \\
    \adjincludegraphics[height=\aalh,trim={ {\ach\width} {\cuthch\height} {\ach\width}  {\cuthch\height}},clip]{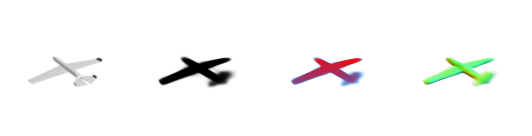}
    \adjincludegraphics[height=\alh,trim={ {\cch\width} {\cuthch\height} {\cch\width}  {\cuthch\height}},clip]{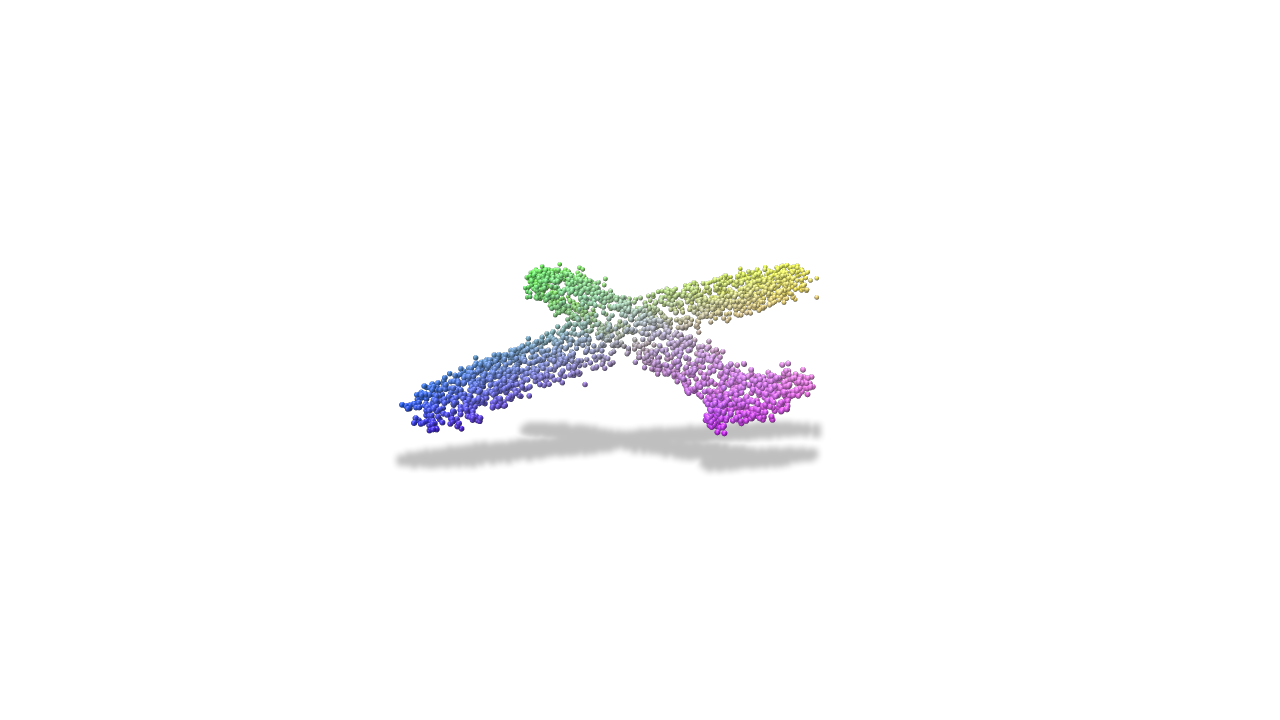}\hfill
    \adjincludegraphics[height=\alh,trim={ {\cch\width} {\cuthch\height} {\cch\width}  {\cuthch\height}},clip]{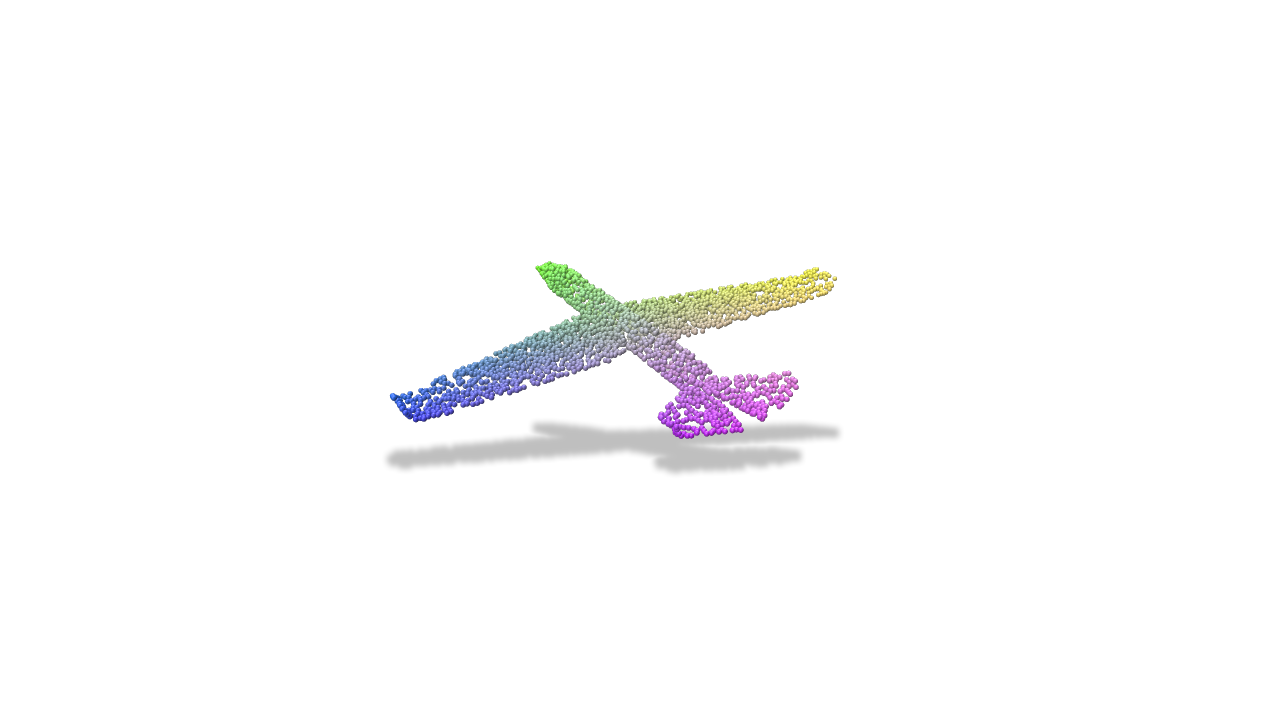}\hfill %
    \adjincludegraphics[height=\aalh,trim={ {\ach\width} {\cuthch\height} {\ach\width}  {\cuthch\height}},clip]{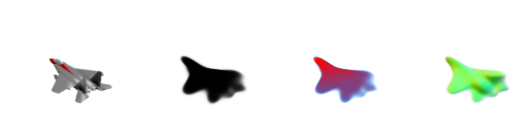}
    \adjincludegraphics[height=\alh,trim={ {\cch\width} {\cuthch\height} {\cch\width}  {\cuthch\height}},clip]{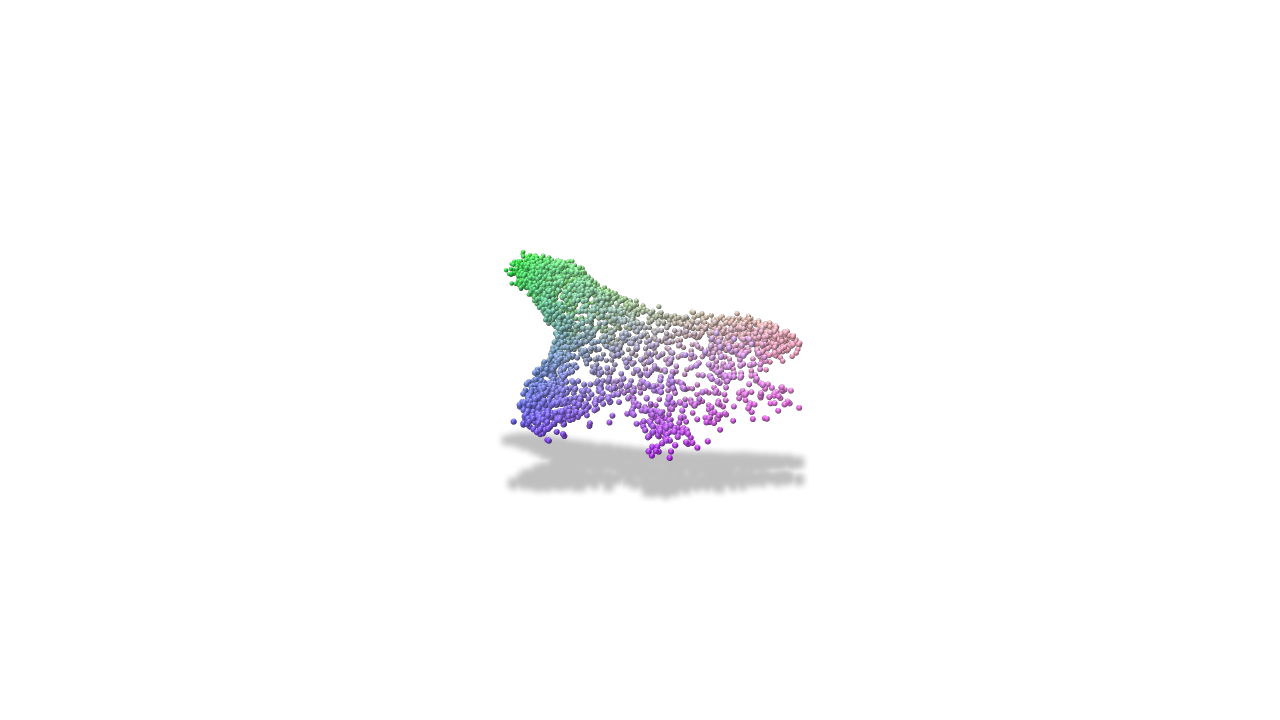}\hfill
    \adjincludegraphics[height=\alh,trim={ {\cch\width} {\cuthch\height} {\cch\width}  {\cuthch\height}},clip]{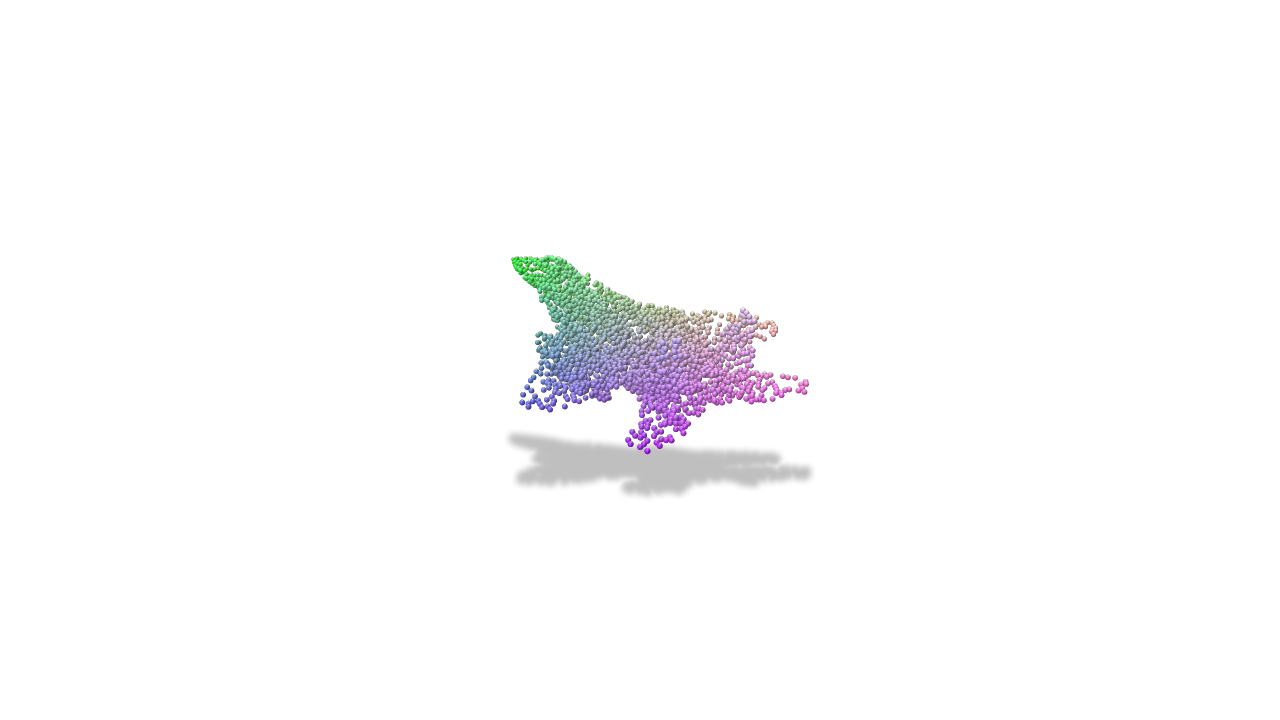} \\
    \adjincludegraphics[height=\aalh,trim={ {\ach\width} {\cuthch\height} {\ach\width}  {\cuthch\height}},clip]{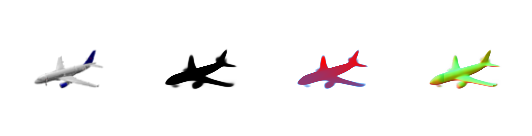}
    \adjincludegraphics[height=\alh,trim={ {\cch\width} {\cuthch\height} {\cch\width}  {\cuthch\height}},clip]{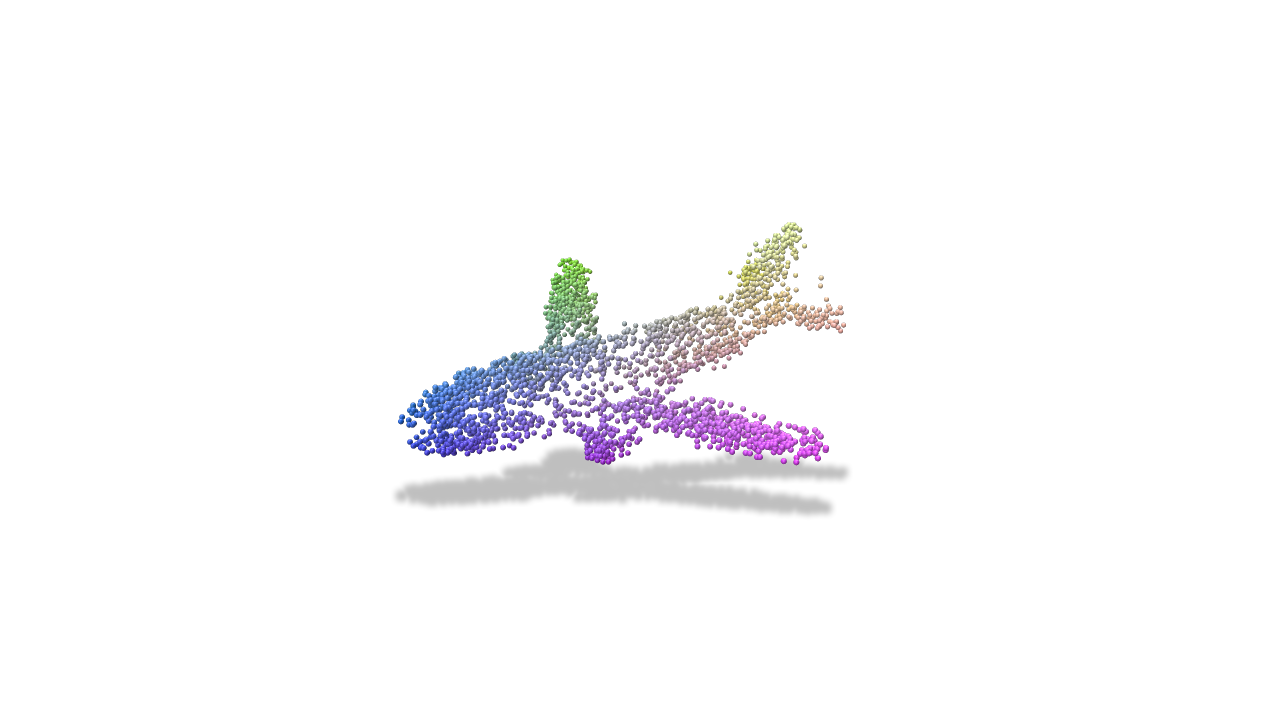}\hfill
    \adjincludegraphics[height=\alh,trim={ {\cch\width} {\cuthch\height} {\cch\width}  {\cuthch\height}},clip]{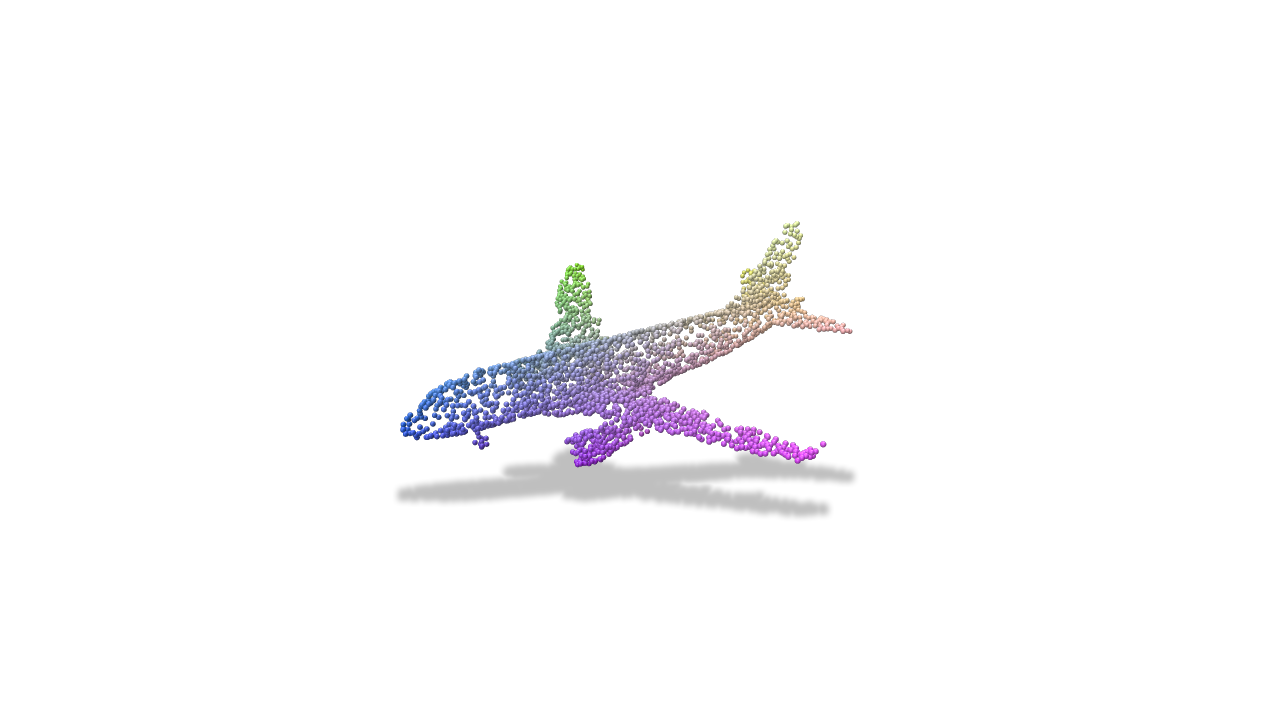}\hfill %
    \adjincludegraphics[height=\aalh,trim={ {\ach\width} {\cuthch\height} {\ach\width}  {\cuthch\height}},clip]{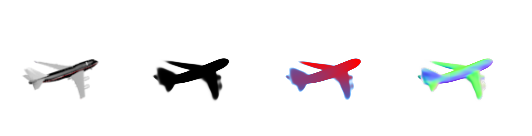}
    \adjincludegraphics[height=\alh,trim={ {\cch\width} {\cuthch\height} {\cch\width}  {\cuthch\height}},clip]{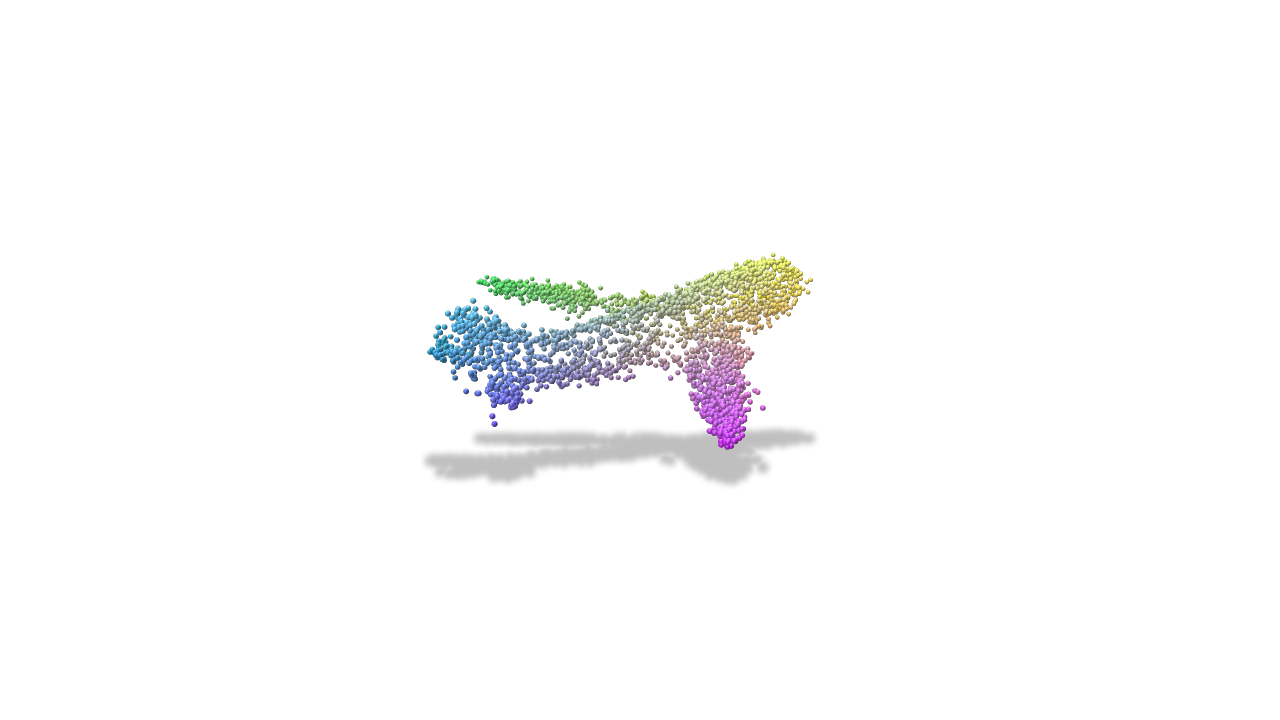}\hfill
    \adjincludegraphics[height=\alh,trim={ {\cch\width} {\cuthch\height} {\cch\width}  {\cuthch\height}},clip]{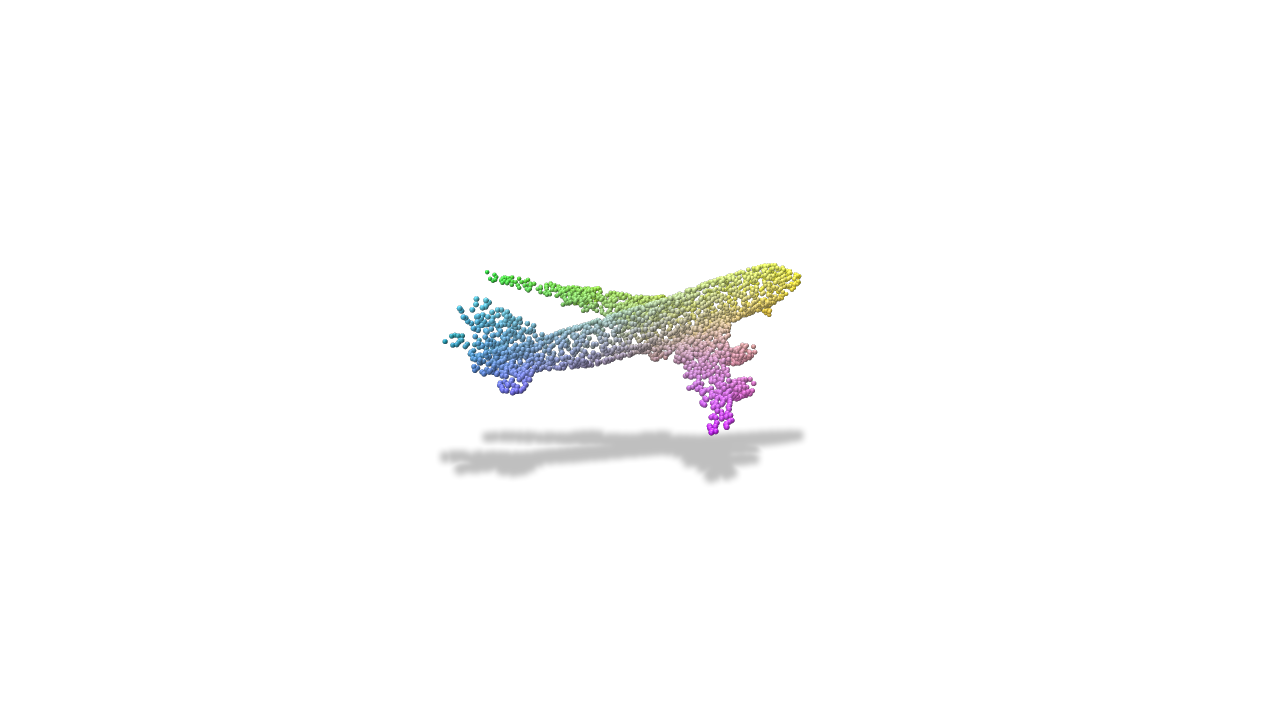} \\
    \adjincludegraphics[height=\aalh,trim={ {\ach\width} {\cuthch\height} {\ach\width}  {\cuthch\height}},clip]{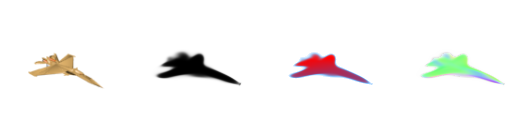}
    \adjincludegraphics[height=\alh,trim={ {\cch\width} {\cuthch\height} {\cch\width}  {\cuthch\height}},clip]{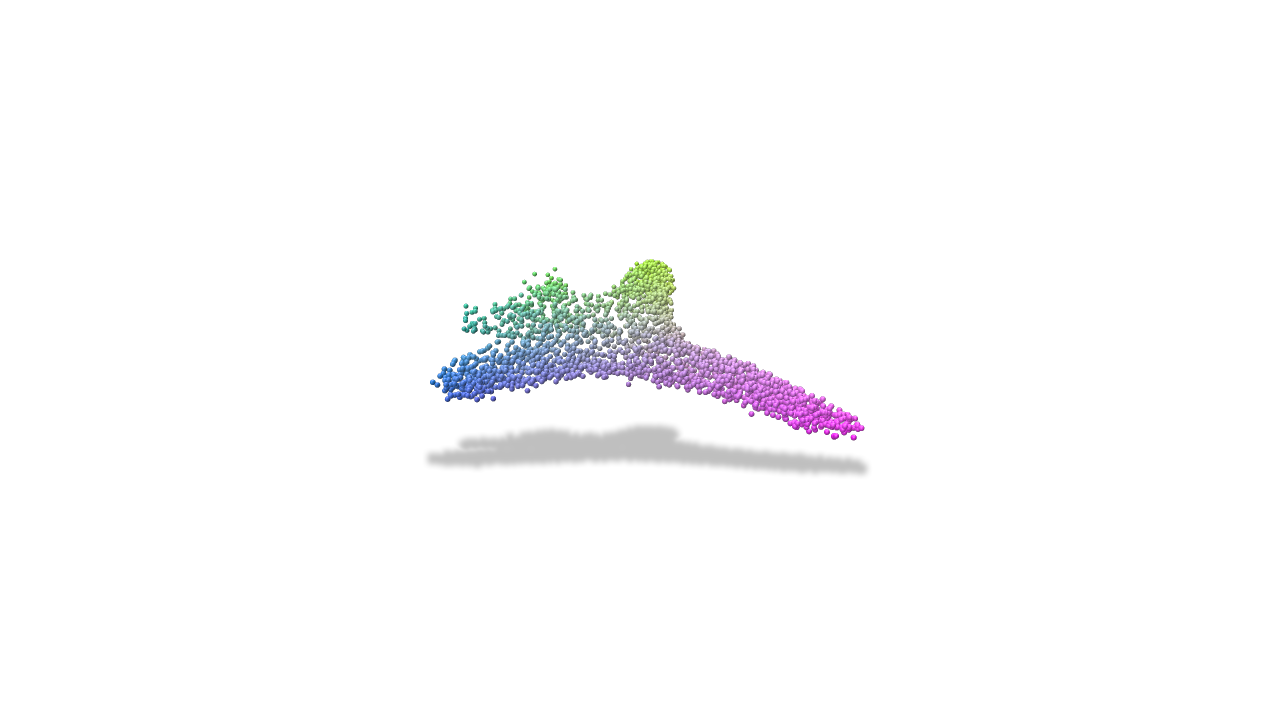}\hfill
    \adjincludegraphics[height=\alh,trim={ {\cch\width} {\cuthch\height} {\cch\width}  {\cuthch\height}},clip]{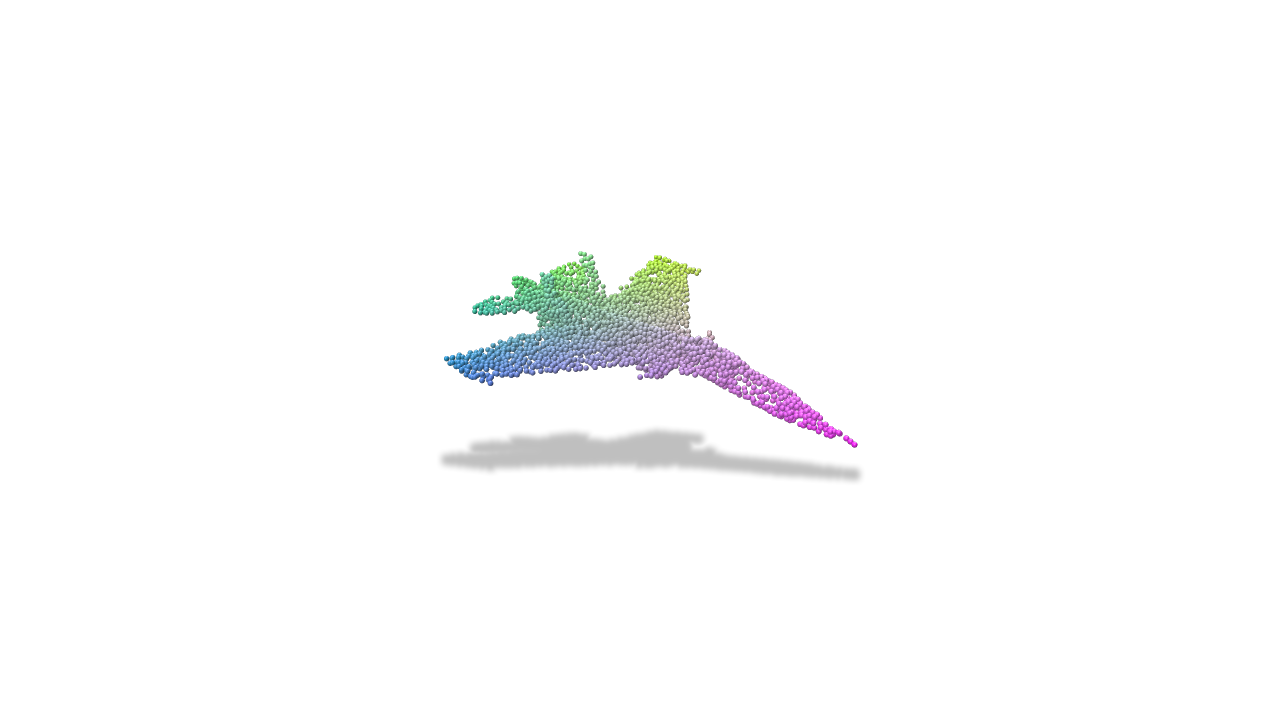}\hfill %
    \adjincludegraphics[height=\aalh,trim={ {\ach\width} {\cuthch\height} {\ach\width}  {\cuthch\height}},clip]{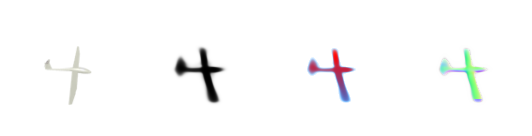}
    \adjincludegraphics[height=\alh,trim={ {\cch\width} {\cuthch\height} {\cch\width}  {\cuthch\height}},clip]{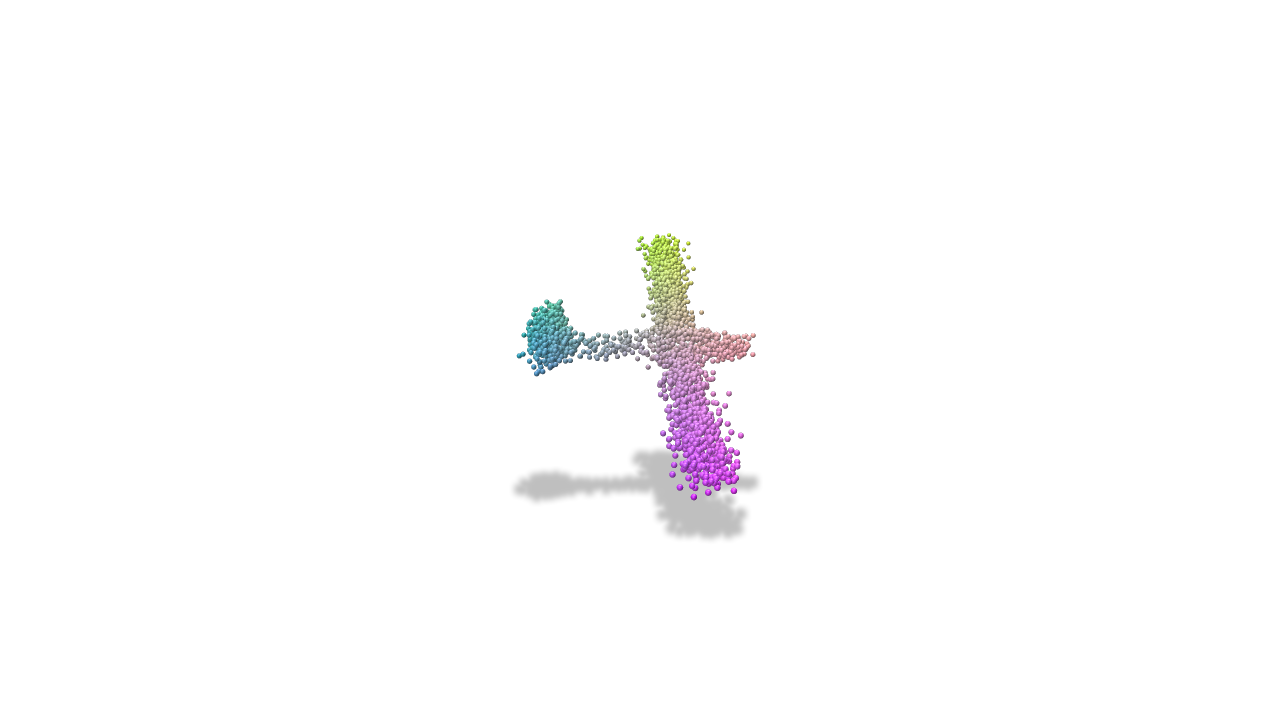}\hfill
    \adjincludegraphics[height=\alh,trim={ {\cch\width} {\cuthch\height} {\cch\width}  {\cuthch\height}},clip]{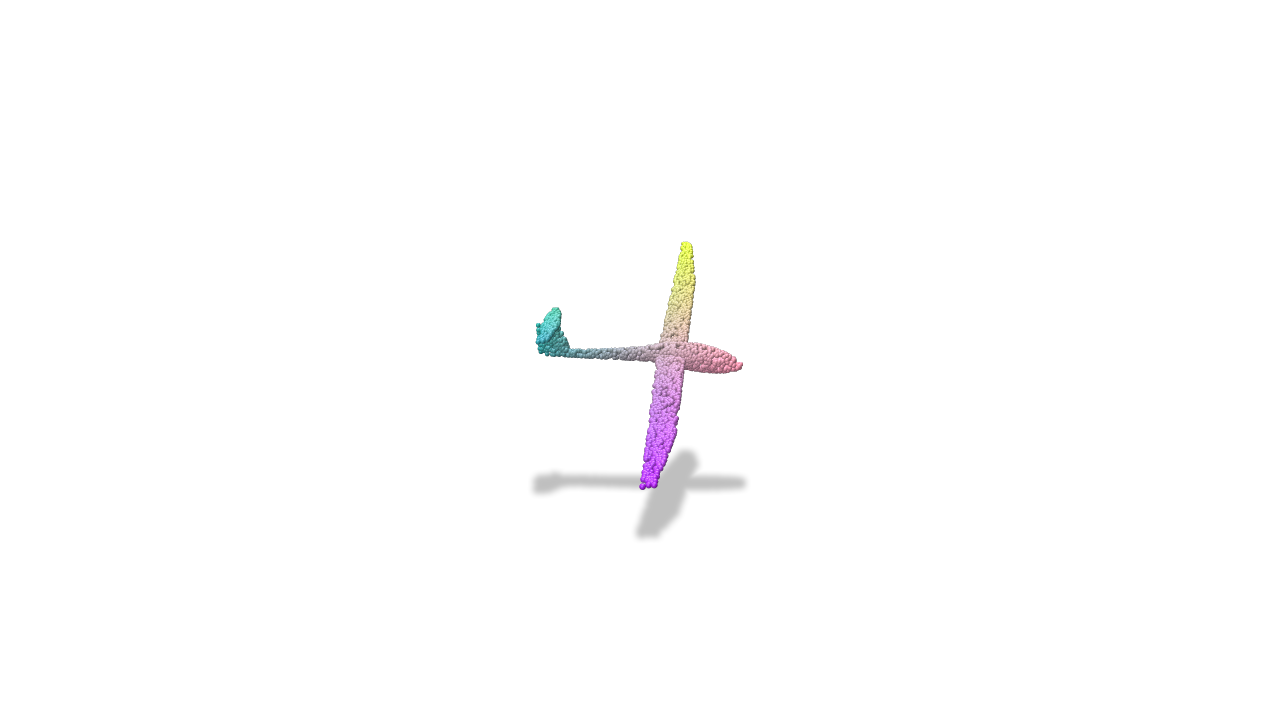} 
    \adjincludegraphics[height=\aalh,trim={ {\ach\width} {\cuthch\height} {\ach\width}  {\cuthch\height}},clip]{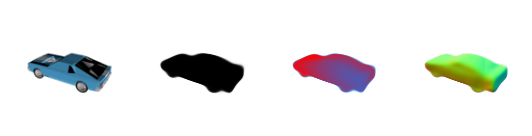}
    \adjincludegraphics[height=\alh,trim={ {\cch\width} {\cuthch\height} {\cch\width}  {\cuthch\height}},clip]{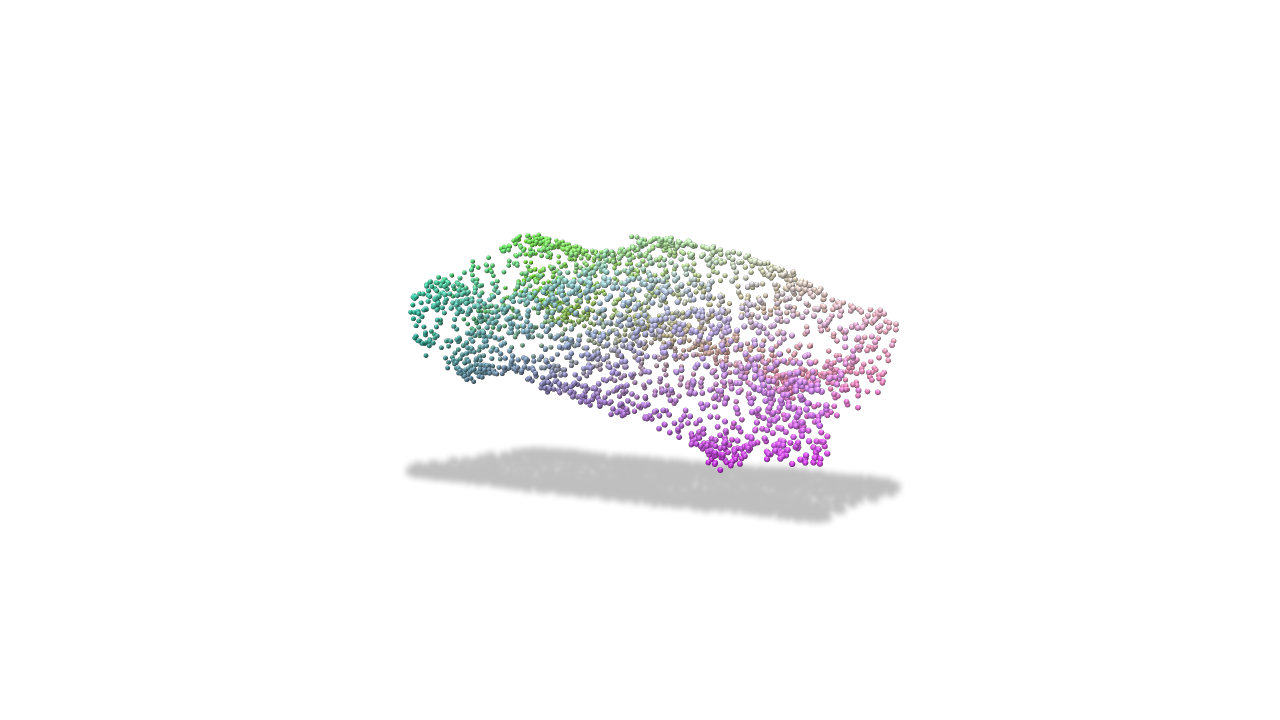}\hfill
    \adjincludegraphics[height=\alh,trim={ {\cch\width} {\cuthch\height} {\cch\width}  {\cuthch\height}},clip]{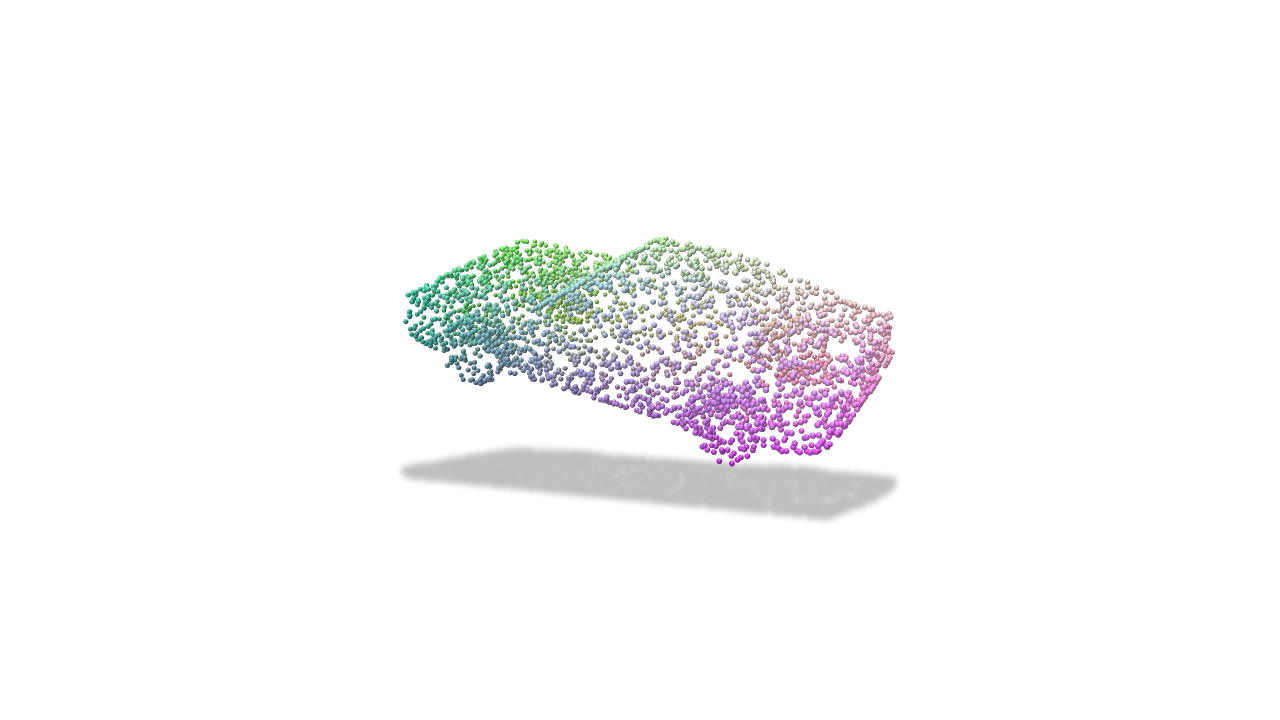}\hfill %
    \adjincludegraphics[height=\aalh,trim={ {\ach\width} {\cuthch\height} {\ach\width}  {\cuthch\height}},clip]{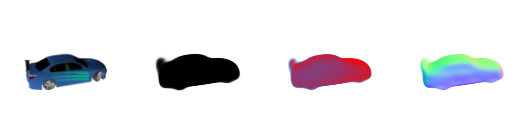}
    \adjincludegraphics[height=\alh,trim={ {\cch\width} {\cuthch\height} {\cch\width}  {\cuthch\height}},clip]{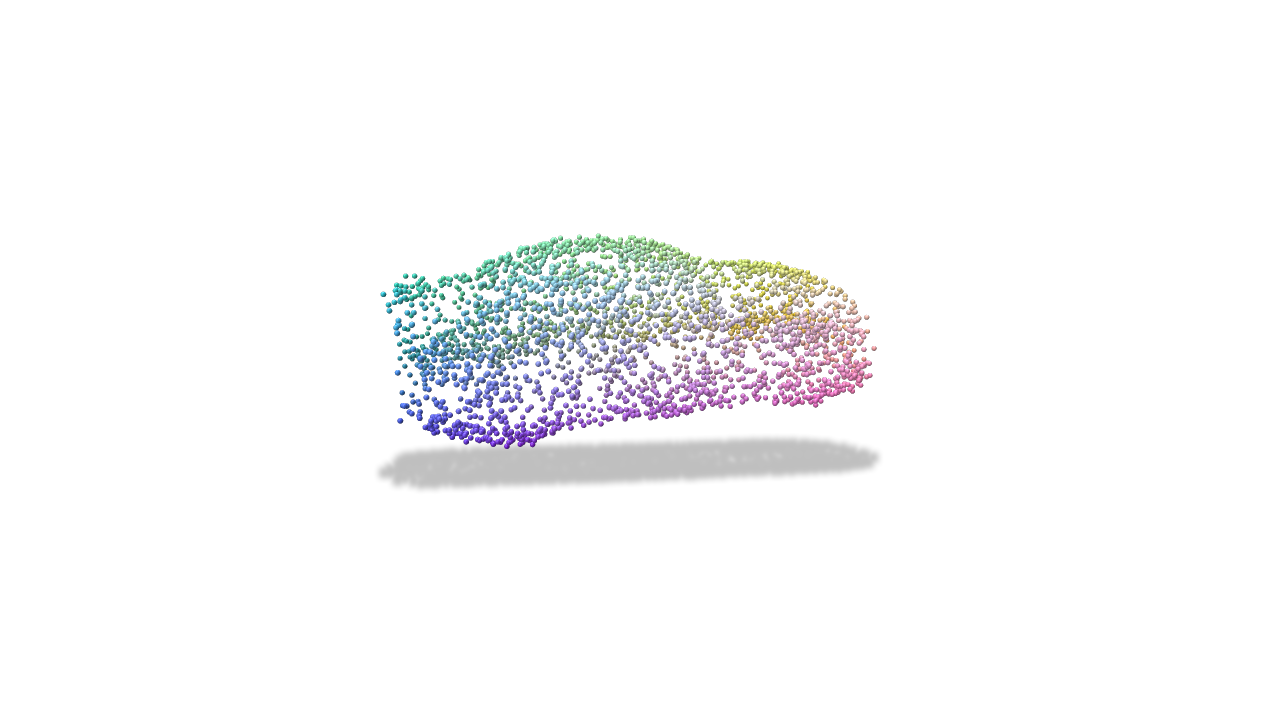}\hfill
    \adjincludegraphics[height=\alh,trim={ {\cch\width} {\cuthch\height} {\cch\width}  {\cuthch\height}},clip]{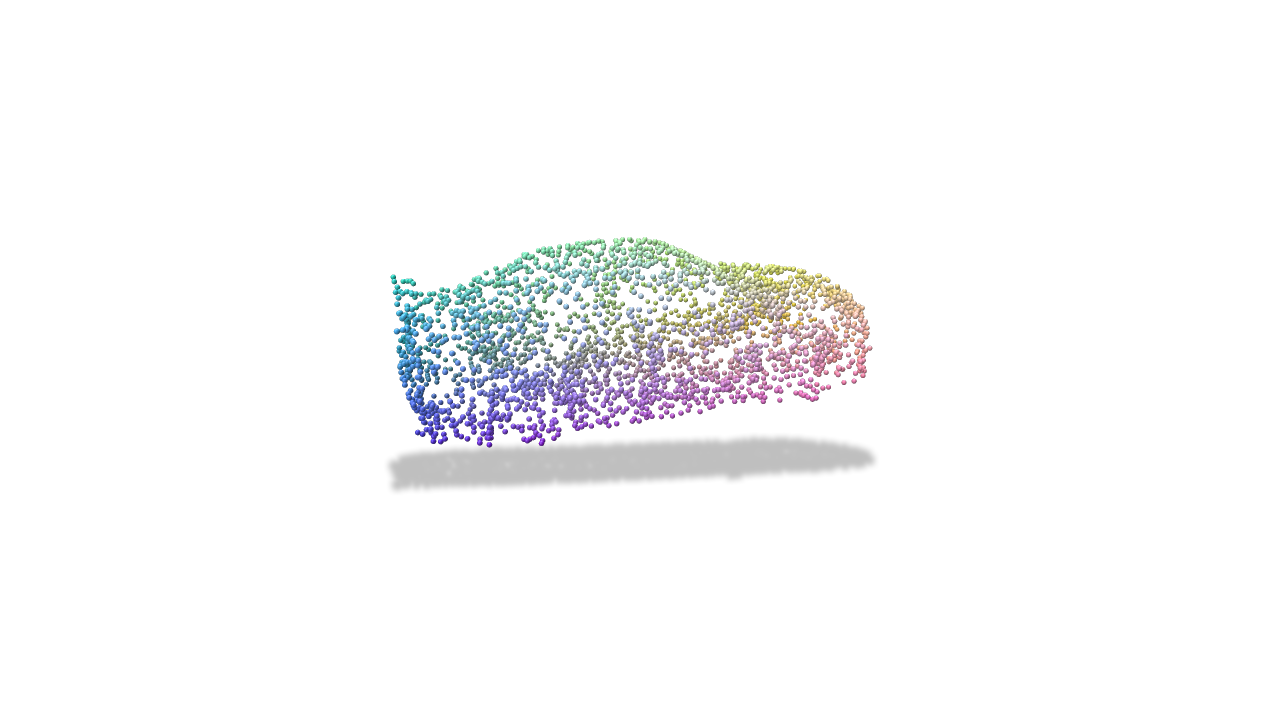} \\
    \adjincludegraphics[height=\aalh,trim={ {\ach\width} {\cuthch\height} {\ach\width}  {\cuthch\height}},clip]{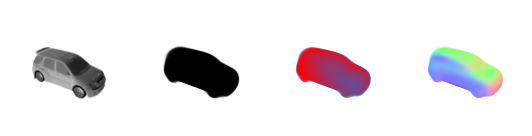}
    \adjincludegraphics[height=\alh,trim={ {\cch\width} {\cuthch\height} {\cch\width}  {\cuthch\height}},clip]{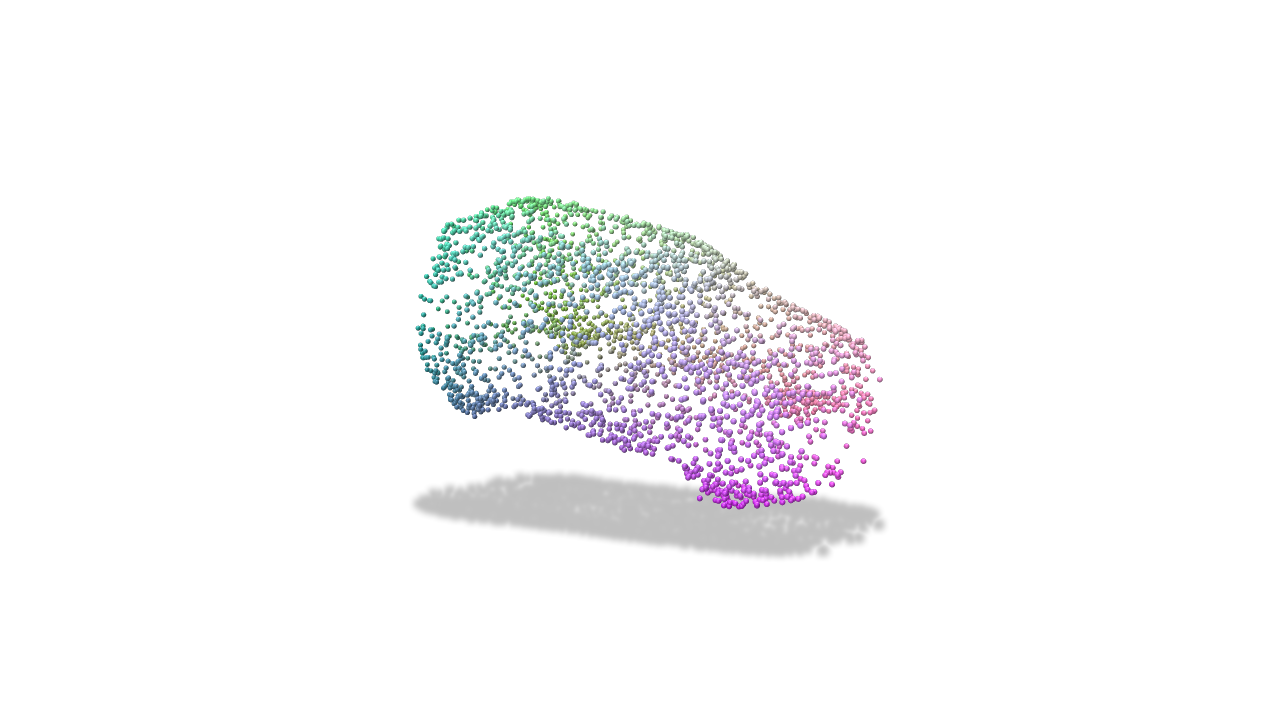}\hfill
    \adjincludegraphics[height=\alh,trim={ {\cch\width} {\cuthch\height} {\cch\width}  {\cuthch\height}},clip]{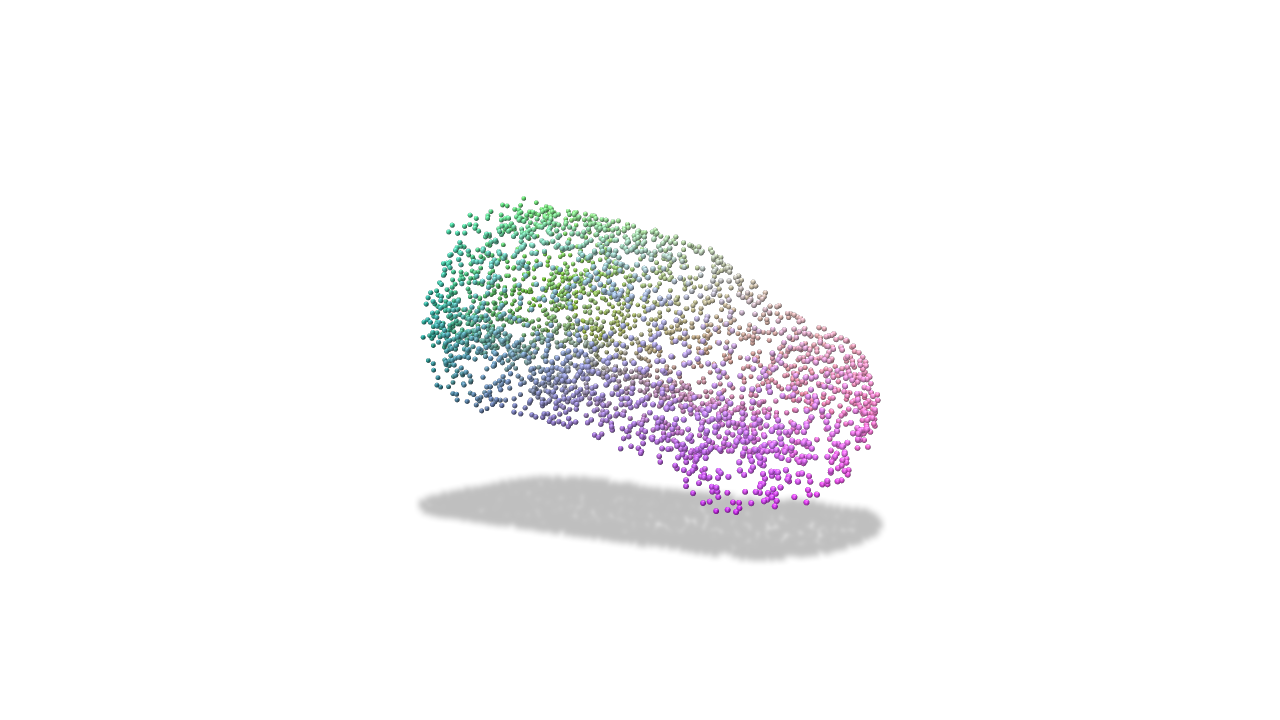}\hfill %
    \adjincludegraphics[height=\aalh,trim={ {\ach\width} {\cuthch\height} {\ach\width}  {\cuthch\height}},clip]{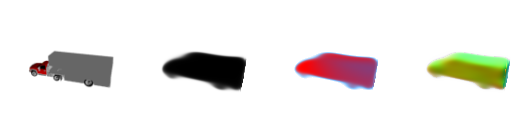}
    \adjincludegraphics[height=\alh,trim={ {\cch\width} {\cuthch\height} {\cch\width}  {\cuthch\height}},clip]{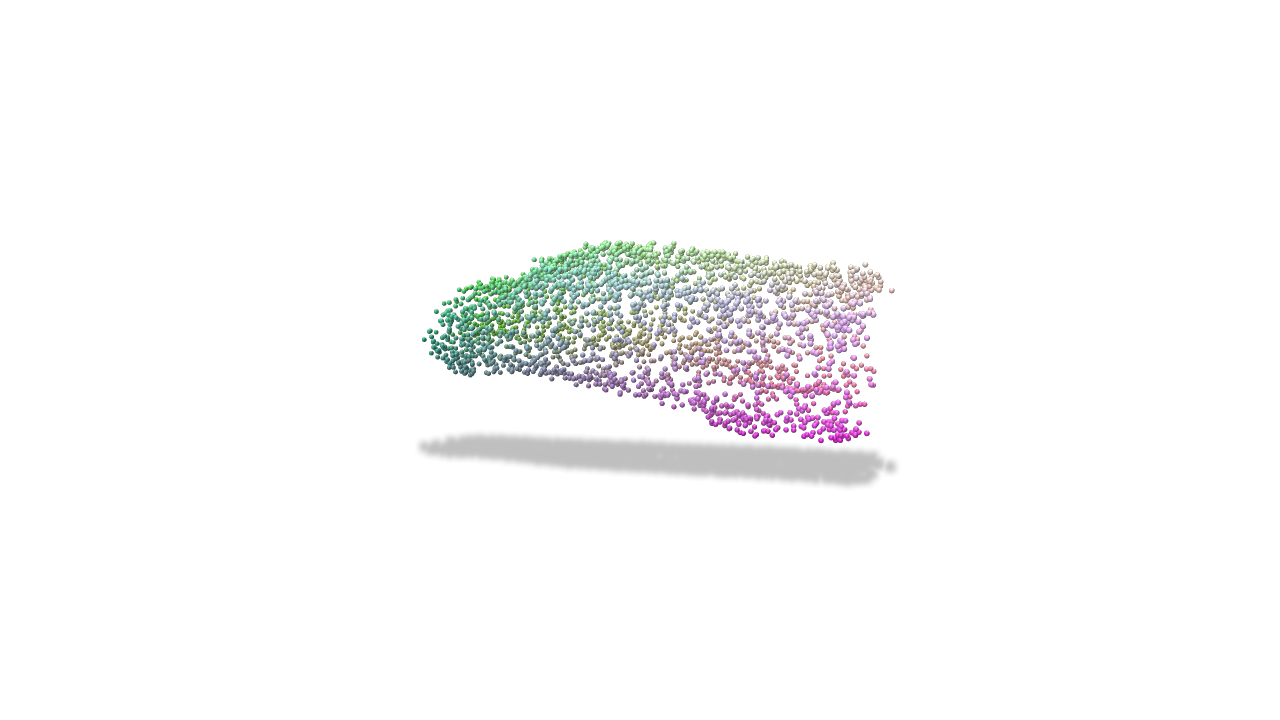}\hfill
    \adjincludegraphics[height=\alh,trim={ {\cch\width} {\cuthch\height} {\cch\width}  {\cuthch\height}},clip]{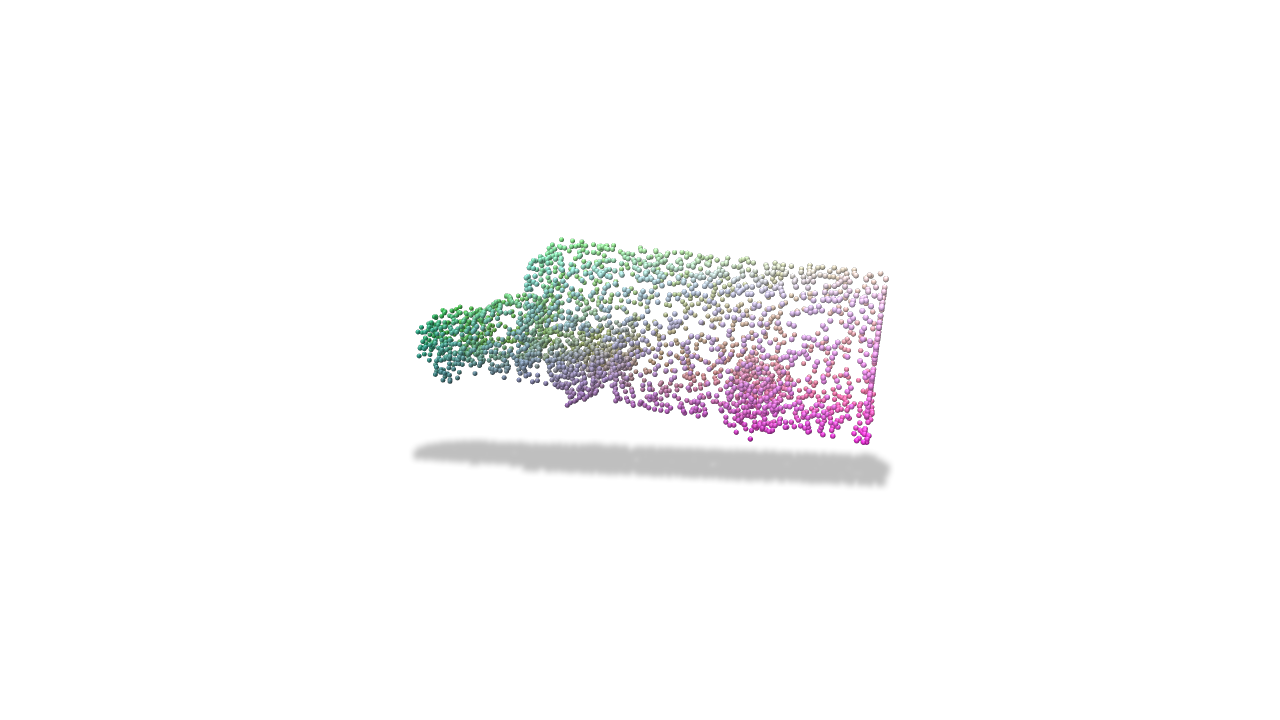} \\
    \adjincludegraphics[height=\aalh,trim={ {\ach\width} {\cuthch\height} {\ach\width}  {\cuthch\height}},clip]{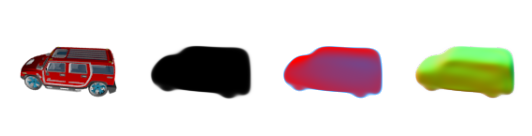}
    \adjincludegraphics[height=\alh,trim={ {\cch\width} {\cuthch\height} {\cch\width}  {\cuthch\height}},clip]{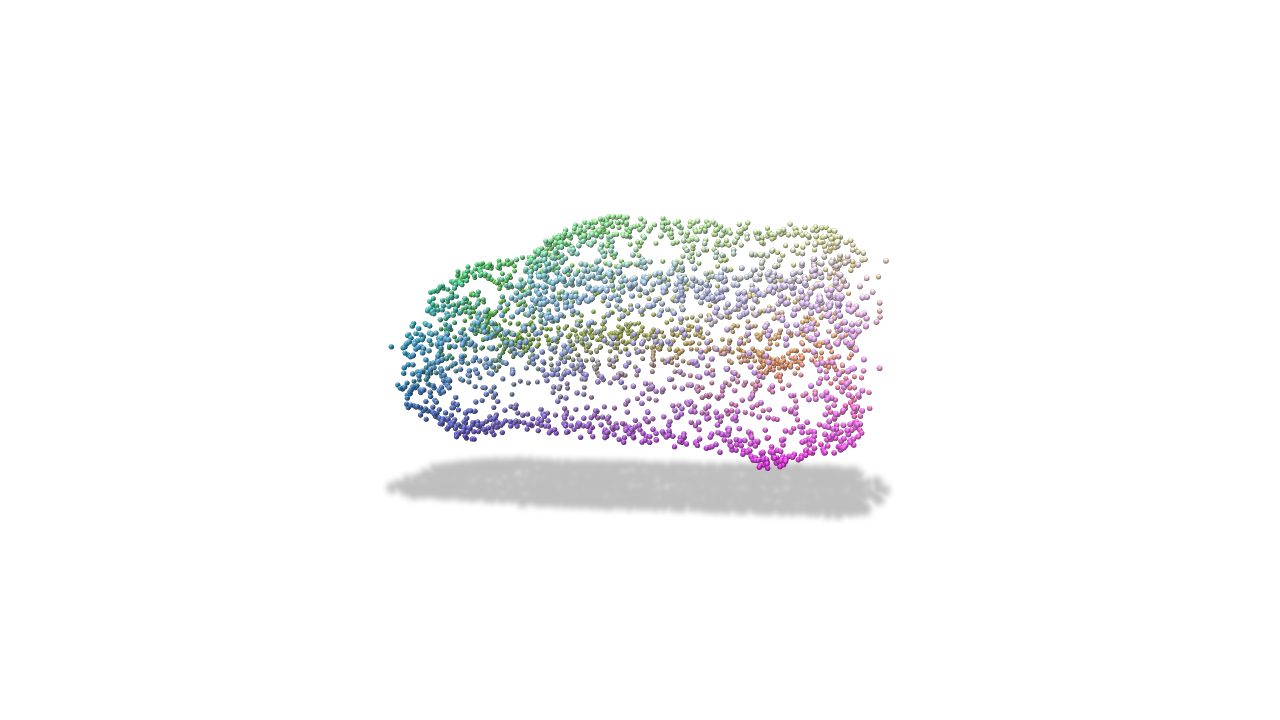}\hfill
    \adjincludegraphics[height=\alh,trim={ {\cch\width} {\cuthch\height} {\cch\width}  {\cuthch\height}},clip]{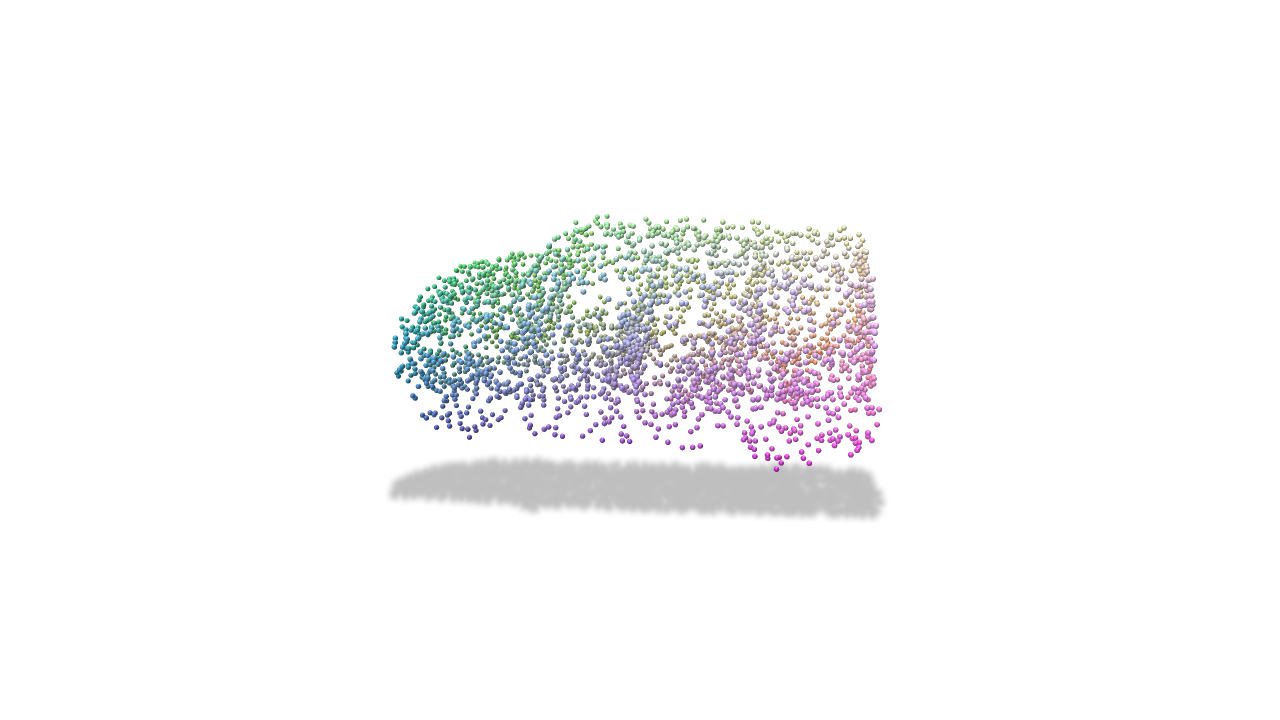}\hfill %
    \adjincludegraphics[height=\aalh,trim={ {\ach\width} {\cuthch\height} {\ach\width}  {\cuthch\height}},clip]{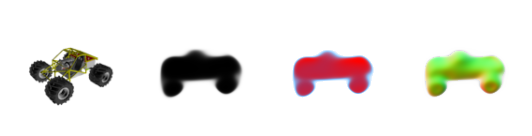}
    \adjincludegraphics[height=\alh,trim={ {\cch\width} {\cuthch\height} {\cch\width}  {\cuthch\height}},clip]{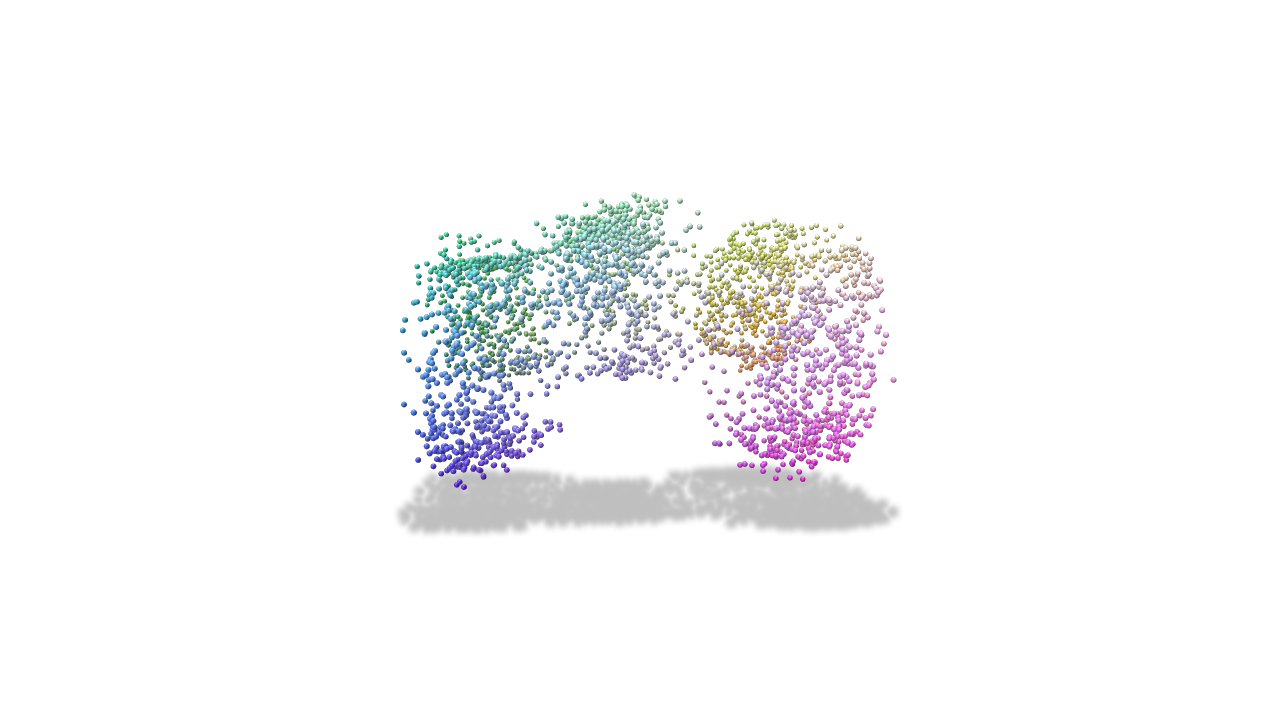}\hfill
    \adjincludegraphics[height=\alh,trim={ {\cch\width} {\cuthch\height} {\cch\width}  {\cuthch\height}},clip]{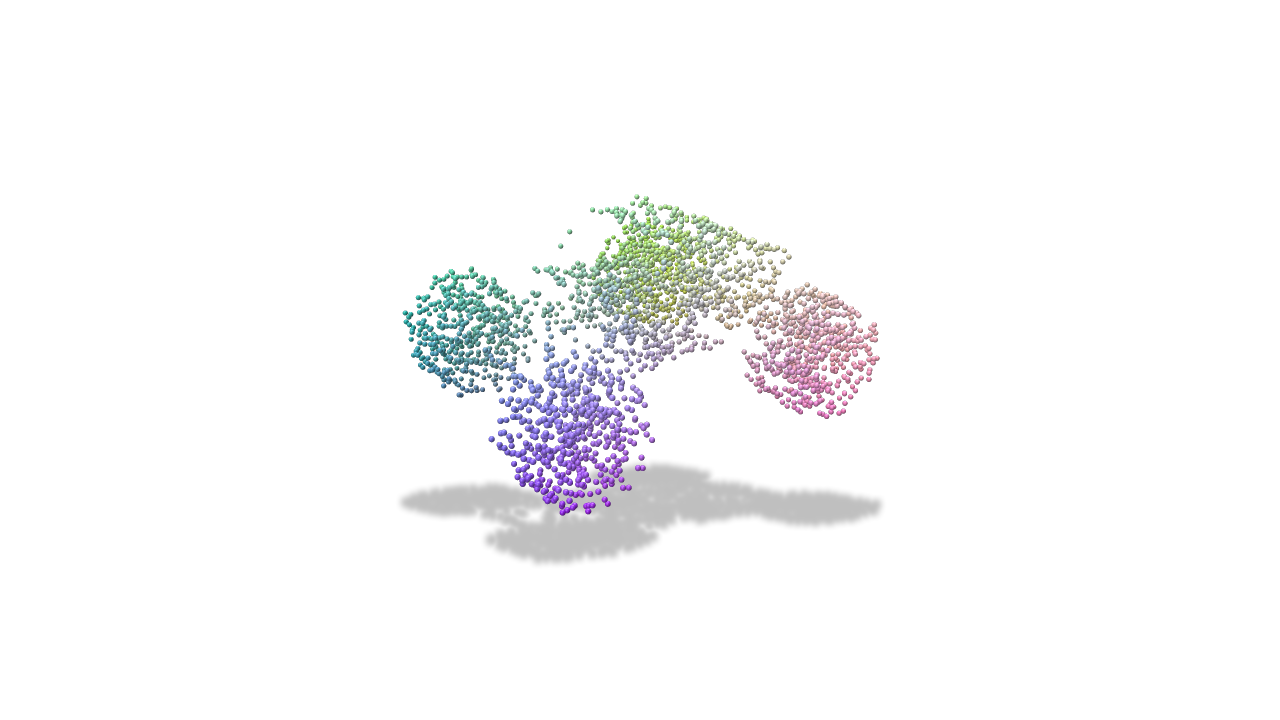}\\
    \caption{
    Single-image 3D reconstruction visualizations on held-out data. 
    Per inset, columns represent 
    (i) the input RGB image, 
    (ii) the visibility $\widehat{\xi}$, 
    (iii) the depth $\widehat{d}$, 
    (iv) the normals $\widehat{n}$, 
    (v) the sampled point cloud (PC) from the DDF, and 
    (vi) a sample from the ground-truth PC.
    Quantities (ii-v) are all differentiably computed directly from the CPDDF and $\widehat{\Pi}$, per point or pixel (i.e., no post-processing needed).
    PC colours denote 3D coordinates.
    A high-error example is in the lower-right of each category.
    See also Fig.\ \ref{fig:si3drvis} and \S\ref{app:sec:si3draddvis}.
    }
    \label{app:fig:si3drvis}
\end{figure*}

\section{Single-Image 3D Reconstruction}
\label{appendix:si3dr}

\textbf{Architecture and Optimization.}
Our CPDDF is a modulated SIREN \cite{mehta2021modulated} with layer sizes \verb|(512, 512, 512,| \verb|256, 256,| \verb|256, 256)|. 
Note that we use a softplus activation instead of ReLU, when multiplying the modulator to the intermediate features.
The encoders (one ResNet-18 for the camera, and another for inferring the latent conditioning shape vector) take 128 $\times$ 128 RGBA images as input. 
However, when rendering the visibility mask for $\mathcal{L}_M$, we output 64 $\times$ 64 images.
We set $\gamma_{R,S} = 1$, $\gamma_{R,\Pi} = 5$, and $\gamma_{R,M} = 10$.
We also used small weight decays on the camera and shape predictors ($10^{-6}$ and $10^{-3}$, respectively). %
Training ran for 100K iterations with AdamW \cite{loshchilov2018decoupled} and a batch size of 32. 
We used $\dim(z_s) = 512$ and a SIREN initializer of $\omega_0 = 1$. 

\textbf{Shape-fitting Loss.}
We use the same settings as the single-shape fitting experiments, with slight modifications:
$\gamma_d = 5$,
$\gamma_\xi = 10$,
$\gamma_n = 10$,
$\gamma_V = 1$,
$\gamma_\mathrm{E,d} = 0.1$, 
$\gamma_\mathrm{E,\xi} = 0.25$, 
and
$\gamma_T = 0.1$.
An additional loss on the variance of $\xi$ is also applied 
(to reduce the visibility entropy at each point, which leads to fuzzy renders):
\begin{equation}
    \mathcal{L}_{V,\xi} = \gamma_{V,\xi} \xi(p,v)[ 1 - \xi(p,v) ],
\end{equation}
where we set
$\gamma_{V,\xi} = 0.25$.
This encourages less blurriness in the shape output due to uncertainty in the visibility field.
For minibatches, we sample 1.2K (A and U) and 300 (S, B, T, and O) oriented points per shape, with each minibatch containing 32 shapes.

\textbf{Camera Loss.}
The camera fitting loss $\mathcal{L}_\Pi$ utilizes a camera representation only involving extrinsic position (the camera is assumed to be looking at the origin). 
In particular, we use the azimuth, elevation, and radius representation of the camera position.
Before computing the $L_2$ loss, we z-score normalize each element, based on the training set statistics.
We also restrict the predicted camera $\widehat{\Pi}$ to be within the range of the parameters observed in the dataset.

\textbf{Rendering Scale Factor.}
We note that an additional scale factor is needed for rendering DDFs for ShapeNet. 
Since ShapeNet shapes are normalized with an instance-dependent measure (the bounding box diagonal), one needs to know the scale to reproduce the output image.
This is an issue as our CPDDF always outputs a shape in $[-1,1]^3$, with training data normalized to have longest axis-aligned bounding box length equal to two.
Further, it creates an ambiguity (with respect to the output image) with the camera position (radius from the shape).
At train time, before rendering to compute $\mathcal{L}_M$, 
we use the ground-truth scale factor. 
At test time, we estimate it by sampling a point cloud and measuring the diagonal.

\textbf{Data Extraction.}
We use the ShapeNet-v1 \cite{shapenet2015} data and splits from Pixel2Mesh \cite{wang2018pixel2mesh,wang2020pixel2mesh}, with the renders by Choy et al.\  \cite{choy20163d}.
For DDF training, 
per data type,
we sample 20K (A and U) and 10K (S, B, T, and O) training samples per shape,
using the watertight form of the meshes (via \cite{Kaolin}), 
decimated to 10K triangles.
These samples are only used for training, not evaluation, and are in a canonical aligned pose.
We set the maximum offset size for O-type data as $\epsilon_O = 0.02$ (see \S\ref{sec:datatypes}); remaining parameters are the same as those used for single shapes (see \S\ref{appendix:singlefits}).

\textbf{Explicit Sampling Details.}
We also remark that our explicit sampling algorithm slightly oversamples points initially (when requesting a point set of size $n_p$, we actually sample $(1 + \varepsilon_p) n_p$ via $p \sim \mathcal{U}[\mathcal{B}]$).
The final point cloud, however, is sorted by visibility (i.e., by $\xi(p,\widehat{v}^{\,*}(p))$) and only the top $n_p$ points are returned.
We used $\varepsilon_p = 0.1$ in all experiments.
This is to prevent outputting non-visible points.

\textbf{PC-SIREN Baseline Details.}
Recall that the PC-SIREN is our architecture-matched baseline, with identical encoders and a nearly identical decoder architecture to the DDF one. 
Here, the decoder is a mapping $f_b : \mathbb{R}^3 \rightarrow \mathbb{R}^3$, which is trained to compute $f_b(p)\in S$ from $p\sim[-1,1]^3$, but uses an identical set of SIREN hidden layers as the DDF. 
A set of sampled random points can thus be mapped into a point cloud of arbitrary size.
The baseline has 25,645,574 parameters, while the DDF-based model has 25,647,367 parameters.
The camera loss $\mathcal{L}_\Pi$ is unchanged and the mask-matching loss $\mathcal{L}_M$ is not used.
The shape-fitting loss $\mathfrak{L}_S$ is replaced with a standard Chamfer loss $D_C$ \cite{chamfer}, computed with 1024 points per shape with a batch-size of 32. The remaining aspects of training remain the same.

\subsection{Ablation with $N_H=1$}

\label{appendix:si3dr:nh1}

Recall that $N_H$ is the number of times to ``cycle'' the points (projecting them towards the surface via the DDF) when sampling an explicit point cloud shape from a DDF (see \S\ref{sec:app:si3dr}).
We show results with $N_H=1$ in Table \ref{tab:si3dr1h}.
In most cases, it is slightly worse than using $N_H = 3$, by 1-3 $F$-score units;
occasionally, however, it is  marginally better: on Planes-$\widehat{\Pi}$, it has slightly lower $D_C$, though this does not translate to better $F$-score.

\begin{table} %
    \centering
    \begin{tabular}{cr|cccc}
         &  & \multicolumn{4}{c}{DDF}  \\
         &  & $\Pi_g$-L & $\Pi_g$-S  & $\widehat{\Pi}$-L  & $\widehat{\Pi}$-S  \\\hline 
           & $D_C$  $ \downarrow$ & 0.477 & 0.532 & 0.861 & 0.928  \\ 
    Chairs & $F_{\tau}$ $ \uparrow$ & 54.37 & 47.26 & 46.81 & 40.35  \\ 
          & $F_{2\tau}$ $ \uparrow$ & 71.62 & 66.33 & 63.09 & 58.09  \\\hline 
              & $D_C$  $ \downarrow$ & 0.201 & 0.231 & 0.748 & 0.799  \\ 
    Planes & $F_{\tau}$  $ \uparrow$ & 80.69 & 76.71 & 63.86 & 60.48   \\ 
     &      $F_{2\tau}$  $ \uparrow$ & 90.23 & 88.49 & 75.30 & 73.55  \\ \hline
    & $D_C$  $ \downarrow$ & 0.235 & 0.309 & 0.545 & 0.628  \\ 
    Cars & $F_{\tau}$  $ \uparrow$ & 68.21 & 57.36 & 59.84 & 49.98  \\ 
     & $F_{2\tau}$  $ \uparrow$ & 83.99 & 76.41 & 76.87 & 69.32  \\ 
    \end{tabular}
    \caption{
        Single-image 3D reconstruction results with $N_H = 1$.
    }
    \label{tab:si3dr1h}
\end{table}

\subsection{Additional Visualizations}
\label{app:sec:si3draddvis}

Some additional visualizations are shown in Fig.\ \ref{app:fig:si3drvis}. 
For highly novel inputs, we also observe that, sometimes, the network does not adapt well to the shape (e.g., see the chairs example in the second row and second column).
While much error is due to the incorrectly predicted camera, the DDF outputs can also be a bit blurrier, especially when it is uncertain about the shape.
This can occur on thin structures, which are hard to localize (e.g., the chair legs in either row one and column one, or row three column two), or atypical inputs (e.g., row three, column two of the cars examples), where the network does not have enough examples to obtain high quality geometry.
One can also see some non-uniform densities from our point cloud sampling algorithm (e.g., concentrations of points on the chair legs in column two, or on the wheels of several examples of cars).

\begin{figure}
    \centering
    \begin{tikzpicture}
    \node[xshift=0cm,yshift=0cm,draw,rounded corners] (PN) at (-0.0,0.0) {$(P,N)$};
    \node[xshift=0cm,yshift=0cm,draw,circle,fill={rgb,255:red,255; green,128; blue,128},inner sep=1.3pt] (zs1) [right=0.4in of PN] {$z_s$};
    \node[xshift=0cm,yshift=0cm,draw,rounded corners] (d1) [right=0.4in of zs1] {$d(p,v|z_s)$};
    \draw [-latex,thick] (PN.east) -- node[right,above] {$E$} (zs1.west);
    \draw [-latex,thick,densely dashed] (zs1.east) -- node[right,above] { } (d1.west);
    \end{tikzpicture}\\[3mm]
    \begin{tikzpicture}
    \node[xshift=0cm,yshift=0cm,draw,circle,fill={rgb,255:red,160; green,203; blue,250},inner sep=0.9pt] (pi) at (0,0) {$\Pi$};
    \node[xshift=0cm,yshift=0cm,draw,circle,fill={rgb,255:red,255; green,128; blue,128},inner sep=1.3pt] (zs) [below left=0.25in and 0.25in of pi] {$z_s$};
    \path let \p1=(pi), \p2=(zs) in node[] (h1) at (\x1,\y2) { };
    \node[xshift=0cm,yshift=0cm,draw,rounded corners] (In) [right=0.4in of h1] {$I_n$};
    \node[xshift=0cm,yshift=0cm,draw,circle,fill={rgb,255:red,85; green,181; blue,98},inner sep=1.1pt] (zt) [above right=0.2in and 0.25in of In] {$z_T$};
    \path let \p1=(zt), \p2=(zs) in node[draw,circle,inner sep=0pt] (h2) at (\x1,\y2) { $+$ };
    \node[xshift=0cm,yshift=0cm,draw,rounded corners] (It) [right=0.4in of h2] {$I_T$};
    \draw [thick,densely dashed] (zs.east) --  node[right,above] {} (h1.center) ;
    \draw [-latex,thick,densely dashed] (h1.center) --  node[right,above] { $\nabla_p d$ } (In.west) ;
    \draw [-latex,thick] (h2.east) --  node[right,above] { $f_T$ } (It.west) ;
    \draw [thick,densely dashed] (pi.south) --  node[right,above] {} (h1.center) ;
    \draw [-latex,thick] (zt.south) -- node[] {} (h2.north) ;
    \draw [-latex,thick] (In.east) -- node[] {} (h2.west) ;
    \end{tikzpicture}
    \caption{
    Two-stage unpaired generative modelling architecture. 
    \textit{Upper inset}: 
    VAE formulation mapping a point cloud $P$ and associated normals $N$ to latent shape vector $z_s$ via PointNet encoder $E$, and decoding into depth values via conditioning a PDDF. 
    \textit{Lower inset}: 
    latent shape $z_s$ and camera $\Pi$ are randomly sampled and used to render a surface normals image $I_n$, via the derivatives of the learned conditional PDDF 
    (see Property \hyperref[property2]{II}).
    The normals map $I_n$ is then concatenated ($\oplus$) with a sampled latent texture $z_T$, and used to compute the final RGB image $I_T = f_T(I_n,z_T)$.
    Coloured circles indicate random variables sampled from a particular distribution; dashed lines indicate computation with multiple forward passes (per point or pixel). 
    }
    \label{fig:ugenarch}
\end{figure}
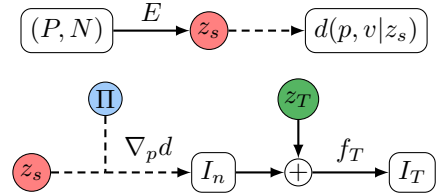

\begin{figure*}%
    \centering
    \adjincludegraphics[width=0.99\textwidth,trim={{.0\width} 0 {.0\width}  0},clip]{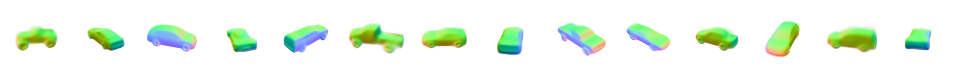}    
    \adjincludegraphics[width=0.99\textwidth,trim={{.0\width} 0 {.0\width}  0},clip]{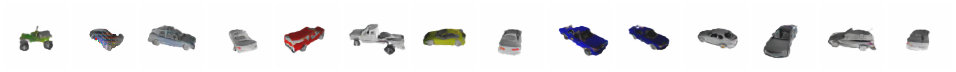}
    \adjincludegraphics[width=0.99\textwidth,trim={{.0\width} 0 {.0\width} 0},clip]{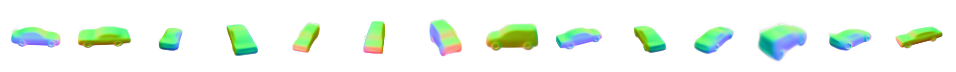}
    \adjincludegraphics[width=0.99\textwidth,trim={{.0\width} 0 {.0\width}  0},clip]{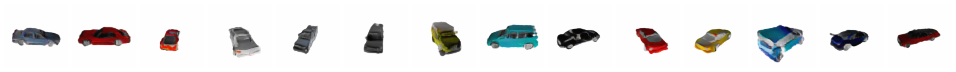}
\caption{
Additional example samples from the ShapeVAE and translational image GAN.
}
\label{fig:gangenssuppa}
\end{figure*}

\begin{figure}%
    \centering
    \adjincludegraphics[width=0.235\textwidth,trim={{.0\width} 0 {.0\width}  0},clip]{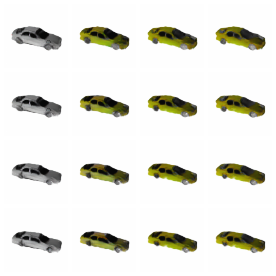}
    \hfill
    \adjincludegraphics[width=0.235\textwidth,trim={ {.0\width} 0 {.0\width}  0},clip]{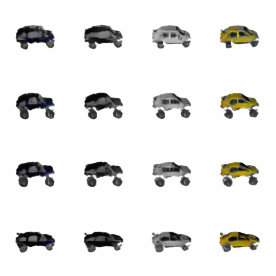}\\
\caption{
Additional example interpolations from the ShapeVAE and translational image GAN.
}
\label{fig:gangenssuppb}
\end{figure}

\section{Generative Modelling}
\label{appendix:genmodel}

See Fig.\ \ref{fig:ugenarch} for a diagram of our two-stage 3D-to-2D modality translation architecture.
See also Fig.\ \ref{fig:gangenssuppa} and \ref{fig:gangenssuppb} for additional sample visualizations (as in Fig.\ \ref{fig:gangens}).

\subsection{Shape VAE}
The first stage learns a VAE \cite{kingma2013auto,rezende2014stochastic}
on 3D shape, in order to 
(i) obtain a latent shape variable $z_s$ that is approximately Gaussian distributed, and 
(ii) learn a CPDDF using $z_s$ that can encode a shape in a form that is easily and efficiently rendered, yet still encodes higher order shape information.
We learn a PointNet encoder \cite{qi2017pointnet} $E$ 
    that maps a point cloud with normals $(P,N)$ 
    to a latent shape vector $z_s = E(P,N)$.
The conditional PDDF (CPDDF) then acts as the decoder, 
    computing depth values as $d(p,v|z_s)$ for a given input.
To implement conditional field computations, 
    we use the modulated SIREN approach \cite{mehta2021modulated}.
This can be trained by the following $\beta$-VAE loss
\cite{higgins2016beta}:
\begin{equation}
    \mathfrak{L}_{\mathrm{VAE}} = 
        \mathfrak{L}_S + \beta \mathcal{L}_\mathrm{KL},
\end{equation}
where $\mathcal{L}_\mathrm{KL}$ is the standard KL divergence loss between a Gaussian prior and the approximate VAE posterior, 
and $\mathfrak{L}_S$ acts analogously to the reconstruction likelihood.
We use single categories of ShapeNet \cite{shapenet2015} to fit the Shape VAE.

\textbf{Data.}
Data is extracted from 1200 randomly chosen shapes from ShapeNet-v1 cars \cite{shapenet2015}, 
    sampling 60k (A and U) and 30K (S, B, T, and O) oriented points.
We downsample shapes to 10K triangles before extraction.
For minibatches, we sample 1.2K (A and U) and 300 (S, B, T, and O) oriented points per shape, with each minibatch containing 32 shapes.
We also sample point clouds $P$ (with normals $N$) of size 1024 to send to the PointNet at each minibatch.

\textbf{Architectures.}
The PointNet encoder follows the standard classification architecture 
    \cite{qi2017pointnet},
    with four \verb|Conv1D-BatchNorm-ReLU| blocks 
    (sizes: 64, 64, 128, 1024), 
    followed 
    by max pooling and a multilayer perceptron with \verb|Linear-BatchNorm-ReLU| blocks (two hidden layers of size 512; dropout probability 0.1).
No point transformers are used.
The final output is of size $2\mathrm{dim}(z_s)$.
We then compute the approximate variational posterior $q(z_s|P,N) = \mathcal{N}(z_s|\mu(P,N),\Sigma(P,N))$ 
with two networks for mean $\mu$ and diagonal log-variance matrix $\Sigma$, each structured as 
\verb|Linear-ReLU-Linear| 
(in which all layers are of dimensionality $\mathrm{dim}(z_s)$; note that each posterior parameter network takes half of the vector output from the PointNet encoder as input).

For the decoder,
    we use eight layers \verb|(512,| \verb|512,| \verb|512, 512, 256, 256, 256, 256)| (with the modulated SIREN \cite{mehta2021modulated}).
We set $\mathrm{dim}(z_s) = 400$ and $\mathrm{dim}(z_T) = 64$.

\textbf{Training.}
We run for 100K iterations, using 
$\beta = 0.05$,
$\gamma_d = 5$,
$\gamma_\xi = 10$,
$\gamma_n = 10$,
$\gamma_V = 1$,
$\gamma_\mathrm{E,d} = 0.1$, 
$\gamma_\mathrm{E,\xi} = 0.1$, 
$\gamma_{V,\xi} = 0.1$,
and
$\gamma_T = 0.1$.
Adam is used for optimization 
(learning rate $10^{-4}$; $\beta_1 = 0.9$, $\beta_2 = 0.999$).

\subsection{Image GAN}

After training a Shape VAE, 
    the CPDDF decoder can be used to render a surface normals image $I_n$
    (see \S\ref{sec:app:rendering} and Fig.\ \ref{fig:singobjfits}).
To perform generation,
    we first sample latent shape $z_s\sim\mathcal{N}(0,I)$
    and
    camera $\Pi$, 
        which includes extrinsics (position and orientation) and focal length,
    following with normals map rendering to get $I_n$.
We then use a convolutional network $f_T$ to obtain the RGB image $I_T$, 
    based on the residual image-to-image translation architecture 
    from CycleGAN \cite{CycleGAN2017}.
This is done by sampling $z_T\sim\mathcal{N}(0,I)$
    and
    computing $I_T = f_T(I_n,z_T)$,
    where
    one concatenates $z_T$ to each pixel of $I_n$ before processing.
Notice that $z_s$, $z_T$, and $\Pi$ are independent, while the final texture (appearance) depends directly on $z_T$, and indirectly on $z_s$ and $\Pi$, through $I_n$.

For training,
    a non-saturating GAN loss \cite{goodfellow2014generative} 
    with a zero-centered gradient penalty for the discriminator \cite{roth2017stabilizing,mescheder2018training} 
    is used 
    (as in \cite{niemeyer2021giraffe,chan2021pi}).
To ensure that $f_T$ preserves information regarding 3D shape and latent texture, we use two consistency losses:
\begin{equation}
    \mathcal{L}_\mathrm{C,S} = \mathrm{MSE}(I_n, \widehat{I}_n)
    \;\;\;\&\;\;\;
    \mathcal{L}_\mathrm{C,T} = \mathrm{MSE}(z_T, \widehat{z}_T),
\end{equation}
where $\mathrm{MSE}$ is the mean squared error, 
$\widehat{I}_n = g_T(I_T)$ (with $g_T$ having identical architecture to $f_T$),
and 
$\widehat{z}_T = h_T(I_T)$ 
(with $h_T$ implemented as a ResNet-20 \cite{he2016deep,Idelbayev18a}).
The first loss, $\mathcal{L}_\mathrm{C,S}$,
encourages the fake RGB image $I_T$ to retain the 3D shape information from $I_n$ through the translation process
(i.e., implicitly forcing the image resemble the input normals)
while the second loss, $\mathcal{L}_\mathrm{C,T}$,
does the same for the latent texture 
(implicitly strengthening textural consistency across viewpoints).
The final loss for the GAN image generator is
\begin{equation}
    \mathfrak{L}_{\mathrm{GAN}} = 
    \mathcal{L}_\mathrm{A} + 
    \gamma_{C,S} \mathcal{L}_\mathrm{C,S} + 
    \gamma_{C,T} \mathcal{L}_\mathrm{C,T}, 
\end{equation}
where $\mathcal{L}_\mathrm{A}$ is the adversarial loss for the generator and
the last two terms enforce consistency
(see also \cite{miyauchi2018shape,kaya2020self,aumentado2020cycle}).
We fit to the dataset of ShapeNet renders from Choy et al.\ \cite{choy20163d}. 
Note that information on the correspondence with the 3D models is not used (i.e., the images and shapes are treated independently).

\textbf{Generation process.}
Recall that our generation process can be written
$I_T = G(z_s,z_T,\Pi) = f_T(I_n(z_s,\Pi)\oplus z_T)$, 
where $I_n(z_s,\Pi)$ denotes the CPDDF normals render 
(see \S\ref{sec:app:rendering}) and $\oplus$ refers to concatenating $z_T$ to every pixel of $I_n$.
For latent sampling, 
$z_s,z_T\sim\mathcal{N}(0,I)$, while the camera $\Pi$ is sampled from the upper hemisphere above the object, oriented toward the origin at a fixed distance and with a fixed focal length.
The image size was set to $64\times 64$.

\textbf{Networks.}
The translation network $f_T$ exactly follows the architecture from CycleGAN \cite{CycleGAN2017} consisting of residual blocks 
(two downsampling layers, then six resolution-retaining layers, followed by two upsampling layers), using the code from \cite{cyclegancode}.
The normals consistency network, $g_T$, has identical architecture to $f_T$, 
    while the texture consistency network, $h_T$, is a ResNet-20 \cite{he2016deep,Idelbayev18a}.
We utilize the convolutional discriminator implementation from Mimicry \cite{lee2020mimicry}, based on DCGAN \cite{radford2015unsupervised}.

\textbf{Training.}
Our image GAN is trained in the standard alternating manner, 
    using two critic training steps for every generator step.
The non-saturating loss \cite{goodfellow2014generative} was used, along with a zero-centered gradient penalty \cite{roth2017stabilizing,mescheder2018training} (with a weight of 10 during critic training).
We used the following loss weights:
$\gamma_{C,S} = 1$ %
and
$\gamma_{C,T} = 1$ %
For optimization, we use Adam 
(learning rate $10^{-4}$; $\beta_1 = 0.0$, $\beta_2 = 0.9$)
for 100K iterations 
(with the same reduce-on-plateau scheduler as in Appendix \ref{appendix:singlefits}).

\textbf{2D GAN Comparison.}
As mentioned in the paper, we trained a convolutional GAN with the same loss and critic architecture using the Mimicry library \cite{lee2020mimicry}. 
We evaluated both our model and the 2D GAN with Frechet Inception Distance (FID), using torch-fidelity \cite{obukhov2020torchfidelity} with 50K samples.

%% file: PaperForReview.bbl
\begin{thebibliography}{10}\itemsep=-1pt

\bibitem{aumentado2020cycle}
Tristan Aumentado-Armstrong, Alex Levinshtein, Stavros Tsogkas, Konstantinos~G
  Derpanis, and Allan~D Jepson.
\newblock Cycle-consistent generative rendering for {2D-3D} modality
  translation.
\newblock In {\em 2020 International Conference on 3D Vision (3DV)}, pages
  230--240. IEEE, 2020.

\bibitem{barron2021mip}
Jonathan~T Barron, Ben Mildenhall, Matthew Tancik, Peter Hedman, Ricardo
  Martin-Brualla, and Pratul~P Srinivasan.
\newblock Mip-{NeRF}: A multiscale representation for anti-aliasing neural
  radiance fields.
\newblock {\em arXiv preprint arXiv:2103.13415}, 2021.

\bibitem{chan2021pi}
Eric~R Chan, Marco Monteiro, Petr Kellnhofer, Jiajun Wu, and Gordon Wetzstein.
\newblock pi-{GAN}: Periodic implicit generative adversarial networks for
  {3D}-aware image synthesis.
\newblock In {\em Proceedings of the IEEE/CVF Conference on Computer Vision and
  Pattern Recognition}, pages 5799--5809, 2021.

\bibitem{shapenet2015}
Angel~X. Chang, Thomas Funkhouser, Leonidas Guibas, Pat Hanrahan, Qixing Huang,
  Zimo Li, Silvio Savarese, Manolis Savva, Shuran Song, Hao Su, Jianxiong Xiao,
  Li Yi, and Fisher Yu.
\newblock {ShapeNet: An Information-Rich 3D Model Repository}.
\newblock Technical Report arXiv:1512.03012 [cs.GR], Stanford University ---
  Princeton University --- Toyota Technological Institute at Chicago, 2015.
\newblock \url{https://shapenet.org/}. Terms of use:
  {\url{https://shapenet.org/terms}}.

\bibitem{chibane2020neural}
Julian Chibane, Aymen Mir, and Gerard Pons-Moll.
\newblock Neural unsigned distance fields for implicit function learning.
\newblock {\em arXiv preprint arXiv:2010.13938}, 2020.

\bibitem{choy20163d}
Christopher~B Choy, Danfei Xu, JunYoung Gwak, Kevin Chen, and Silvio Savarese.
\newblock {3D-R2N2}: A unified approach for single and multi-view {3D} object
  reconstruction.
\newblock In {\em Proceedings of the European Conference on Computer Vision
  ({ECCV})}, 2016.

\bibitem{curless1996volumetric}
Brian Curless and Marc Levoy.
\newblock A volumetric method for building complex models from range images.
\newblock In {\em Proceedings of the 23rd annual conference on Computer
  graphics and interactive techniques}, pages 303--312, 1996.

\bibitem{aman_dalmia_2020_3902941}
Aman Dalmia.
\newblock dalmia/siren.
\newblock {\em \url{https://doi.org/10.5281/zenodo.3902941}}, June 2020.
\newblock Zenodo: 10.5281/zenodo.3902941. \url{https://github.com/dalmia/siren}
  (MIT License).

\bibitem{gao2020learning}
Jun Gao, Wenzheng Chen, Tommy Xiang, Clement~Fuji Tsang, Alec Jacobson, Morgan
  McGuire, and Sanja Fidler.
\newblock Learning deformable tetrahedral meshes for {3D} reconstruction.
\newblock {\em arXiv preprint arXiv:2011.01437}, 2020.

\bibitem{garbin2021fastnerf}
Stephan~J Garbin, Marek Kowalski, Matthew Johnson, Jamie Shotton, and Julien
  Valentin.
\newblock Fast{NeRF}: High-fidelity neural rendering at 200fps.
\newblock {\em arXiv preprint arXiv:2103.10380}, 2021.

\bibitem{gkioxari2019mesh}
Georgia Gkioxari, Jitendra Malik, and Justin Johnson.
\newblock Mesh {R-CNN}.
\newblock In {\em Proceedings of the IEEE/CVF International Conference on
  Computer Vision}, pages 9785--9795, 2019.

\bibitem{goel2020shape}
Shubham Goel, Angjoo Kanazawa, and Jitendra Malik.
\newblock Shape and viewpoint without keypoints.
\newblock In {\em European Conference on Computer Vision}, pages 88--104.
  Springer, 2020.

\bibitem{goodfellow2014generative}
Ian Goodfellow, Jean Pouget-Abadie, Mehdi Mirza, Bing Xu, David Warde-Farley,
  Sherjil Ozair, Aaron Courville, and Yoshua Bengio.
\newblock Generative adversarial nets.
\newblock {\em Advances in neural information processing systems}, 27, 2014.

\bibitem{gropp2020implicit}
Amos Gropp, Lior Yariv, Niv Haim, Matan Atzmon, and Yaron Lipman.
\newblock Implicit geometric regularization for learning shapes.
\newblock {\em arXiv preprint arXiv:2002.10099}, 2020.

\bibitem{chamfer}
Thibault Groueix.
\newblock Pytorch chamfer distance.
\newblock {\em ThibaultGROUEIX/ChamferDistancePytorch}, Nov. 2021.
\newblock \url{https://github.com/ThibaultGROUEIX/ChamferDistancePytorch} (MIT
  License).

\bibitem{he2016deep}
Kaiming He, Xiangyu Zhang, Shaoqing Ren, and Jian Sun.
\newblock Deep residual learning for image recognition.
\newblock In {\em Proceedings of the IEEE conference on computer vision and
  pattern recognition}, pages 770--778, 2016.

\bibitem{hedman2021baking}
Peter Hedman, Pratul~P Srinivasan, Ben Mildenhall, Jonathan~T Barron, and Paul
  Debevec.
\newblock Baking neural radiance fields for real-time view synthesis.
\newblock {\em arXiv preprint arXiv:2103.14645}, 2021.

\bibitem{henderson2020leveraging}
Paul Henderson, Vagia Tsiminaki, and Christoph~H Lampert.
\newblock Leveraging {2D} data to learn textured {3D} mesh generation.
\newblock In {\em Proceedings of the IEEE/CVF Conference on Computer Vision and
  Pattern Recognition}, pages 7498--7507, 2020.

\bibitem{heusel2017gans}
Martin Heusel, Hubert Ramsauer, Thomas Unterthiner, Bernhard Nessler, and Sepp
  Hochreiter.
\newblock {GANs} trained by a two time-scale update rule converge to a local
  nash equilibrium.
\newblock {\em Advances in neural information processing systems}, 30, 2017.

\bibitem{higgins2016beta}
Irina Higgins, Loic Matthey, Arka Pal, Christopher Burgess, Xavier Glorot,
  Matthew Botvinick, Shakir Mohamed, and Alexander Lerchner.
\newblock Beta-{VAE}: Learning basic visual concepts with a constrained
  variational framework.
\newblock {\em {International Conference on Learning Representations
  ({ICLR})}}, 2017.

\bibitem{Idelbayev18a}
Yerlan Idelbayev.
\newblock Proper {ResNet} implementation for {CIFAR10/CIFAR100} in {PyTorch}.
\newblock \url{https://github.com/akamaster/pytorch_resnet_cifar10}.
\newblock Accessed: Nov 2021. (BSD-2-clause License).

\bibitem{eldar}
Eldar Insafutdinov and Alexey Dosovitskiy.
\newblock Unsupervised learning of shape and pose with differentiable point
  clouds.
\newblock {\em Advances in Neural Information Processing Systems}, pages
  2807--2817, 2018.

\bibitem{Kaolin}
Krishna~Murthy Jatavallabhula, Edward Smith, Jean-Francois Lafleche,
  Clement~Fuji Tsang, Artem Rozantsev, Wenzheng Chen, Tommy Xiang, Rev
  Lebaredian, and Sanja Fidler.
\newblock Kaolin: A {PyTorch} library for accelerating {3D} deep learning
  research.
\newblock {\em arXiv:1911.05063}, 2019.
\newblock \url{https://github.com/NVIDIAGameWorks/kaolin} (Apache License).

\bibitem{jiang2020sdfdiff}
Yue Jiang, Dantong Ji, Zhizhong Han, and Matthias Zwicker.
\newblock Sdfdiff: Differentiable rendering of signed distance fields for {3D}
  shape optimization.
\newblock In {\em Proceedings of the IEEE/CVF Conference on Computer Vision and
  Pattern Recognition}, pages 1251--1261, 2020.

\bibitem{kato2018neural}
Hiroharu Kato, Yoshitaka Ushiku, and Tatsuya Harada.
\newblock Neural {3D} mesh renderer.
\newblock In {\em Proceedings of the IEEE conference on computer vision and
  pattern recognition}, pages 3907--3916, 2018.

\bibitem{kaya2020self}
Berk Kaya and Radu Timofte.
\newblock Self-supervised 2d image to {3D} shape translation with disentangled
  representations.
\newblock In {\em 2020 International Conference on 3D Vision (3DV)}, pages
  1039--1048. IEEE, 2020.

\bibitem{Kellnhofer:2021:nlr}
Petr Kellnhofer, Lars Jebe, Andrew Jones, Ryan Spicer, Kari Pulli, and Gordon
  Wetzstein.
\newblock Neural lumigraph rendering.
\newblock In {\em CVPR}, 2021.

\bibitem{kingma2014adam}
Diederik~P Kingma and Jimmy Ba.
\newblock Adam: A method for stochastic optimization.
\newblock {\em arXiv preprint arXiv:1412.6980}, 2014.

\bibitem{kingma2013auto}
Diederik~P Kingma and Max Welling.
\newblock Auto-encoding variational {Bayes}.
\newblock {\em arXiv preprint arXiv:1312.6114}, 2013.

\bibitem{kreyszigdg}
Erwin Kreyszig.
\newblock {\em Differential Geometry}.
\newblock University of Toronto Press, 1959.
\newblock Mathematical Expositions, No. 11. Dover edition.

\bibitem{krishnamurthy1996fitting}
Venkat Krishnamurthy and Marc Levoy.
\newblock Fitting smooth surfaces to dense polygon meshes.
\newblock In {\em Proceedings of the 23rd annual conference on Computer
  graphics and interactive techniques}, pages 313--324, 1996.

\bibitem{kulkarni2015deep}
Tejas~D Kulkarni, Will Whitney, Pushmeet Kohli, and Joshua~B Tenenbaum.
\newblock Deep convolutional inverse graphics network.
\newblock {\em arXiv preprint arXiv:1503.03167}, 2015.

\bibitem{lee2020mimicry}
Kwot~Sin Lee and Christopher Town.
\newblock Mimicry: Towards the reproducibility of {GAN} research.
\newblock {\em CVPR Workshop on AI for Content Creation}, 2020.
\newblock \url{https://github.com/kwotsin/mimicry} (MIT License).

\bibitem{lin2020sdf}
Chen-Hsuan Lin, Chaoyang Wang, and Simon Lucey.
\newblock {SDF-SRN}: Learning signed distance {3D} object reconstruction from
  static images.
\newblock {\em arXiv preprint arXiv:2010.10505}, 2020.

\bibitem{lindell2021autoint}
David~B Lindell, Julien~NP Martel, and Gordon Wetzstein.
\newblock Autoint: Automatic integration for fast neural volume rendering.
\newblock In {\em Proceedings of the IEEE/CVF Conference on Computer Vision and
  Pattern Recognition}, pages 14556--14565, 2021.

\bibitem{liu2019soft}
Shichen Liu, Tianye Li, Weikai Chen, and Hao Li.
\newblock Soft rasterizer: A differentiable renderer for image-based {3D}
  reasoning.
\newblock In {\em Proceedings of the IEEE/CVF International Conference on
  Computer Vision}, pages 7708--7717, 2019.

\bibitem{liu2019learning}
Shichen Liu, Shunsuke Saito, Weikai Chen, and Hao Li.
\newblock Learning to infer implicit surfaces without {3D} supervision.
\newblock {\em arXiv preprint arXiv:1911.00767}, 2019.

\bibitem{liu2020dist}
Shaohui Liu, Yinda Zhang, Songyou Peng, Boxin Shi, Marc Pollefeys, and Zhaopeng
  Cui.
\newblock Dist: Rendering deep implicit signed distance function with
  differentiable sphere tracing.
\newblock In {\em Proceedings of the IEEE/CVF Conference on Computer Vision and
  Pattern Recognition}, pages 2019--2028, 2020.

\bibitem{loshchilov2018decoupled}
Ilya Loshchilov and Frank Hutter.
\newblock Decoupled weight decay regularization.
\newblock In {\em International Conference on Learning Representations}, 2019.

\bibitem{martel2021acorn}
Julien~NP Martel, David~B Lindell, Connor~Z Lin, Eric~R Chan, Marco Monteiro,
  and Gordon Wetzstein.
\newblock Acorn: Adaptive coordinate networks for neural scene representation.
\newblock {\em arXiv preprint arXiv:2105.02788}, 2021.

\bibitem{mehta2021modulated}
Ishit Mehta, Micha{\"e}l Gharbi, Connelly Barnes, Eli Shechtman, Ravi
  Ramamoorthi, and Manmohan Chandraker.
\newblock Modulated periodic activations for generalizable local functional
  representations.
\newblock {\em arXiv preprint arXiv:2104.03960}, 2021.

\bibitem{mescheder2018training}
Lars Mescheder, Andreas Geiger, and Sebastian Nowozin.
\newblock Which training methods for {GANs} do actually converge?
\newblock In {\em International conference on machine learning}, pages
  3481--3490. PMLR, 2018.

\bibitem{mescheder2019occupancy}
Lars Mescheder, Michael Oechsle, Michael Niemeyer, Sebastian Nowozin, and
  Andreas Geiger.
\newblock Occupancy networks: Learning {3D} reconstruction in function space.
\newblock In {\em Proceedings of the IEEE/CVF Conference on Computer Vision and
  Pattern Recognition}, pages 4460--4470, 2019.

\bibitem{mildenhall2020nerf}
Ben Mildenhall, Pratul~P Srinivasan, Matthew Tancik, Jonathan~T Barron, Ravi
  Ramamoorthi, and Ren Ng.
\newblock {NeRF}: Representing scenes as neural radiance fields for view
  synthesis.
\newblock In {\em European conference on computer vision}, pages 405--421.
  Springer, 2020.

\bibitem{miyauchi2018shape}
Yutaro Miyauchi, Yusuke Sugano, and Yasuyuki Matsushita.
\newblock Shape-conditioned image generation by learning latent appearance
  representation from unpaired data.
\newblock In {\em Asian Conference on Computer Vision}, pages 438--453.
  Springer, 2018.

\bibitem{nguyen2018rendernet}
Thu Nguyen-Phuoc, Chuan Li, Stephen Balaban, and Yong-Liang Yang.
\newblock {RenderNet}: a deep convolutional network for differentiable
  rendering from {3D} shapes.
\newblock In {\em Proceedings of the 32nd International Conference on Neural
  Information Processing Systems}, pages 7902--7912, 2018.

\bibitem{niemeyer2021giraffe}
Michael Niemeyer and Andreas Geiger.
\newblock Giraffe: Representing scenes as compositional generative neural
  feature fields.
\newblock In {\em Proceedings of the IEEE/CVF Conference on Computer Vision and
  Pattern Recognition}, pages 11453--11464, 2021.

\bibitem{niemeyer2020differentiable}
Michael Niemeyer, Lars Mescheder, Michael Oechsle, and Andreas Geiger.
\newblock Differentiable volumetric rendering: Learning implicit {3D}
  representations without {3D} supervision.
\newblock In {\em Proceedings of the IEEE/CVF Conference on Computer Vision and
  Pattern Recognition}, pages 3504--3515, 2020.

\bibitem{obukhov2020torchfidelity}
Anton Obukhov, Maximilian Seitzer, Po-Wei Wu, Semen Zhydenko, Jonathan Kyl, and
  Elvis Yu-Jing Lin.
\newblock High-fidelity performance metrics for generative models in {PyTorch},
  2020.
\newblock Version: 0.3.0, DOI: 10.5281/zenodo.4957738 (Apache License).

\bibitem{oechsle2021unisurf}
Michael Oechsle, Songyou Peng, and Andreas Geiger.
\newblock Unisurf: Unifying neural implicit surfaces and radiance fields for
  multi-view reconstruction.
\newblock {\em arXiv preprint arXiv:2104.10078}, 2021.

\bibitem{park2019deepsdf}
Jeong~Joon Park, Peter Florence, Julian Straub, Richard Newcombe, and Steven
  Lovegrove.
\newblock {DeepSDF}: Learning continuous signed distance functions for shape
  representation.
\newblock In {\em Proceedings of the IEEE/CVF Conference on Computer Vision and
  Pattern Recognition}, pages 165--174, 2019.

\bibitem{pytorch}
Adam Paszke, Sam Gross, Francisco Massa, Adam Lerer, James Bradbury, Gregory
  Chanan, Trevor Killeen, Zeming Lin, Natalia Gimelshein, Luca Antiga, Alban
  Desmaison, Andreas Kopf, Edward Yang, Zachary DeVito, Martin Raison, Alykhan
  Tejani, Sasank Chilamkurthy, Benoit Steiner, Lu Fang, Junjie Bai, and Soumith
  Chintala.
\newblock Pytorch: An imperative style, high-performance deep learning library.
\newblock In H. Wallach, H. Larochelle, A. Beygelzimer, F. d'Alch'{e} Buc, E.
  Fox, and R. Garnett, editors, {\em Advances in Neural Information Processing
  Systems 32}, pages 8024--8035. Curran Associates, Inc., 2019.

\bibitem{pavllo2020convolutional}
Dario Pavllo, Graham Spinks, Thomas Hofmann, Marie-Francine Moens, and Aurelien
  Lucchi.
\newblock Convolutional generation of textured {3D} meshes.
\newblock {\em arXiv preprint arXiv:2006.07660}, 2020.

\bibitem{qi2017pointnet}
Charles~R Qi, Hao Su, Kaichun Mo, and Leonidas~J Guibas.
\newblock Pointnet: Deep learning on point sets for {3D} classification and
  segmentation.
\newblock In {\em Proceedings of the IEEE conference on computer vision and
  pattern recognition}, pages 652--660, 2017.

\bibitem{radford2015unsupervised}
Alec Radford, Luke Metz, and Soumith Chintala.
\newblock Unsupervised representation learning with deep convolutional
  generative adversarial networks.
\newblock {\em arXiv preprint arXiv:1511.06434}, 2015.

\bibitem{rebain2021derf}
Daniel Rebain, Wei Jiang, Soroosh Yazdani, Ke Li, Kwang~Moo Yi, and Andrea
  Tagliasacchi.
\newblock {DeRF}: Decomposed radiance fields.
\newblock In {\em Proceedings of the IEEE/CVF Conference on Computer Vision and
  Pattern Recognition}, pages 14153--14161, 2021.

\bibitem{reiser2021kilonerf}
Christian Reiser, Songyou Peng, Yiyi Liao, and Andreas Geiger.
\newblock Kilo{NeRF}: Speeding up neural radiance fields with thousands of tiny
  mlps.
\newblock {\em arXiv preprint arXiv:2103.13744}, 2021.

\bibitem{rezende2014stochastic}
Danilo~Jimenez Rezende, Shakir Mohamed, and Daan Wierstra.
\newblock Stochastic backpropagation and approximate inference in deep
  generative models.
\newblock In {\em International conference on machine learning}, pages
  1278--1286. PMLR, 2014.

\bibitem{romaszko2017vision}
Lukasz Romaszko, Christopher~KI Williams, Pol Moreno, and Pushmeet Kohli.
\newblock Vision-as-inverse-graphics: Obtaining a rich {3D} explanation of a
  scene from a single image.
\newblock In {\em Proceedings of the IEEE International Conference on Computer
  Vision Workshops}, pages 851--859, 2017.

\bibitem{rosenfeld1968distance}
Azriel Rosenfeld and John~L Pfaltz.
\newblock Distance functions on digital pictures.
\newblock {\em Pattern recognition}, 1(1):33--61, 1968.

\bibitem{roth2017stabilizing}
Kevin Roth, Aurelien Lucchi, Sebastian Nowozin, and Thomas Hofmann.
\newblock Stabilizing training of generative adversarial networks through
  regularization.
\newblock {\em arXiv preprint arXiv:1705.09367}, 2017.

\bibitem{schwarz2020graf}
Katja Schwarz, Yiyi Liao, Michael Niemeyer, and Andreas Geiger.
\newblock Graf: Generative radiance fields for {3D}-aware image synthesis.
\newblock {\em arXiv preprint arXiv:2007.02442}, 2020.

\bibitem{polyscope}
Nicholas Sharp et~al.
\newblock Polyscope, 2019.
\newblock \url{www.polyscope.run} v1.2.0. (MIT License).

\bibitem{shin2018pixels}
Daeyun Shin, Charless~C Fowlkes, and Derek Hoiem.
\newblock Pixels, voxels, and views: A study of shape representations for
  single view {3D} object shape prediction.
\newblock In {\em Proceedings of the IEEE conference on computer vision and
  pattern recognition}, pages 3061--3069, 2018.

\bibitem{sitzmann2020implicit}
Vincent Sitzmann, Julien Martel, Alexander Bergman, David Lindell, and Gordon
  Wetzstein.
\newblock Implicit neural representations with periodic activation functions.
\newblock {\em Advances in Neural Information Processing Systems}, 33, 2020.

\bibitem{sitzmann2021light}
Vincent Sitzmann, Semon Rezchikov, William~T Freeman, Joshua~B Tenenbaum, and
  Fredo Durand.
\newblock Light field networks: Neural scene representations with
  single-evaluation rendering.
\newblock {\em arXiv preprint arXiv:2106.02634}, 2021.

\bibitem{sitzmann2019scene}
Vincent Sitzmann, Michael Zollh{\"o}fer, and Gordon Wetzstein.
\newblock Scene representation networks: Continuous {3D}-structure-aware neural
  scene representations.
\newblock {\em arXiv preprint arXiv:1906.01618}, 2019.

\bibitem{stanfordscanningrepo}
{{S}tanford Computer Graphics Laboratory}.
\newblock The {S}tanford 3{D} scanning repository.
\newblock \url{http://graphics.stanford.edu/data/3Dscanrep/}.
\newblock Accessed: 09/08/21.

\bibitem{takikawa2021neural}
Towaki Takikawa, Joey Litalien, Kangxue Yin, Karsten Kreis, Charles Loop, Derek
  Nowrouzezahrai, Alec Jacobson, Morgan McGuire, and Sanja Fidler.
\newblock Neural geometric level of detail: Real-time rendering with implicit
  {3D} shapes.
\newblock In {\em Proceedings of the IEEE/CVF Conference on Computer Vision and
  Pattern Recognition}, pages 11358--11367, 2021.

\bibitem{tatarchenko2019single}
Maxim Tatarchenko, Stephan~R Richter, Ren{\'e} Ranftl, Zhuwen Li, Vladlen
  Koltun, and Thomas Brox.
\newblock What do single-view {3D} reconstruction networks learn?
\newblock In {\em Proceedings of the IEEE/CVF Conference on Computer Vision and
  Pattern Recognition}, pages 3405--3414, 2019.

\bibitem{tulsiani2020implicit}
Shubham Tulsiani, Nilesh Kulkarni, and Abhinav Gupta.
\newblock Implicit mesh reconstruction from unannotated image collections.
\newblock {\em arXiv preprint arXiv:2007.08504}, 2020.

\bibitem{tulsiani2017multi}
Shubham Tulsiani, Tinghui Zhou, Alexei~A Efros, and Jitendra Malik.
\newblock Multi-view supervision for single-view reconstruction via
  differentiable ray consistency.
\newblock In {\em Proceedings of the IEEE conference on computer vision and
  pattern recognition}, pages 2626--2634, 2017.

\bibitem{turk1994zippered}
Greg Turk and Marc Levoy.
\newblock Zippered polygon meshes from range images.
\newblock In {\em Proceedings of the 21st annual conference on Computer
  graphics and interactive techniques}, pages 311--318, 1994.

\bibitem{venkatesh2020dude}
Rahul Venkatesh, Sarthak Sharma, Aurobrata Ghosh, Laszlo Jeni, and Maneesh
  Singh.
\newblock Dude: Deep unsigned distance embeddings for hi-fidelity
  representation of complex {3D} surfaces.
\newblock {\em arXiv preprint arXiv:2011.02570}, 2020.

\bibitem{wang2018pixel2mesh}
Nanyang Wang, Yinda Zhang, Zhuwen Li, Yanwei Fu, Wei Liu, and Yu-Gang Jiang.
\newblock Pixel2mesh: Generating {3D} mesh models from single {RGB} images.
\newblock In {\em Proceedings of the European Conference on Computer Vision
  (ECCV)}, pages 52--67, 2018.

\bibitem{wang2020pixel2mesh}
Nanyang Wang, Yinda Zhang, Zhuwen Li, Yanwei Fu, Hang Yu, Wei Liu, Xiangyang
  Xue, and Yu-Gang Jiang.
\newblock Pixel2mesh: {3D} mesh model generation via image guided deformation.
\newblock {\em IEEE transactions on pattern analysis and machine intelligence},
  2020.

\bibitem{wang2021neus}
Peng Wang, Lingjie Liu, Yuan Liu, Christian Theobalt, Taku Komura, and Wenping
  Wang.
\newblock Neus: Learning neural implicit surfaces by volume rendering for
  multi-view reconstruction.
\newblock {\em arXiv preprint arXiv:2106.10689}, 2021.

\bibitem{wu2017marrnet}
Jiajun Wu, Yifan Wang, Tianfan Xue, Xingyuan Sun, Bill Freeman, and Josh
  Tenenbaum.
\newblock Marrnet: {3D} shape reconstruction via {2.5D} sketches.
\newblock {\em Advances in Neural Information Processing Systems}, 30:540--550,
  2017.

\bibitem{yan2016perspective}
Xinchen Yan, Jimei Yang, Ersin Yumer, Yijie Guo, and Honglak Lee.
\newblock Perspective transformer nets: Learning single-view {3D} object
  reconstruction without {3D} supervision.
\newblock {\em Advances in Neural Information Processing Systems},
  29:1696--1704, 2016.

\bibitem{yang2021deep}
Mingyue Yang, Yuxin Wen, Weikai Chen, Yongwei Chen, and Kui Jia.
\newblock Deep optimized priors for {3D} shape modeling and reconstruction.
\newblock In {\em Proceedings of the IEEE/CVF Conference on Computer Vision and
  Pattern Recognition}, pages 3269--3278, 2021.

\bibitem{yariv2021volume}
Lior Yariv, Jiatao Gu, Yoni Kasten, and Yaron Lipman.
\newblock Volume rendering of neural implicit surfaces.
\newblock {\em arXiv preprint arXiv:2106.12052}, 2021.

\bibitem{yu2021plenoctrees}
Alex Yu, Ruilong Li, Matthew Tancik, Hao Li, Ren Ng, and Angjoo Kanazawa.
\newblock Plenoctrees for real-time rendering of neural radiance fields.
\newblock {\em arXiv preprint arXiv:2103.14024}, 2021.

\bibitem{yu2021pixelnerf}
Alex Yu, Vickie Ye, Matthew Tancik, and Angjoo Kanazawa.
\newblock Pixel{NeRF}: Neural radiance fields from one or few images.
\newblock In {\em Proceedings of the IEEE/CVF Conference on Computer Vision and
  Pattern Recognition}, pages 4578--4587, 2021.

\bibitem{yuille2006vision}
Alan Yuille and Daniel Kersten.
\newblock Vision as {Bayesian} inference: analysis by synthesis?
\newblock {\em Trends in cognitive sciences}, 10(7):301--308, 2006.

\bibitem{CycleGAN2017}
Jun-Yan Zhu, Taesung Park, Phillip Isola, and Alexei~A Efros.
\newblock Unpaired image-to-image translation using cycle-consistent
  adversarial networks.
\newblock In {\em Computer Vision (ICCV), 2017 IEEE International Conference
  on}, 2017.

\bibitem{cyclegancode}
Jun-Yan Zhu, Taesung Park, Phillip Isola, and Alexei~A Efros.
\newblock Cyclegan and pix2pix in {PyTorch}.
\newblock {\em junyanz/pytorch-CycleGAN-and-pix2pix}, Nov. 2021.
\newblock \url{https://github.com/junyanz/pytorch-CycleGAN-and-pix2pix} (BSD
  License).

\bibitem{zhu2018visual}
Jun-Yan Zhu, Zhoutong Zhang, Chengkai Zhang, Jiajun Wu, Antonio Torralba, Josh
  Tenenbaum, and Bill Freeman.
\newblock Visual object networks: Image generation with disentangled {3D}
  representations.
\newblock {\em Advances in Neural Information Processing Systems}, 31:118--129,
  2018.

\bibitem{zobeidi2021deep}
Ehsan Zobeidi and Nikolay Atanasov.
\newblock A deep signed directional distance function for object shape
  representation.
\newblock {\em arXiv preprint arXiv:2107.11024}, 2021.

\end{thebibliography}
